%% file: main.tex
\documentclass[runningheads]{llncs}

\usepackage{graphicx}
\usepackage{algorithm2e}
\usepackage{amsmath,amssymb,amsfonts}
\usepackage{algorithmic}
\usepackage{subfigure}
\usepackage{tikz}
\usepackage{pgfplots}
\usepackage{multirow}
\usepackage{color}

\SetKwComment{Comment}{/* }{ */}

\begin{document}

\title{Generation of Non-Deterministic Synthetic Face Datasets Guided by Identity Priors}

\author{Marcel Grimmer\inst{1} \and Haoyu Zhang \inst{1} \and Raghavendra Ramachandra \inst{1} \and Kiran Raja \inst{1} \and
Christoph Busch\inst{1, 2}}
\authorrunning{Grimmer et al.}

\institute{NBL - Norwegian Biometrics Laboratory, NTNU, Norway \and
da/sec - Biometrics and Internet Securiy Research Group, HDA, Germany}

\maketitle              

\input{content/00-abstract}
\input{content/01-introduction}

\input{content/02-sota}
\input{content/03-proposed}

\input{content/04-settings}
\input{content/05-experiments}

\input{content/06-conclusion}

\newpage
\input{content/appendix}

\bibliographystyle{splncs04}
\bibliography{references.bib}

\end{document}

%% file: content/00-abstract.tex
\begin{abstract}
Enabling highly secure applications (such as border crossing) with face recognition requires extensive biometric performance tests through large scale data. However, using real face images raises concerns about privacy as the laws do not allow the images to be used for other purposes than originally intended. Using representative and subsets of face data can also lead to unwanted demographic biases and cause an imbalance in datasets. One possible solution to overcome these issues is to replace real face images with synthetically generated samples. While generating synthetic images has benefited from recent advancements in computer vision, generating multiple samples of the same synthetic identity resembling real-world variations is still unaddressed, i.e., mated samples. This work proposes a non-deterministic method for generating mated face images by exploiting the well-structured latent space of StyleGAN. Mated samples are generated by manipulating latent vectors, and more precisely, we exploit Principal Component Analysis (PCA) to define semantically meaningful directions in the latent space and control the similarity between the original and the mated samples using a pre-trained face recognition system. We create a new dataset of synthetic face images (SymFace) consisting of 77,034 samples including 25,919 synthetic IDs. Through our analysis using well-established face image quality metrics, we demonstrate the differences in the biometric quality of synthetic samples mimicking characteristics of real biometric data. The analysis and results thereof indicate the use of synthetic samples created using the proposed approach as a viable alternative to replacing real biometric data. 
\keywords{Biometrics, Face recognition, Synthetic Face Image Generation, Deep learning,  StyleGAN}
\end{abstract}

%% file: content/01-introduction.tex
\section{Introduction}

\input{images/intra-identity-variation}

The popularity of biometric recognition has increased steadily along with the development of more accurate and convenient recognition technologies. According to ISO/IEC 2382-37:2017 \cite{ISO_2382_37}, biometrics refers to the automated recognition of individuals based on their biological and behavioural characteristics. In particular, the human face has proven to be sufficiently unique and an easy-to-capture biometric characteristic, leading to a wide range of real-world applications, including border control, passport issuance, and civilian ID management. Driven by the promising performance of current face recognition systems, the Smart Borders program has been initiated within the European Union to establish the Entry-Exit System (EES)~\cite{EU-Regulation-EES-InternalDocument-2017}, an automated IT system for registering travellers from third-countries, replacing the current system of manual stamping of passports. This system aims to help bona fide third-country nationals travel more easily while also identifying more efficiently over-stayers and cases of document and identity fraud. To perform automatically, EES will register the person's name, type of the travel document and biometric data (face images and/or fingerprints).

A requirement for deploying biometric recognition at the European borders is complying with the high standards defined in the best practices for automated border control of the European Border and Coast Guard Agency (Frontex)~\cite{FRONTEX-BorderControl-BestPractices-InternalDocument-2015}. The compliance with these guidelines must be validated by conducting large-scale biometric performance tests which require large datasets. As the collection of real face images is expensive, time-consuming, and privacy-concerning, generating synthetic face images has become an attractive and viable alternative. Driven by the advancements in technology, approaches like StyleGAN and StyleGAN2 \cite{karras2019style}\cite{karras2020analyzing} have shown promises to create large scale face datasets with unique identities. 

While the synthetic image generation approaches are well used in various applications, the applicability of those images in biometrics is limited. Specifically, the biometric data used for training algorithms and performance testing need to mimic the real data with variations in pose, varying expressions, occlusions and illumination changes reflecting realistic conditions for any particular identity. In essence, each synthetic identity should accompany a set of variations that can compose what is referred to as mated samples for obtaining comparison scores. Specifically, the synthetic data should represent intra-class variations similar to bona fide data while preserving the identity information.  The mated samples essentially are required to generate the genuine score distribution to gauge the biometric performance such as False Match Rate (FMR) and False Non-Match Rate (FNMR). However, despite the recent advancements of synthetic image generation~\cite{karras2019style}\cite{karras2020analyzing}, it continues to be a technical issue to create synthetic datasets with mated samples that are representative and comparable to real face images captured at border control scenarios (e.g. frontal head poses without face occlusions). 

\subsection{Our contributions}

This work tackles the above-described challenge by introducing a new technique for generating synthetic mated samples. More precisely, a pre-trained StyleGAN generator~\cite{karras2019style} is utilised to generate synthetic face images of distinct synthetic individuals ("base images"). Each base image is represented by a latent vector $w_{1\times512}$, acting as a compressed version of the original image and reflecting the internal data representation learned by StyleGAN. Motivated by the idea of editing facial attributes by shifting the corresponding latent vector in a specific direction in the latent space~\cite{shen2020interfacegan}, we propose to generate mated faces in a non-deterministic manner. We assert that such an approach for attribute editing leads to a better approximation of the natural intra-identity variation of bona fide mated samples, as can be compared in Figure \ref{fig:intra-identity-variation}.

As the components of the latent vector space can represent various possible semantics, the principal components can be interpreted as semantically meaningful directions in the latent space of StyleGAN. Concretely, extracting the Principal Components ~\cite{Pearson-PCA-Misc-1901} from a latent vector space of $50,000$ to 512 leads to obtaining semantically meaningful values. Inspired by such an argument, we create the mated samples by shifting the latent vectors into the directions given by the most relevant eigenvectors (i.e. the principal components). However, as the latent vectors are moved farther from their original locations, the risk of losing the identity information increases, we, therefore, employ a pre-trained face recognition system (FR)~\cite{Deng-ArcFace-CVPR-2019} to obtain the distance between the original and edited image dynamically to ensure the preservation of identity information from mated samples for the original identity used for editing. We refer to non-deterministic face editing as changing multiple semantics in an unsupervised manner, as opposed to controlled face editing, where specific facial attributes are chosen to be edited. 

With such a rationale of our proposed approach, we create a new dataset of face images with synthetic identities and mated samples for each identity in this work which we refer to as Synthetic Mated Face Dataset (SymFace Dataset). The dataset consists of $77,034$ samples with an average number of three mated samples per synthetic identity. To better approximate a semi-controlled capturing environment, images with extreme characteristics are sorted out, taking into account illumination conditions, head poses rotation and inter-eye distance. Also, the study concentrates on adult face images due to the limited training data available from young children and seniors. We refer to Figure \ref{fig:filtered-out-imgs} to get an impression of typical images filtered out by our filtering pipeline.

We further evaluate the quality of our proposed synthetic dataset by comparing its properties to real face images taken from FRGC v2.0~\cite{FRGC_DB}. Among other approaches for conducting such an analysis, we translate the biometric quality of each image to a quality score between [0, 100] using Face Quality Assessment Algorithms (FQAAs)~\cite{ISO_29794_5_2021}. In this context, a high-quality score indicates that the corresponding biometric sample is well suited for biometric recognition. On the opposite, low-quality scores deteriorate the recognition accuracy due to the low quality of the input image. This understanding of biometric quality corresponds to the terminology specified by ISO/IEC 29794-1~\cite{ISO_29794_1:2016}, defining the utility of a biometric sample as the prediction of the biometric recognition performance. In this work, two FQAAs are used for estimating and comparing the biometric quality: FaceQnet v1~\cite{Hernandez-faceqnetv1-arxiv-2020} and SER-FIQ~\cite{Terhorst-SERFIQ-CVPR-2020}. At this point, the reader is referred to Section~\ref{sec:sota} to obtain a more detailed description of these methods. 

In the rest of the paper, Section 2 summarises the conceptual ideas of generating synthetic face images and mated samples. Next, Section 3 provides a detailed description of the proposed PCA-FR-Guided sampling approach. Section 4 details the newly created SymFace Dataset, and finally, Section 5 gives an overview of the experimental results, followed by a conclusion about the key findings in Section 6.

%% file: images/intra-identity-variation.tex
\begin{figure*}[h]
\centering
\subfigure{
    \begin{minipage}[t]{0.15\linewidth}
    \includegraphics[width=0.77in]{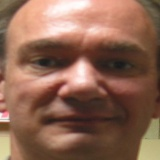}
    \end{minipage}%
}%
\subfigure{
    \begin{minipage}[t]{0.15\linewidth}
    \includegraphics[width=0.77in]{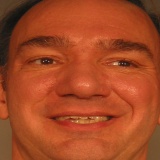}
    \end{minipage}%
}%
\subfigure{
    \begin{minipage}[t]{0.15\linewidth}
    \includegraphics[width=0.77in]{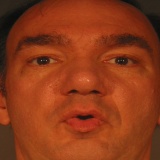}
    \end{minipage}%
}%
\subfigure{
    \begin{minipage}[t]{0.15\linewidth}
    \includegraphics[width=0.77in]{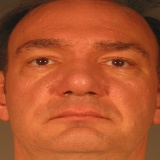}
    \end{minipage}%
}%
\subfigure{
    \begin{minipage}[t]{0.15\linewidth}
    \includegraphics[width=0.77in]{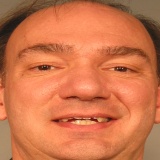}
    \end{minipage}%
}%

\subfigure{
    \begin{minipage}[t]{0.15\linewidth}
    \includegraphics[width=0.77in]{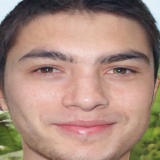}
    \end{minipage}%
}%
\subfigure{
    \begin{minipage}[t]{0.15\linewidth}
    \includegraphics[width=0.77in]{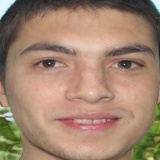}
    \end{minipage}%
}%
\subfigure{
    \begin{minipage}[t]{0.15\linewidth}
    \includegraphics[width=0.77in]{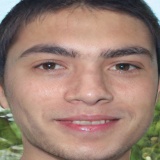}
    \end{minipage}%
}%
\subfigure{
    \begin{minipage}[t]{0.15\linewidth}
    \includegraphics[width=0.77in]{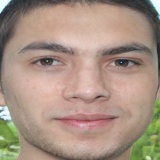}
    \end{minipage}%
}%
\subfigure{
    \begin{minipage}[t]{0.15\linewidth}
    \includegraphics[width=0.77in]{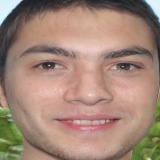}
    \end{minipage}%
}%
\caption{Comparison of the intra-identity variation between FRGC v2.0 (top) and SymFace (bottom).}
\label{fig:intra-identity-variation}
\end{figure*}

%% file: content/02-sota.tex
\section{Related Works}
\label{sec:sota}

\subsection{Synthetic Image Generation}
\label{sec:base_image_generation}
In 2019, Karras et al.~\cite{karras2019style} presented a style-based generator architecture for generative adversarial networks (StyleGAN), capable of generating synthetic images with high resolutions (1024x1024) and realistic appearances. In addition to their proposed GAN architecture, the authors
web crawled high-quality human face images from a social media platform (Flickr) to create a new dataset (FFHQ), covering a wide variation of soft biometrics.

Despite the recent success of deep generative networks, most generators are still operating as black-boxes and lack a deeper understanding of the latent space. To address these weaknesses and improve the disentanglement properties of the latent space, StyleGAN maps initially drawn latent vectors to an intermediate latent space, which turns out to encode facial features in a more disentangled manner. Further, Adaptive Instance Normalization (AdaIN)~\cite{Huang-AdaIn-CVPR-2017} enables to fuse the styles of different faces on multiple feature levels. Furthermore, stochastic variation is achieved by adding Gaussian noise to the feature maps after each convolution operation to vary fine-grained details. Recently, StyleGAN2 has been published by the same authors~\cite{karras2020analyzing}, improving the architectural design and fixing the characteristic artefacts occurring in the synthetic images generated by StyleGAN. 

In StyleGAN and StyleGAN2, synthetic images are generated by randomly sampling from a known distribution (latent space). If these latent vectors are drawn from tail regions of the distribution, the quality of the generated face images deteriorates while the diversity of facial attributes increases. To balance this trade-off, a truncation factor can be used to stabilise the sampling: the truncated latent code $w'$ is calculated as $w' = \Bar{w} + \psi (w-\Bar{w})$ where $\Bar{w}$ indicates the latent spaces' center of mass and $\psi$ denotes the truncation factor. Following the empirical analysis of Zhang et al.~\cite{Zhang-SynthData-FR-IWBF-2021}, we choose a truncation factor of $\psi=0.75$. In~\cite{Zhang-SynthData-FR-IWBF-2021}, the authors have shown that the biometric performance of synthetic samples generated with StyleGAN and StyleGAN2 are similar and comparable to bona fide images from FRGC v2.0~\cite{FRGC_DB}. Hence, this work uses StyleGAN for generating synthetic base images to enable the implementation of PCA-FR-Guided sampling to operate within the framework of InterFaceGAN~\cite{shen2020interfacegan}.

\subsection{Mated Sample Generation}

Though it has been shown in~\cite{Zhang-SynthData-FR-IWBF-2021} that single synthetic face images can achieve comparable performances as bona fide samples for face recognition, mated samples are more commonly required in biometric performance evaluations. Given a synthetic base image, mated samples can be derived by editing facial attributes to simulate the factors of variation present in bona fide samples. With the groundbreaking work of Shen et al.~\cite{shen2020interfacegan}, InterFaceGAN was introduced as a framework enabling editing facial attributes of synthetic identities through manipulating latent vectors in the latent space. In this context, the latent space reflects the internal data representation of StyleGAN and structures various semantics learned from the training dataset. Further, the innovative architecture of StyleGAN significantly reduces the entanglement of the encoded semantics, which provides optimal conditions for controlled modifications on facial attributes.

The main contribution of InterFaceGAN is based on the observation that the latent space can be divided into linear subspaces according to binary semantics, such as "smile" or "no smile". Concretely, linear Support Vector Machines (SVMs)~\cite{Cortes-SVM-MachineLearning-1995} are used to divide the latent space into subspaces for each facial attribute of interest. Once the SVMs are trained, facial attributes are modified by shifting the latent vectors into the perpendicular direction of the previously found boundaries, thereby causing continuous changes. The same principle has been adopted by Colbois et al.~\cite{colbois2021use}, who manipulate yaw angle, illumination, and a smile by approximating the bona fide conditions of Multi-PIE~\cite{Gross-MultiPIE-IVC-2010}.   

\subsection{Limitations in State-of-the-art}
Although InterFaceGAN generates visually appealing mated samples, their applicability for general biometric performance tests is still limited and understudied. As shown in Figure \ref{fig:intra-identity-variation}, mated samples collected in real-world scenarios naturally include several variations varying at the same time, for instance, pose, illumination, expression and a combination of them. In contrast, controlled face editing focuses on changing only a few semantics while leaving others fixed. Therefore, controlled modifications are useful to determine the vulnerability of face recognition systems for targeted semantics while only representing a small subset of the potential diversity in bona fide datasets. Motivated by this observation, we introduce PCA-FR-Guided sampling as a technique for generating non-deterministic mated samples to either replace or complement existing test datasets.   

%% file: content/03-proposed.tex
\section{PCA-FR-Guided Sampling}

This section introduces our new method for generating mated samples, which we refer to as PCA-FR-Guided Sampling. As described in Section~\ref{sec:sota}, semantic modifications can be caused by moving latent vectors in the latent space. However, this approach still leaves two questions unanswered: 1) How to choose semantically meaningful directions? 2) How to choose the distance to preserve identities while maximising the intra-identity variation?

\begin{figure}[h]
\centering
\includegraphics[width=\textwidth]{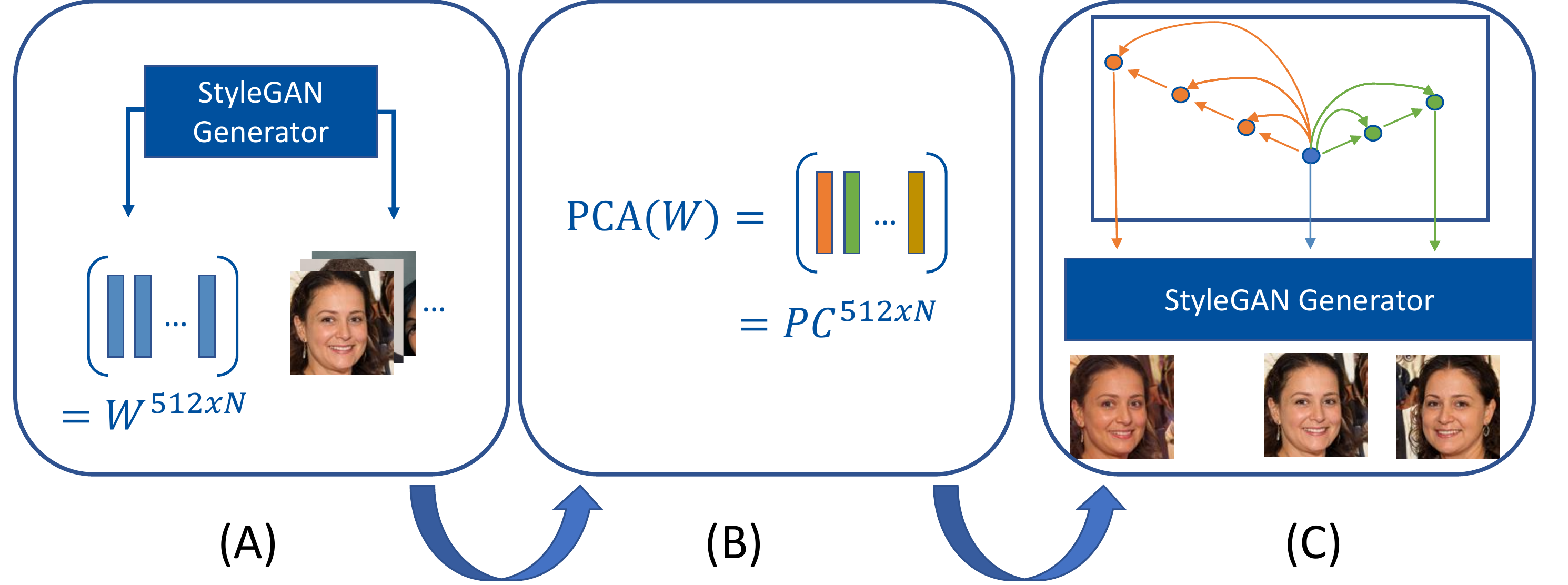}
\caption{Overview of the proposed PCA-FR-Guided Sampling with $N=50,000$ denoting the number of latent vectors concatenated as matrix $W$ to obtain the principal components (PCs). A detailed workflow is given in Algorithm \ref{alg:pca-fr-guided-sampling}.}
\label{fig:pca-fr-sampling-overview}
\end{figure}

Aiming to find solutions for the aforementioned questions, Figure~\ref{fig:pca-fr-sampling-overview} provides an overview of the PCA-FR-Guided sampling technique. After generating an initial synthetic dataset with StyleGAN with a truncation factor of $\psi=0.75$ (A), PCA is applied to extract semantically meaningful directions from the corresponding latent vectors (B). The idea is to extract the latent direction with the most variance, leading to effective variation after image generation. Finally, the latent vectors are moved along the principal component axes while adjusting the distance dynamically by measuring the similarity between the original and the shifted mated sample in a step-wise manner (C). Algorithm~\ref{alg:pca-fr-guided-sampling} provides a detailed workflow of the PCA-FR-Guided mated sample generation process proposed in this work. 

\begin{algorithm}
\SetKwRepeat{Do}{do}{while}
\SetKwInOut{Input}{input}
\SetKw{KwIn}{in}
\Input{$latentVectors, \ N, \ stepSize, \ threshold, \ Generator$}
$components = PCA(latentVectors)$\;
\For{$w$ \KwIn $latentVectors$}{
    $img = Generator(v)$\;
    \For{$c$ \KwIn $components$}{
        $i = 1$\;
        \Do{$recognised$}{
            $w\_moved = shift\_in\_lspace(w, c, stepSize\cdot i)$\;
            $mated\_img = Generator(w\_moved)$\;
            $recognised = ArcFace(img, mated\_img, threshold)$\;
            $i = i + 1$\;
            \uIf{$recognised$}{
                $save(mated\_img)$\;
                }
            }
        }
    }
\caption{PCA-FR-Guided sampling algorithm for generating mated-samples.}
\label{alg:pca-fr-guided-sampling}
\end{algorithm}

We specifically employ $stepSize$ and the verification $threshold$ as controlling parameters to balance the trade-off between intra-class variation and identity-retaining factor for generated mated samples. In other words, increasing the comparison threshold decreases the distance between the original latent vector and the shifted latent vector, thus generating more similar faces with fewer factors of variation. On the other hand, decreasing the step size approaches the given threshold with smaller steps, thus yielding mated samples closer to the desired similarity tolerance \footnote{We have chosen $stepSize=0.2$ and $threshold=0.8$ empirically, considering the quality of the mated samples and the algorithm's efficiency. However, other values can also be used on application scenarios.}.

%% file: content/04-settings.tex
\section{Synthetic Mated Face (SymFace) Dataset}
\label{sec:dataset}

This section describes the structure of our synthetic mated face dataset (SymFace) and the reference dataset used for the comparison part in Section~\ref{sec:experiments}. 

Each mated sample is generated based on a synthetic face image randomly generated by StyleGAN. As StyleGAN was trained using images crawled from social media, the diversity of the generated images roughly corresponds to approximate capturing scenarios "in the wild". As described in section~\ref{sec:base_image_generation}, a truncation factor of $\psi=0.75$ was chosen to generate 50,000 unique identity images with high resolutions of 1024x1024 pixels and this is referred to as base images.

However, not all images generated from StyleGAN satisfy the minimum criteria needed for biometric applications. For instance, in a border-crossing scenario, factors such as minimum inter-eye distance (IED), illumination metrics, predicted head poses~\cite{albiero2021img2pose}, and estimated ages~\cite{zhang2019c3ae} are needed in accordance to  ISO/IEC TR 29794-5:2010~\cite{ISO_29794_5} and ICAO 9303~\cite{ICAO9303}. Accounting for this, we discard all such images not meeting the criteria of minimum inter-eye distance (IED), illumination metrics, predicted head poses~\cite{albiero2021img2pose}, and estimated ages~\cite{zhang2019c3ae}. The SymFace Dataset thus has a total of 25,919 images which we deem as usable for further analysis in this work, and a sample of such images that are eliminated by our filtering pipeline is illustrated in Figure \ref{fig:filtered-out-imgs}. As it can be observed from Figure \ref{fig:filtered-out-imgs}, despite these images looking visually pleasing, they fail to meet the quality standards with respect to ISO/IEC TR 29794-5:2010~\cite{ISO_29794_5} and ICAO 9303~\cite{ICAO9303}.

\input{images/filtered-out-imgs}

Finally, the filtered base images are used as a basis for generating two mated samples for each synthetic identity by using our proposed PCA-FR-Guided sampling technique. Though we selected the first and second principal components, our experiments indicate that each of the 512 principal components can be used to obtain semantically meaningful mated samples. In addition, we apply InterFaceGAN to create three additional datasets, each of which includes mated samples with single semantics edited (yaw angle, illumination quality, and smile).
\input{tables/datasets}

\subsection{Reference Biometric Dataset}
\begin{figure}
    \centering
    \includegraphics[width=1\linewidth]{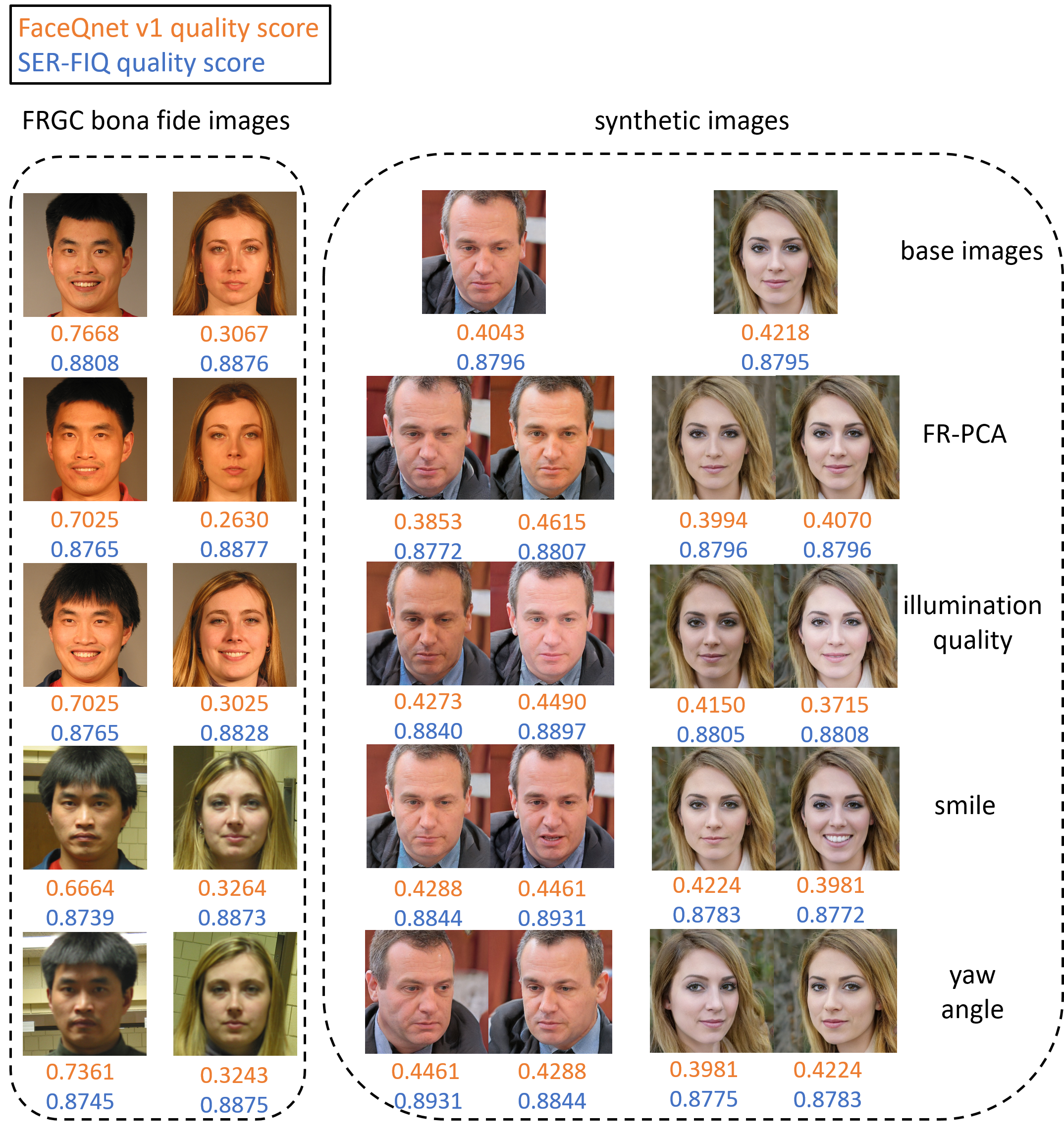}
    \caption{Examples images of bona fide and synthetic images evaluated in Section \ref{sec:experiments}.}
    \label{fig:example-images}
\end{figure}

Further, we employ FRGC v2.0~\cite{FRGC_DB} as a reference dataset, including 24,025 bona fide images captured in constrained conditions that resemble the image quality in a border-crossing scenario. Finally, we analyse biometric use cases of the SymFace dataset by studying the characteristics and comparing the same against the FRGC v2.0 dataset. A concise overview of the above-described datasets is given in Table \ref{tab:datasets}, listing the number of samples counted during different development stages. Moreover, Figure \ref{fig:example-images} presents example images extracted from all datasets, annotated with quality scores obtained by SER-FIQ and FaceQnet v1.

%% file: images/filtered-out-imgs.tex
\begin{figure*}[h]
\centering
\subfigure{
    \begin{minipage}[t]{0.15\linewidth}
    \includegraphics[width=0.7in]{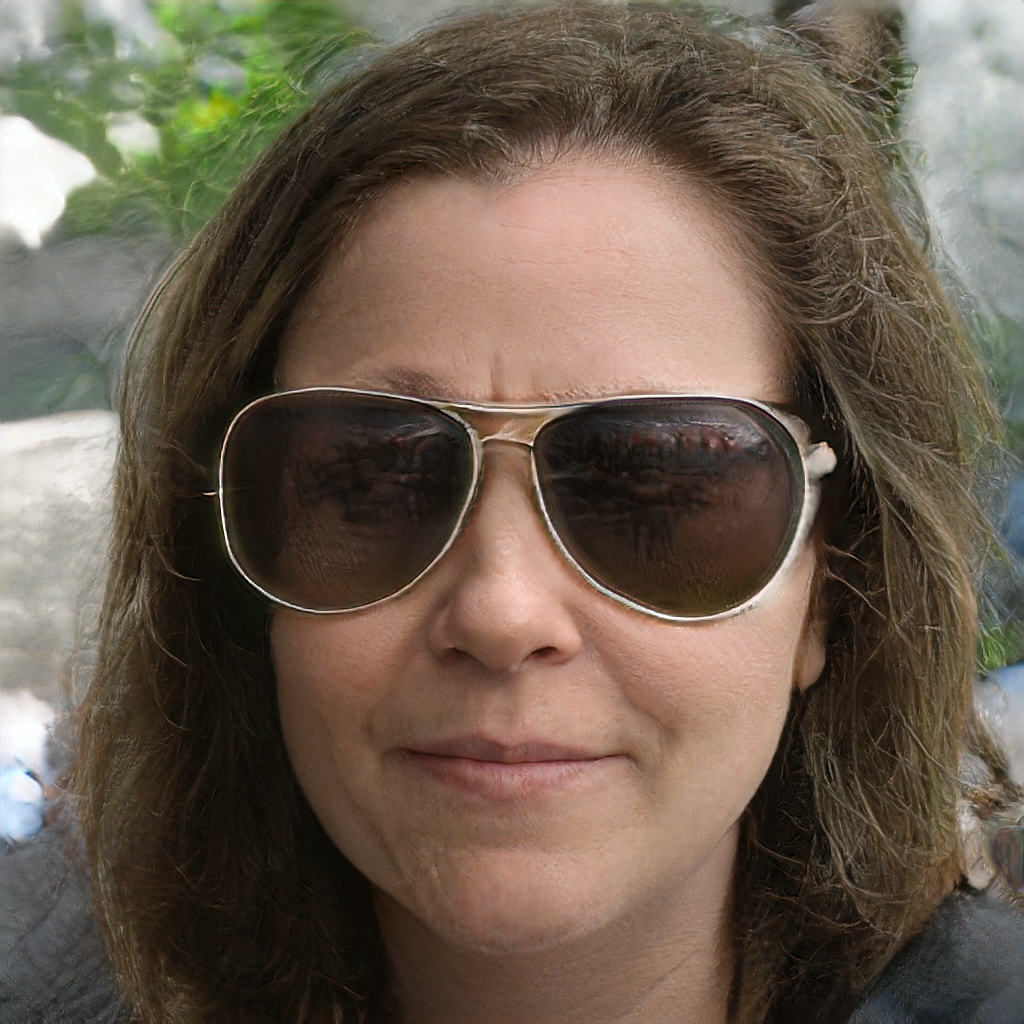}
    \end{minipage}%
}%
\subfigure{
    \begin{minipage}[t]{0.15\linewidth}
    \includegraphics[width=0.7in]{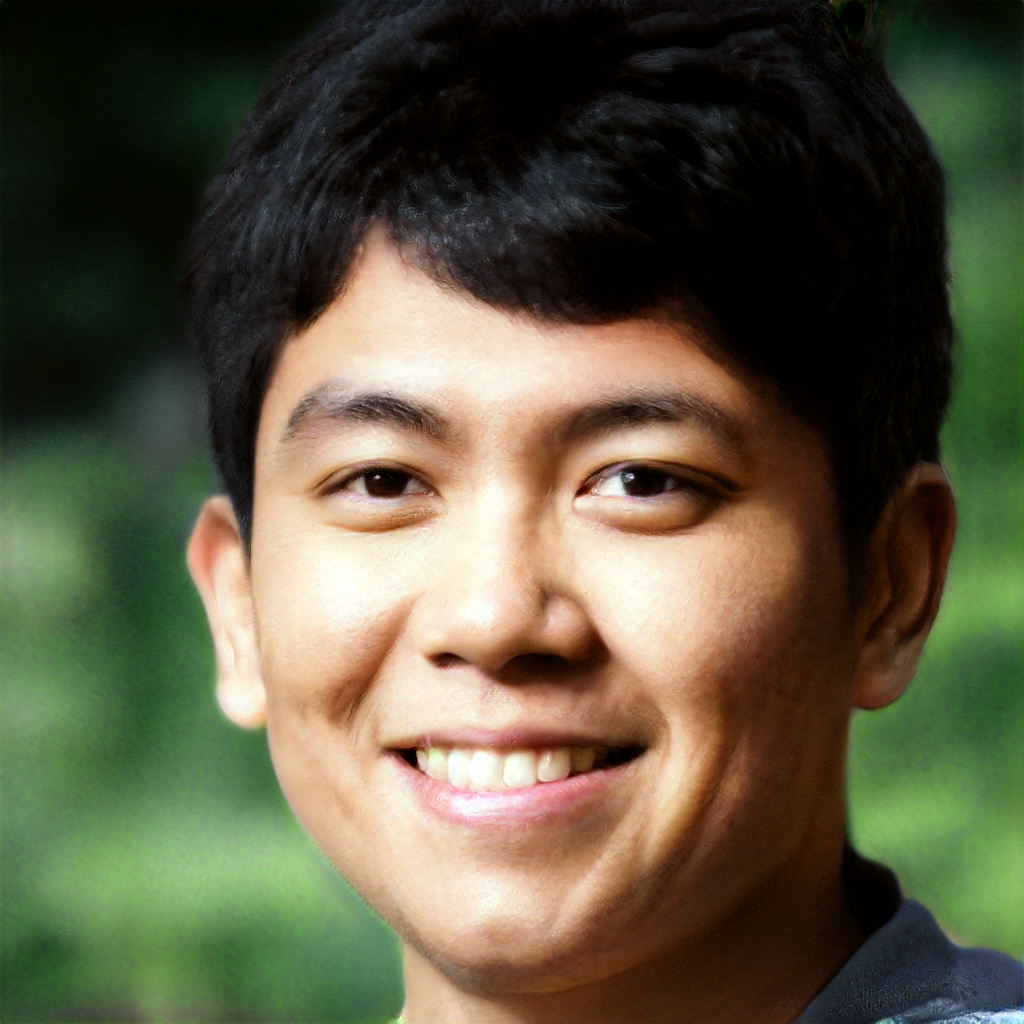}
    \end{minipage}%
}%
\subfigure{
    \begin{minipage}[t]{0.15\linewidth}
    \includegraphics[width=0.7in]{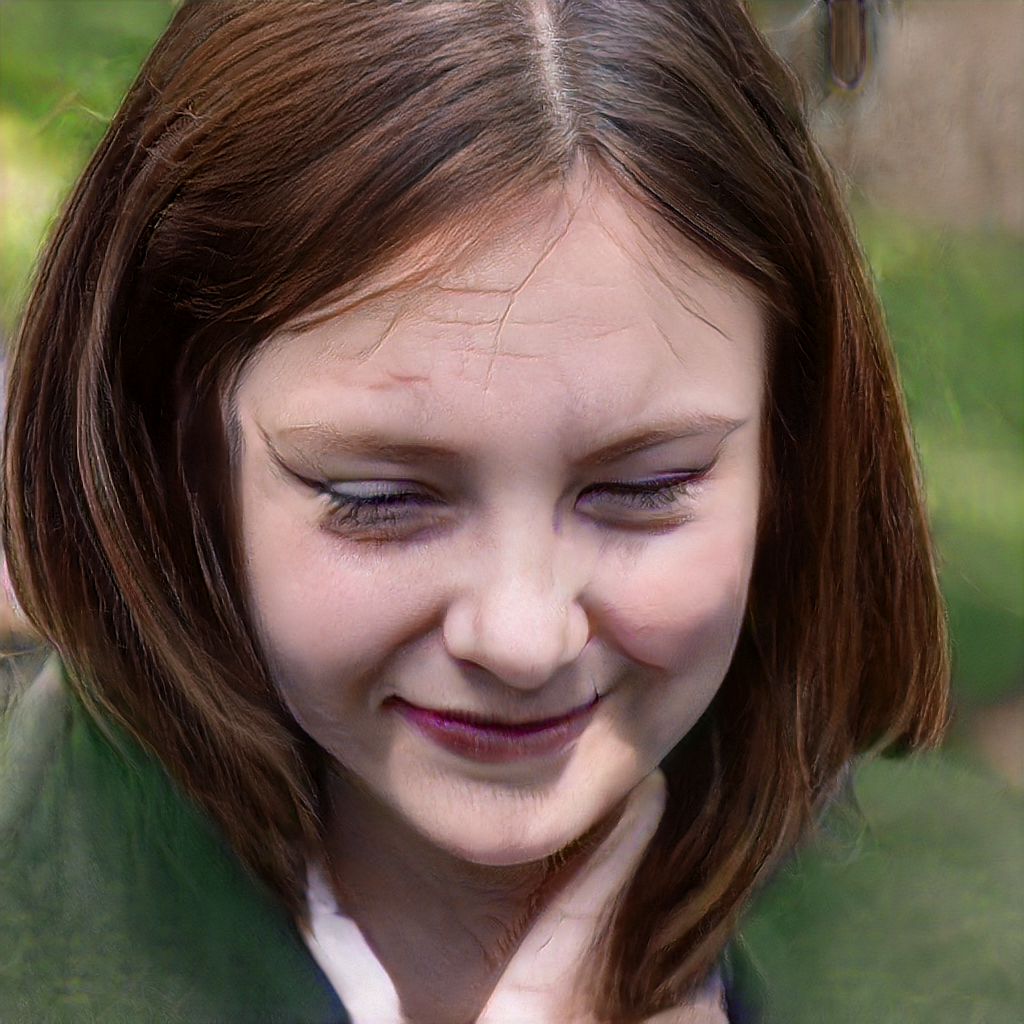}
    \end{minipage}%
}%
\subfigure{
    \begin{minipage}[t]{0.15\linewidth}
    \includegraphics[width=0.7in]{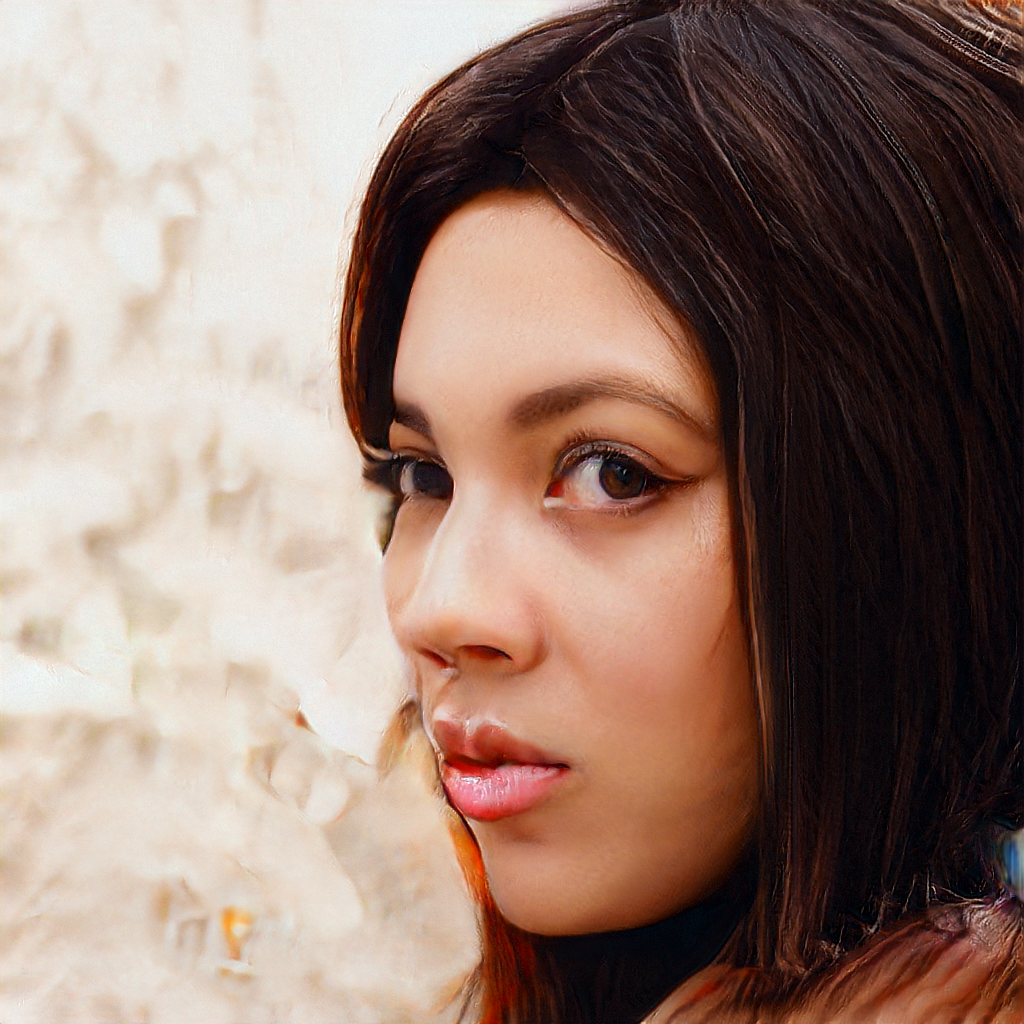}
    \end{minipage}%
}%
\subfigure{
    \begin{minipage}[t]{0.15\linewidth}
    \includegraphics[width=0.7in]{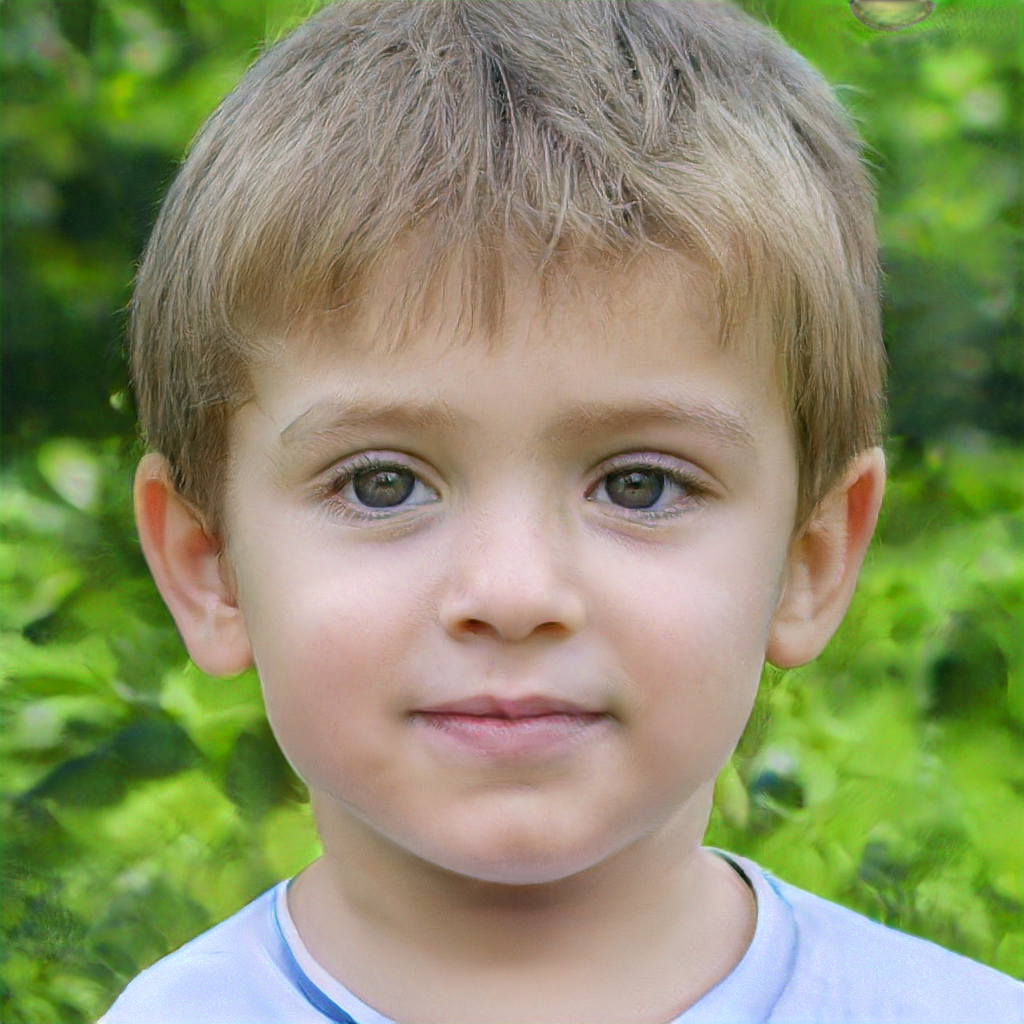}
    \end{minipage}%
}%
\caption{Low quality images filtered out by our filtering pipeline - from left to right: IED, illumination, pitch angle, yaw angle, age.}
\label{fig:filtered-out-imgs}
\end{figure*}

%% file: tables/datasets.tex
\begin{table}[]
\centering
\begin{tabular}{|l|c|c|c|c|c|}
\hline
                & \multicolumn{1}{l|}{SymFace} & \multicolumn{1}{l|}{FRGC v2.0} & \multicolumn{1}{l|}{Illumination Quality} & \multicolumn{1}{l|}{Smile} & \multicolumn{1}{l|}{Yaw} \\ \hline
\# Base Images  & 50,000                       & /                              & 50,000                                    & 50,000                     & 50,000                   \\ \hline

- Filtering     & 25,919                       & /                              & 25,919                                    & 25,919                     & 25,919                   \\ \hline

+ Mated Samples & 77,757                      & 24,025                         & 77,757                                     & 77,757                     & 77,757                   \\ \hline

- Filtering     & 77,034                       & 17,919                         & 74,183                                   & 74,574                     & 60,504                   \\ \hline

\end{tabular}
\caption{Dataset sizes in different development stages after applying our filtering pipeline and generating mated samples.}
\label{tab:datasets}
\end{table}

%% file: content/05-experiments.tex
\section{Experimental Results}
\label{sec:experiments}

The biometric utility of the synthetic database, especially for mated samples, can be evaluated by measuring the biometric performance or by validating the quality of the samples according to well-established face image quality metrics. We employ both approaches by first evaluating the Face quality assessment algorithms (FQAAs) on the newly created SymFace Dataset and compare it against similar characteristics of the FRGC v2.0 dataset. We then evaluate the mated and non-mated comparison score distributions obtained by applying the pre-trained VGGFace2~\cite{Qiong-VGGFace2-arxiv-2018} face recognition model to verify the biometric utility by analysing the score distribution. We provide a summary of the employed FQAAs for the convenience of the reader.

\subsection{Face Image Quality Assessment}
Face quality assessment algorithms (FQAAs) are used as indicators of how the quality of a face image contributes to the overall accuracy of a face recognition system. In this work, two representative FQAAs are utilised to evaluate the generated mated samples' biometric quality:

\begin{itemize}
    \item \textbf{FaceQnet v1} is a deep learning-based FQAA proposed by Hernandez-Ortega et al.~\cite{Hernandez-faceqnetv1-arxiv-2020}, aiming to predict the general utility of a face image, independent of a specific face recognition system. For the quality score prediction, a pre-trained network of ResNet-50~\cite{He-ResNet-CVPR-2016}is fine-tuned as a feature extractor on a small subset of the VGGFace2 dataset~\cite{Qiong-VGGFace2-arxiv-2018}, including 300 subjects. FaceQnet v1 follows a supervised learning approach, which means that the ground truth quality scores are required for fine-tuning the model. However, finding representative quality scores that accurately reflect general utility criteria is a challenging task. Therefore, the authors propose to determine the utility of an image by comparing it to an ICAO 9303~\cite{ICAO9303} compliant image, knowing that the sample with unknown image quality can only cause low comparison scores. The performance of FaceQnet v1 has been benchmarked against other FQAAs and proven competitive in the ongoing quality assessment evaluation of the National Institute of Standards and Quality (NIST)\cite{Grother-FRVT-Identification-NISTIR-2018}.
    
    \item \textbf{SER-FIQ}~\cite{Terhorst-SERFIQ-CVPR-2020} is an unsupervised technique that is not dependent on previously extracted ground truths for training a prediction model. Compared to FaceQnet v1, which outputs the general utility of a face image, SER-FIQ focuses on predicting the utility for a specific face recognition system. More precisely, the quality scores are based on the variations of face embeddings stemming from random subnetworks of a face recognition model. The authors argue that a high variation between the embeddings of the same sample functions as a robustness indication, which is assumed to be synonymous with image quality. The computational complexity of SER-FIQ increases quadratically with the number of random subnetworks, which leads to a trade-off between the efficiency of the algorithm and the expected accuracy of the quality predictions. In this work, we are following the authors' recommendation, choosing $N=100$ stochastic embeddings. The comparison of the authors against state-of-the-art FQAA approaches indicates that SER-FIQ significantly outperformed alternative methods.
    
\end{itemize}

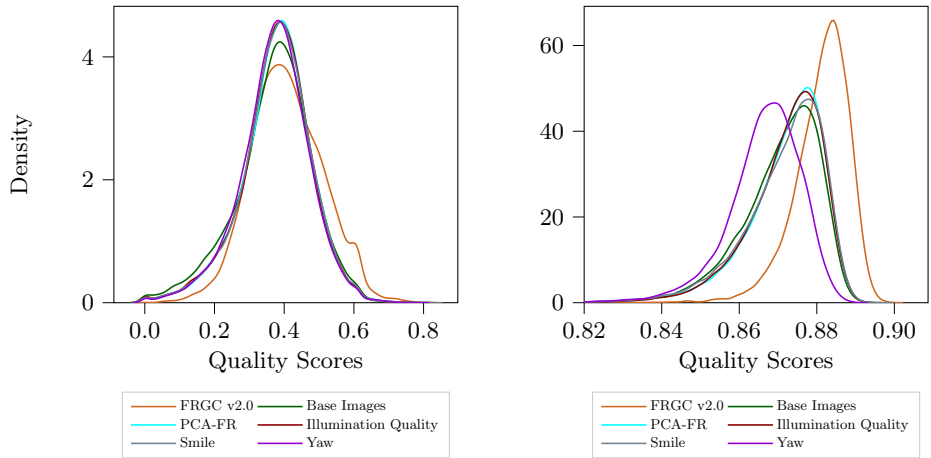
\begin{figure*}[htbp]
\centering
\subfigure{
    \begin{minipage}[t]{0.5\linewidth}
    \flushleft
    \input{images/faceqnetv1-nisk-qscores}
    \end{minipage}%
}%
\subfigure{
    \begin{minipage}[t]{0.5\linewidth}
    \flushright
    \input{images/serfiq-nisk-qscores}
    \end{minipage}%
}%
\caption{Quality score distributions of two FQAAs: FaceQnet v1 (left) and SER-FIQ (right).}
\label{fig:qscore-distributions}
\end{figure*}

The distributions of the quality scores predicted with FaceQnet v1 and SER-FIQ are shown in Figure~\ref{fig:qscore-distributions}. On the left, the well-aligned curves indicate that the average biometric quality across the evaluated datasets is nearly identical. However, looking at the SER-FIQ quality scores reveals a discrepancy between the distributions of the synthetic and bona fide images. We explain this observation with a wider range of yaw angles present in the synthetic datasets, a factor known by the authors of SER-FIQ to decrease the utility estimations~\cite{Terhorst-SERFIQ-CVPR-2020}. The same behaviour is reflected by the left-shifted purple curve, thereby validating the negative impact of yaw angle variations on the biometric quality. Overall, the analysis of the utility scores does not reveal significant differences between bona fide and synthetic images. Moreover, except for yaw angle manipulations, these differences even vanish when comparing only synthetic datasets. Hence, the biometric quality of mated samples generated with PCA-FR-Guided sampling and InterFaceGAN are similar as both are products of the same generator. Further, the generation of mated samples has not deteriorated the biometric quality, as indicated by the overlapping areas to the base image distributions.

\begin{figure*}[htbp]
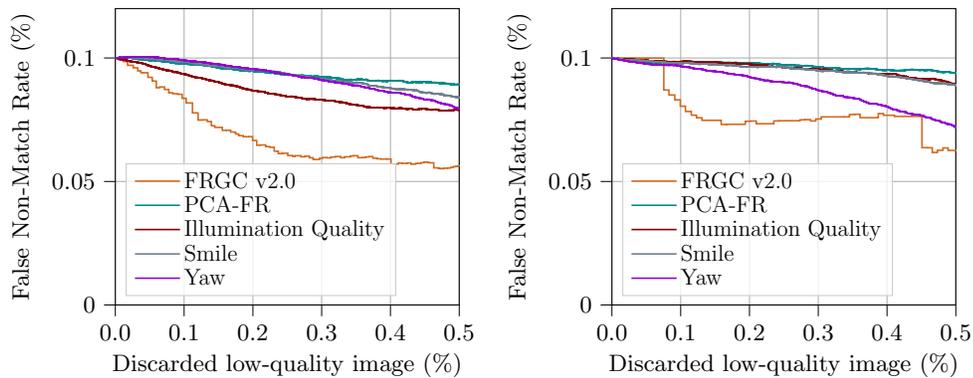

\centering
\subfigure{
    \begin{minipage}[t]{0.45\textwidth}
    \input{images/faceqnetv1-edc-nisk-paper}
    \end{minipage}%
}%
\hfill
\subfigure{
    \begin{minipage}[t]{0.45\textwidth}
    \input{images/serfiq-edc-nisk-paper}
    \end{minipage}%
}%
\caption{EDC curves based on paired quality scores derived with FaceQnet v1 (left) and SER-FIQ (right). False non-match rates are computed with ArcFace \cite{Deng-ArcFace-CVPR-2019}.}
\label{fig:edc-curves}
\end{figure*}

To further investigate the credibility of the FQAAs, Error-vs-Discard Characteristic (EDC) curves are shown in Figure~\ref{fig:edc-curves}. EDCs are commonly used to compare the performance of multiple FQAAs as suggested by the third version of ISO/IEC WD 29794-1:2021~\cite{ISO_29794_1:2021}. For each face comparison, a paired quality score is defined as the minimum of the single quality scores predicted with the FQAAs. Finally, EDC curves are obtained by measuring the FNMRs by increasingly discarding the lowest quality images from the test set. Hence, decreasing EDC curves indicate lower misclassification rates; thus, the underlying FQAA could predict the biometric quality.

In Figure~\ref{fig:edc-curves}, all EDC curves share the same decreasing trend. However, the orange curves (FRGC v2.0) are steeper, indicating that both FQAAs are more accurate in predicting the biometric quality of bona fide images than synthetic samples. One reason for this observation might be rooted in an increased intra-identity variation of the bona fide images, which is still challenging to mimic with synthetic substitutes. This assumption fits with the analysis of the mated
comparison scores presented in the following subsection.

\begin{figure*}[htbp]
\centering
\subfigure{
    \begin{minipage}{0.49\linewidth}
    \flushleft
    \input{images/comparison-frgc}
    \end{minipage}%
}%
\subfigure{
    \begin{minipage}{0.49\linewidth}
    \flushright
    \input{images/comparison-frpca}
    \end{minipage}%
}%

\subfigure{
    \begin{minipage}{0.49\linewidth}
    \flushleft
    \input{images/comparison-base}
    \end{minipage}%
}%
\subfigure{
    \begin{minipage}{0.49\linewidth}
    \flushright
    \input{images/comparison-illumination-quality}
    \end{minipage}%
}%

\subfigure{
    \begin{minipage}{0.49\linewidth}
    \flushleft
    \input{images/comparison-yaw}
    \end{minipage}%
}%
\subfigure{
    \begin{minipage}{0.49\linewidth}
    \flushright
    \input{images/comparison-smile}
    \end{minipage}%
}%
\caption{Mated and non-mated comparison scores computed with VGGFace2 \cite{Qiong-VGGFace2-arxiv-2018}. Thick solid line represents the kernel density curve of mated comparison scores, Thick dotted line represents the non-mated comparison scores and black dashed line represents the threshold $@$ FMR$=0.1\%$ on LFW \cite{LFWTech} dataset. Note that our base image dataset includes a single image per identity, therefore only depicting the non-mated comparison score distribution.}
\label{fig:comparison_scores}
\end{figure*}
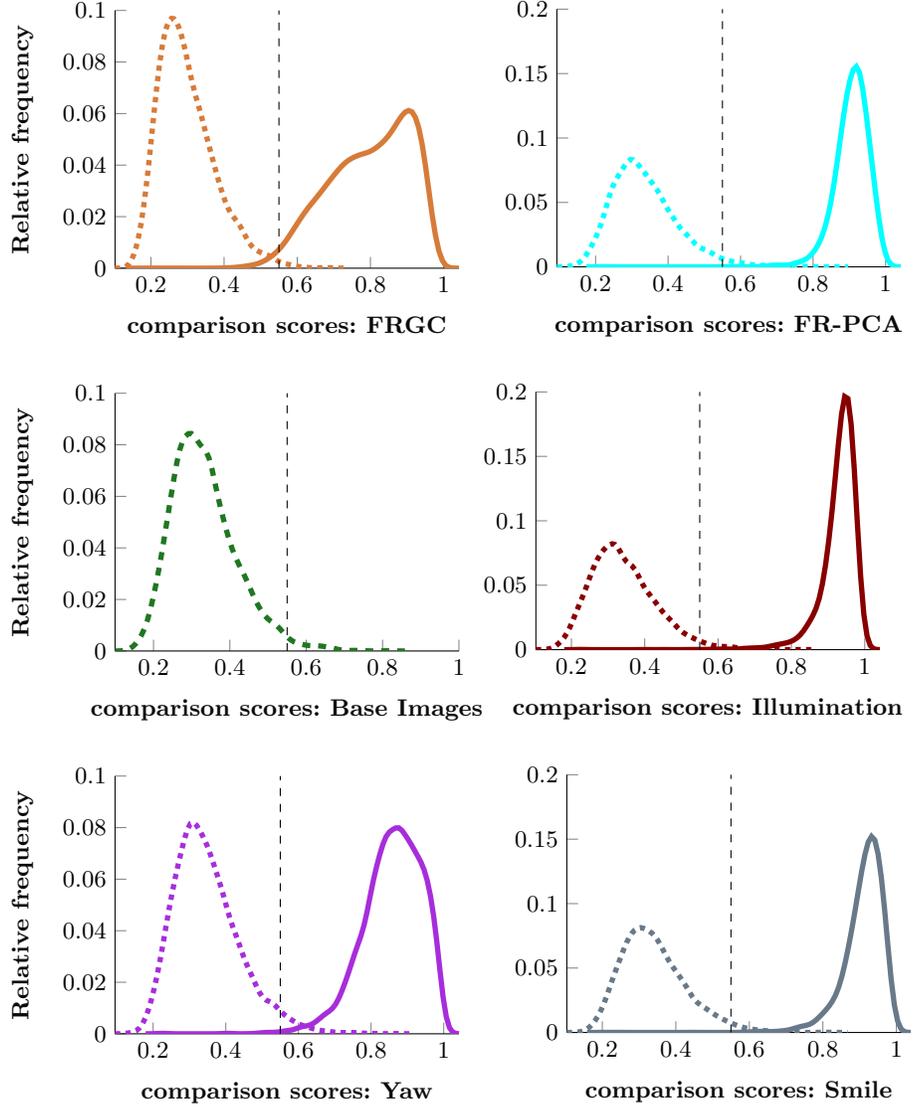

\subsection{Biometric Performance}

After evaluating the biometric quality with dedicated FQAAs, the synthetic face images are further assessed in Figure~\ref{fig:comparison_scores}, visualising the comparison score distributions. The non-mated comparison score distributions of all synthetic datasets are clearly below the vertically marked threshold, therefore indicating that the proportion of non-mated lookalikes is not significantly increased compared to non-mated bona fide samples. However, the mated comparison scores reveal more significant differences between the datasets. It is visible that the thick orange curve (FRGC v2.0) is heavier tailed on the left side than all synthetic mated distributions. Again, this observation re-validates the findings of the last subsection, tracing back the differences to a lower similarity of mated samples caused by varying facial attributes. Moreover, the mated comparison scores of the synthetic datasets reveal minor differences: While our proposed PCA-FR-Guided sampling performs similar to the controlled manipulation of smiles, editing yaw angles widens the span of mated comparison scores significantly. Overall, the well-separated distributions of mated and non-mated comparison scores lead us to conclude that either editing single (InterFaceGAN) or multiple (PCA-FR-Guided sampling) semantics can be promising for generating synthetic datasets for biometric performance tests. In addition, a quantitative analysis, measuring the Kullback-Leibler Divergences between the distributions presented in this section, is provided in Table \ref{tab:qscores-kl-divergences} and Table \ref{tab:cscores-kl-divergences} (appendix).


%% file: images/faceqnetv1-nisk-qscores.tex
\begin{tikzpicture}

\tikzset{lw/.style = {line width=4pt}}

\definecolor{color0}{rgb}{0.823529411764706,0.411764705882353,0.117647058823529}
\definecolor{color1}{rgb}{0,1,1}
\definecolor{color2}{rgb}{0.43921568627451,0.501960784313725,0.564705882352941}
\definecolor{color3}{rgb}{0.580392156862745,0,0.827450980392157}

\begin{axis}[
legend cell align={left},
legend style={
  fill opacity=0.8,
  draw opacity=1,
  text opacity=1,
  at={(0.5,-0.3)},
  anchor=north,
  nodes={scale=0.6, transform shape},
  legend columns=2,
  draw=white!80!black
},
tick align=outside,
tick pos=left,
scale only axis,
x grid style={white!69.0196078431373!black},
xlabel={Quality Scores},
width=1.8in,
xmin=-0.0886777652301044, xmax=0.898193425230104,
xtick style={color=black},
xtick={-0.2,0,0.2,0.4,0.6,0.8,1},
xticklabels={−0.2,0.0,0.2,0.4,0.6,0.8,1.0},
y grid style={white!69.0196078431373!black},
ylabel={Density},
ymin=0, ymax=4.82310947101849,
ytick style={color=black}
]
\addplot [semithick, color0]
table {%
-0.0438199838455494 2.48380927903217e-05
-0.0393116641084082 6.12711437595165e-05
-0.034803344371267 0.000137589229035631
-0.0302950246341258 0.00028127685584717
-0.0257867048969846 0.000523545366266105
-0.0212783851598434 0.000887438790456251
-0.0167700654227022 0.00137048960741073
-0.012261745685561 0.00193005284545261
-0.00775342594841983 0.00248365461797317
-0.00324510621127863 0.00293316586014396
0.00126321352586257 0.00320896549032762
0.00577153326300377 0.00331516467006772
0.010279853000145 0.00335089823895657
0.0147881727372862 0.0034925554739992
0.0192964924744274 0.00394289564660725
0.0238048122115686 0.00487148762723879
0.0283131319487098 0.00637418724061134
0.032821451685851 0.00846506503675846
0.0373297714229922 0.0110920409628501
0.0418380911601334 0.0141534728191334
0.0463464108972746 0.0174987414148517
0.0508547306344158 0.0209184724182023
0.055363050371557 0.024150658256718
0.0598713701086982 0.0269270274796548
0.0643796898458394 0.0290571224206699
0.0688880095829806 0.0305163791040906
0.0733963293201218 0.0314941199846703
0.077904649057263 0.0323736074977827
0.0824129687944042 0.033644195799702
0.0869212885315454 0.0357699224503083
0.0914296082686866 0.0390580504319025
0.0959379280058278 0.0435855151567458
0.100446247742969 0.0492351489670336
0.10495456748011 0.0558446880581637
0.109462887217251 0.0633888769491437
0.113971206954393 0.0720583191991952
0.118479526691534 0.0821385681775526
0.122987846428675 0.0937328716790284
0.127496166165816 0.106515525149645
0.132004485902957 0.119721448212825
0.136512805640099 0.132432145150897
0.14102112537724 0.144011492940291
0.145529445114381 0.154439096518722
0.150037764851522 0.164361273799211
0.154546084588663 0.174858804225272
0.159054404325805 0.187074069255129
0.163562724062946 0.201865190135331
0.168071043800087 0.219590520246268
0.172579363537228 0.240055632115235
0.177087683274369 0.26262222718666
0.181596003011511 0.28646575202275
0.186104322748652 0.310937818680311
0.190612642485793 0.335935745384017
0.195120962222934 0.362140884407784
0.199629281960075 0.391004322783667
0.204137601697217 0.424446843180324
0.208645921434358 0.464365728714559
0.213154241171499 0.512136089230368
0.21766256090864 0.56829803246794
0.222170880645781 0.632525556985568
0.226679200382923 0.703836448404116
0.231187520120064 0.780908057976257
0.235695839857205 0.862359884143288
0.240204159594346 0.946937184365557
0.244712479331487 1.03362667785193
0.249220799068629 1.12179405305437
0.25372911880577 1.21140574708184
0.258237438542911 1.30327999941681
0.262745758280052 1.39917856474266
0.267254078017193 1.50153204871844
0.271762397754335 1.61276408284092
0.276270717491476 1.73445028983925
0.280779037228617 1.86669841847221
0.285287356965758 2.00802009350669
0.289795676702899 2.15566824570882
0.294303996440041 2.30618566326343
0.298812316177182 2.45591865899333
0.303320635914323 2.60142876384153
0.307828955651464 2.73987191143481
0.312337275388605 2.86937834001693
0.316845595125747 2.98932598914075
0.321353914862888 3.10032370743481
0.325862234600029 3.20379916711097
0.33037055433717 3.30127259927678
0.334878874074311 3.39357987017457
0.339387193811453 3.48039113231868
0.343895513548594 3.56028921082326
0.348403833285735 3.63141147296673
0.352912153022876 3.69232701644677
0.357420472760017 3.74264802612563
0.361928792497159 3.7830412651341
0.3664371122343 3.81472519404592
0.370945431971441 3.83886908378915
0.375453751708582 3.85627115820186
0.379962071445723 3.86735795326046
0.384470391182865 3.87227499397388
0.388978710920006 3.87088344222569
0.393487030657147 3.86272119521088
0.397995350394288 3.84710467515743
0.402503670131429 3.82339889598917
0.407011989868571 3.79126348942927
0.411520309605712 3.7506665139712
0.416028629342853 3.70167514759877
0.420536949079994 3.64425716098635
0.425045268817135 3.57834791434302
0.429553588554277 3.50424836735339
0.434061908291418 3.42316217417521
0.438570228028559 3.33750798745338
0.4430785477657 3.25067427052564
0.447586867502841 3.16616935159427
0.452095187239983 3.0865433127281
0.456603506977124 3.01271183213364
0.461111826714265 2.94411579236569
0.465620146451406 2.87957280716375
0.470128466188547 2.81816638480236
0.474636785925689 2.75952323459155
0.47914510566283 2.70335885299226
0.483653425399971 2.64876826682865
0.488161745137112 2.59390328672187
0.492670064874253 2.53631861617289
0.497178384611395 2.47374240906177
0.501686704348536 2.40477012098245
0.506195024085677 2.32913510535321
0.510703343822818 2.24756365531072
0.515211663559959 2.1614564670764
0.519719983297101 2.07260308194285
0.524228303034242 1.98293753265365
0.528736622771383 1.89421502482823
0.533244942508524 1.80757130859663
0.537753262245665 1.72313747463345
0.542261581982807 1.63999752157423
0.546769901719948 1.55664674175899
0.551278221457089 1.47181786689273
0.55578654119423 1.38531904536366
0.560294860931371 1.29852992941903
0.564803180668513 1.2143930086757
0.569311500405654 1.13696154938987
0.573819820142795 1.07068595646284
0.578328139879936 1.01958302963377
0.582836459617077 0.986283960896887
0.587344779354219 0.970865925560749
0.59185309909136 0.969565392632643
0.596361418828501 0.973982984366862
0.600869738565642 0.971815710739785
0.605378058302783 0.94982986227458
0.609886378039925 0.898488578700152
0.614394697777066 0.81613764756866
0.618903017514207 0.710280314756275
0.623411337251348 0.594900360757495
0.627919656988489 0.485147034267622
0.632427976725631 0.392198277159525
0.636936296462772 0.320690498014408
0.641444616199913 0.2692753923864
0.645952935937054 0.233133148309479
0.650461255674195 0.206714534373953
0.654969575411337 0.185551226706525
0.659477895148478 0.16689219779237
0.663986214885619 0.1495231494601
0.66849453462276 0.133249554154434
0.673002854359901 0.118377685940707
0.677511174097043 0.105343288575366
0.682019493834184 0.0945081503705894
0.686527813571325 0.0860729097843142
0.691036133308466 0.0800351475236902
0.695544453045607 0.0761536162796744
0.700052772782749 0.0739331673866241
0.70456109251989 0.0726710721501673
0.709069412257031 0.0715806167830949
0.713577731994172 0.0699579578978328
0.718086051731313 0.0673287483459512
0.722594371468455 0.0635226754272442
0.727102691205596 0.0586612582879069
0.731611010942737 0.0530783876071856
0.736119330679878 0.0472079935638337
0.740627650417019 0.0414710446335555
0.745135970154161 0.036185277461565
0.749644289891302 0.0315131579005455
0.754152609628443 0.0274569415037229
0.758660929365584 0.0239002947811688
0.763169249102725 0.0206816453735843
0.767677568839867 0.0176704298250742
0.772185888577008 0.0148146701721506
0.776694208314149 0.0121433731464891
0.78120252805129 0.00973233363502288
0.785710847788431 0.00765887843719315
0.790219167525573 0.00596849423216158
0.794727487262714 0.00466092508139703
0.799235806999855 0.0036914794214271
0.803744126736996 0.00298240073683016
0.808252446474137 0.00244238720807884
0.812760766211279 0.00199074850877857
0.81726908594842 0.00157735791069591
0.821777405685561 0.00118875904949479
0.826285725422702 0.000838150185029141
0.830794045159843 0.000546718395875674
0.835302364896985 0.000327612916887947
0.839810684634126 0.00017957975899109
0.844319004371267 8.98149173587953e-05
0.848827324108408 4.09245663416767e-05
0.853335643845549 1.69739629978881e-05
};
\addlegendentry{FRGC v2.0}
\addplot [semithick, green!39.2156862745098!black]
table {%
-0.0423578736638996 0.000877443932965963
-0.0380481689034584 0.00211630867466041
-0.0337384641430172 0.00466028116927764
-0.029428759382576 0.00937526980645018
-0.0251190546221348 0.017244825206636
-0.0208093498616936 0.0290367289543705
-0.0164996451012524 0.044830799549539
-0.0121899403408112 0.063619882155552
-0.00788023558036995 0.083277049440501
-0.00357053081992875 0.101067938286256
0.000739173940512448 0.114581501083506
0.00504887870095365 0.122643584813887
0.00935858346139486 0.125717432045787
0.0136682882218361 0.125571962630434
0.0179779929822773 0.124430083779476
0.0222876977427185 0.124085098242196
0.0265974025031597 0.125419059940953
0.0309071072636009 0.128459197064843
0.0352168120240421 0.132802621311773
0.0395265167844833 0.138104383063425
0.0438362215449245 0.144379604025816
0.0481459263053657 0.152024465603169
0.0524556310658069 0.161608291234065
0.0567653358262481 0.173571351082307
0.0610750405866893 0.187971083976123
0.0653847453471305 0.204372954188748
0.0696944501075717 0.221912572151558
0.0740041548680129 0.239494249869317
0.0783138596284541 0.256057720559686
0.0826235643888953 0.270839823702566
0.0869332691493365 0.28356691112338
0.0912429739097777 0.294525022615573
0.0955526786702189 0.304475356138137
0.0998623834306601 0.31442792099491
0.104172088191101 0.325350677685254
0.108481792951542 0.337933382174267
0.112791497711984 0.352499319858685
0.117101202472425 0.369068641829056
0.121410907232866 0.387491158649935
0.125720611993307 0.407556500501587
0.130030316753748 0.429057859998928
0.13434002151419 0.451855060761292
0.138649726274631 0.475972953309399
0.142959431035072 0.501685514414125
0.147269135795513 0.529468025538978
0.151578840555954 0.559744637769872
0.155888545316396 0.592512721870584
0.160198250076837 0.627071504979744
0.164507954837278 0.662083727457079
0.168817659597719 0.696021303225167
0.173127364358161 0.727799671338429
0.177437069118602 0.757265662710744
0.181746773879043 0.785273465815775
0.186056478639484 0.813315614201807
0.190366183399925 0.842915158892917
0.194675888160367 0.875086739095547
0.198985592920808 0.910099820871671
0.203295297681249 0.94759776217386
0.20760500244169 0.98695193450359
0.211914707202131 1.02763986598734
0.216224411962572 1.06945340011328
0.220534116723014 1.1124485910219
0.224843821483455 1.15670011172443
0.229153526243896 1.20205275242096
0.233463231004337 1.24808845518769
0.237772935764779 1.2943978876018
0.24208264052522 1.34101363716487
0.246392345285661 1.38869550207279
0.250702050046102 1.43882015193683
0.255011754806543 1.49290501236213
0.259321459566984 1.55206965553951
0.263631164327426 1.61676830650105
0.267940869087867 1.68689368790299
0.272250573848308 1.76208051078627
0.276560278608749 1.84197018970493
0.280869983369191 1.92635480533919
0.285179688129632 2.01528842889418
0.289489392890073 2.109236249745
0.293799097650514 2.20914981418572
0.298108802410955 2.31623157625713
0.302418507171397 2.43127312135244
0.306728211931838 2.5537805495871
0.311037916692279 2.68139114670604
0.31534762145272 2.81007999652743
0.319657326213161 2.93528638847366
0.323967030973603 3.05354668743693
0.328276735734044 3.16384257068504
0.332586440494485 3.26793291863632
0.336896145254926 3.36943799215047
0.341205850015367 3.47209515336024
0.345515554775809 3.57801347074855
0.34982525953625 3.68669600425224
0.354134964296691 3.79515297251344
0.358444669057132 3.89889109549751
0.362754373817573 3.99323073515157
0.367064078578015 4.07439761855444
0.371373783338456 4.14007675707559
0.375683488098897 4.18941057995824
0.379993192859338 4.22261931676202
0.384302897619779 4.24048905924654
0.388612602380221 4.24395843314961
0.392922307140662 4.23396239356305
0.397232011901103 4.21154307430248
0.401541716661544 4.1780420014484
0.405851421421985 4.13508328073545
0.410161126182427 4.08418768527061
0.414470830942868 4.02618131238243
0.418780535703309 3.96081471765969
0.42309024046375 3.88692986472911
0.427399945224191 3.80313253745259
0.431709649984633 3.70857284955917
0.436019354745074 3.6034122813454
0.440329059505515 3.48884211599815
0.444638764265956 3.36681182380389
0.448948469026397 3.23967840417337
0.453258173786839 3.10985054007023
0.45756787854728 2.97941356687619
0.461877583307721 2.84979927332415
0.466187288068162 2.72167286337116
0.470496992828603 2.59515006335022
0.474806697589045 2.47023200239297
0.479116402349486 2.34716964789386
0.483426107109927 2.22652859893632
0.487735811870368 2.10896627467447
0.492045516630809 1.9949314361011
0.496355221391251 1.88449664871824
0.500664926151692 1.7773911961902
0.504974630912133 1.67317391347307
0.509284335672574 1.57145491625421
0.513594040433015 1.47210131883133
0.517903745193457 1.37537824185854
0.522213449953898 1.28197630423858
0.526523154714339 1.19289393107731
0.53083285947478 1.10918852611519
0.535142564235221 1.03166225843819
0.539452268995663 0.96058241776893
0.543761973756104 0.895543357040522
0.548071678516545 0.835546290269071
0.552381383276986 0.77929447378976
0.556691088037428 0.725597091462512
0.561000792797869 0.673711554876614
0.56531049755831 0.623489556934348
0.569620202318751 0.575310571906835
0.573929907079192 0.529895451698746
0.578239611839633 0.488105066318602
0.582549316600075 0.450745354309349
0.586859021360516 0.418314895620018
0.591168726120957 0.390653766750757
0.595478430881398 0.366604223784533
0.59978813564184 0.343956197484705
0.604097840402281 0.319931902617237
0.608407545162722 0.292175123367865
0.612717249923163 0.259810504760249
0.617026954683604 0.22395898846123
0.621336659444045 0.187351134420202
0.625646364204487 0.153225436012734
0.629956068964928 0.124122909596677
0.634265773725369 0.101180604675683
0.63857547848581 0.0841415341672769
0.642885183246252 0.0718702562327295
0.647194888006693 0.0629815778251128
0.651504592767134 0.0562873934384742
0.655814297527575 0.0509765177723536
0.660124002288016 0.0465936794499129
0.664433707048458 0.0429240796804805
0.668743411808899 0.0398635631452807
0.67305311656934 0.0373190406302708
0.677362821329781 0.0351603586202333
0.681672526090222 0.0332256727620527
0.685982230850664 0.0313610431913639
0.690291935611105 0.0294602651629257
0.694601640371546 0.0274766139941339
0.698911345131987 0.0254037188644286
0.703221049892428 0.0232480695480508
0.70753075465287 0.0210181193395177
0.711840459413311 0.0187333520229314
0.716150164173752 0.0164344224393232
0.720459868934193 0.014176162005816
0.724769573694634 0.0120067207594975
0.729079278455075 0.00995348645833738
0.733388983215517 0.0080305827277113
0.737698687975958 0.00626085898667516
0.742008392736399 0.00469071117476842
0.74631809749684 0.00338237357563109
0.750627802257282 0.002387380997963
0.754937507017723 0.00171874988662952
0.759247211778164 0.00133858310819193
0.763556916538605 0.00116672555885783
0.767866621299046 0.00110478214030459
0.772176326059487 0.00106422927607001
0.776486030819929 0.000987673019827897
0.78079573558037 0.000856467532745297
0.785105440340811 0.00068401663791159
0.789415145101252 0.000499976793180585
0.793724849861694 0.000333583520784848
0.798034554622135 0.00020293004776492
0.802344259382576 0.000112505140318697
0.806653964143017 5.68326114063085e-05
0.810963668903458 2.61569063326365e-05
0.8152733736639 1.096790322043e-05
};
\addlegendentry{Base Images}
\addplot [semithick, color1]
table {%
-0.0352781061515267 0.000792313421645662
-0.0309105228233706 0.00226178701126833
-0.0265429394952146 0.00563015513396697
-0.0221753561670585 0.012226891141224
-0.0178077728389025 0.0231843236587874
-0.0134401895107464 0.0384385419739954
-0.00907260618259038 0.055861541903026
-0.00470502285443433 0.0714791711312734
-0.000337439526278287 0.0811926363042902
0.00403014380187776 0.0830871698917287
0.0083977271300338 0.0785697686248817
0.0127653104581898 0.0713707130545347
0.0171328937863459 0.0652546587880413
0.0215004771145019 0.0622526281064926
0.025868060442658 0.0624611012931748
0.030235643770814 0.064998058844483
0.0346032270989701 0.069049701222162
0.0389708104271261 0.0743527466105209
0.0433383937552822 0.0811232546267782
0.0477059770834382 0.089711543708064
0.0520735604115943 0.100206138121824
0.0564411437397503 0.112154367379849
0.0608087270679064 0.124557899962551
0.0651763103960624 0.136186467596156
0.0695438937242185 0.146049281764465
0.0739114770523745 0.153768541506977
0.0782790603805305 0.159689759382009
0.0826466437086866 0.16471035181263
0.0870142270368426 0.169905704554919
0.0913818103649987 0.176109179670593
0.0957493936931547 0.183658630857066
0.100116977021311 0.192455245665971
0.104484560349467 0.202273426683073
0.108852143677623 0.213083555501593
0.113219727005779 0.225169959059133
0.117587310333935 0.239005478847074
0.121954893662091 0.255005571724697
0.126322476990247 0.273316328749459
0.130690060318403 0.29371907383272
0.135057643646559 0.315653378474876
0.139425226974715 0.338335557582263
0.143792810302871 0.360973939730723
0.148160393631027 0.383087910988772
0.152527976959183 0.404853716229153
0.156895560287339 0.427253666936203
0.161263143615495 0.451776461971192
0.165630726943651 0.479671003802257
0.169998310271808 0.511164934034499
0.174365893599964 0.545221850533412
0.17873347692812 0.580086425539528
0.183101060256276 0.614289044333655
0.187468643584432 0.647447270102521
0.191836226912588 0.680381571545795
0.196203810240744 0.714573517683549
0.2005713935689 0.751397204819595
0.204938976897056 0.791557538171716
0.209306560225212 0.8348820837786
0.213674143553368 0.880402963791672
0.218041726881524 0.926705529718554
0.22240931020968 0.972566251653459
0.226776893537836 1.01767206212235
0.231144476865992 1.06291562904811
0.235512060194148 1.10992173766966
0.239879643522304 1.16014220386714
0.24424722685046 1.21436766291884
0.248614810178616 1.27315736177806
0.252982393506772 1.3377207186584
0.257349976834928 1.41020618317268
0.261717560163085 1.4928299810759
0.266085143491241 1.58638061332453
0.270452726819397 1.68929269285779
0.274820310147553 1.79807483301253
0.279187893475709 1.90880637836298
0.283555476803865 2.01866359673477
0.287923060132021 2.1265849926232
0.292290643460177 2.23295504334869
0.296658226788333 2.33884585444871
0.301025810116489 2.4454685053353
0.305393393444645 2.5541409246845
0.309760976772801 2.6666063774878
0.314128560100957 2.78523476477856
0.318496143429113 2.91269258317409
0.322863726757269 3.05103364010124
0.327231310085425 3.20055559245089
0.331598893413581 3.35894859967553
0.335966476741737 3.52122663865921
0.340334060069893 3.68067577218673
0.344701643398049 3.83052183029365
0.349069226726205 3.96551856777068
0.353436810054361 4.08278025321555
0.357804393382518 4.18190188049191
0.362171976710674 4.26481397431718
0.36653956003883 4.33535778362969
0.370907143366986 4.398039235696
0.375274726695142 4.45591413140338
0.379642310023298 4.50872247054175
0.384009893351454 4.55271417093041
0.38837747667961 4.5824577206701
0.392745060007766 4.5934379650345
0.397112643335922 4.5838791853874
0.401480226664078 4.55504619836941
0.405847809992234 4.51023523482137
0.41021539332039 4.45308209886292
0.414582976648546 4.38589810824897
0.418950559976702 4.30878577239494
0.423318143304858 4.22003299556709
0.427685726633014 4.11750665575843
0.43205330996117 4.00001954723974
0.436420893289326 3.8678042060657
0.440788476617482 3.7222303205465
0.445156059945638 3.56558502950389
0.449523643273794 3.4012769886032
0.453891226601951 3.23388569540712
0.458258809930107 3.0683533589997
0.462626393258263 2.90853496057604
0.466993976586419 2.75618366489565
0.471361559914575 2.61122125100447
0.475729143242731 2.47301973486137
0.480096726570887 2.34151358485733
0.484464309899043 2.21714611747863
0.488831893227199 2.09976837430463
0.493199476555355 1.9876374849146
0.497567059883511 1.87765874134377
0.501934643211667 1.76695607570772
0.506302226539823 1.65465163292418
0.510669809867979 1.54244031974808
0.515037393196135 1.43345153864173
0.519404976524291 1.33025401450052
0.523772559852447 1.23350684439982
0.528140143180603 1.1421309864534
0.532507726508759 1.05457809449171
0.536875309836915 0.970017653047584
0.541242893165071 0.888647779592803
0.545610476493227 0.811277048534028
0.549978059821383 0.738807471948297
0.554345643149539 0.671981070233432
0.558713226477696 0.61130415423341
0.563080809805852 0.556968231714922
0.567448393134008 0.508763686517775
0.571815976462164 0.466101714604152
0.57618355979032 0.428224487909833
0.580551143118476 0.394565170977594
0.584918726446632 0.36505220622789
0.589286309774788 0.340016489342068
0.593653893102944 0.319481018982677
0.5980214764311 0.302127524176114
0.602389059759256 0.284829672317861
0.606756643087412 0.263606005702246
0.611124226415568 0.235802814767413
0.615491809743724 0.202008323852169
0.61985939307188 0.16605901154208
0.624226976400036 0.132983479720623
0.628594559728192 0.10645072622725
0.632962143056348 0.0874952347996903
0.637329726384504 0.0749502771174309
0.641697309712661 0.0666998199138585
0.646064893040817 0.0607165927986156
0.650432476368973 0.0555413347922826
0.654800059697129 0.050395125807408
0.659167643025285 0.0451276152295853
0.663535226353441 0.0400296394345298
0.667902809681597 0.0355343162648902
0.672270393009753 0.031941811318506
0.676637976337909 0.0292988628901855
0.681005559666065 0.0274246142222669
0.685373142994221 0.0259869432891223
0.689740726322377 0.0245895949586361
0.694108309650533 0.0229011518894636
0.698475892978689 0.0208091568538269
0.702843476306845 0.018486791111541
0.707211059635001 0.0162810983472243
0.711578642963157 0.014482216227735
0.715946226291313 0.0131454756969897
0.720313809619469 0.0120883766889873
0.724681392947625 0.0110361826239071
0.729048976275781 0.00979210133528521
0.733416559603937 0.0083242644711588
0.737784142932093 0.00674324644659365
0.74215172626025 0.00521540095710577
0.746519309588406 0.00387959948854413
0.750886892916562 0.00280987609260297
0.755254476244718 0.00202244260335438
0.759622059572874 0.0014976450744939
0.76398964290103 0.00119261634190659
0.768357226229186 0.0010451660317306
0.772724809557342 0.000984414873853317
0.777092392885498 0.00095326338854321
0.781459976213654 0.000926740542192045
0.78582755954181 0.000906977075332133
0.790195142869966 0.000897484789988283
0.794562726198122 0.000882115597153466
0.798930309526278 0.000829988229860641
0.803297892854434 0.000720402540653114
0.80766547618259 0.00056191425222016
0.812033059510746 0.000387910298257229
0.816400642838902 0.000235108964256688
0.820768226167058 0.000124605494449455
0.825135809495214 5.7634567732679e-05
0.82950339282337 2.32431969509404e-05
0.833870976151527 8.16918525504265e-06
};
\addlegendentry{PCA-FR}
\addplot [semithick, red!54.5098039215686!black]
table {%
-0.035115102949783 0.000772239179917264
-0.0307476019151118 0.00223298603751339
-0.0263801008804406 0.00563330784411133
-0.0220125998457694 0.0124115254945414
-0.0176450988110983 0.0239182526694381
-0.0132775977764271 0.0404092737951118
-0.00891009674175588 0.0600676716076039
-0.00454259570708469 0.079006126642396
-0.000175094672413507 0.0927668755279004
0.00419240636225768 0.0985782656782231
0.00855990739692886 0.0967374535127866
0.0129274084316 0.0901027885848113
0.0172949094662712 0.0822959921445792
0.0216624105009424 0.07612202546854
0.0260299115356136 0.073088994354407
0.0303974125702848 0.0737097934315256
0.034764913604956 0.0778299862468158
0.0391324146396272 0.0847068675689771
0.0434999156742984 0.0931176347720425
0.0478674167089695 0.101764686461582
0.0522349177436407 0.109830329212848
0.0566024187783119 0.117260140251971
0.0609699198129831 0.124530432739024
0.0653374208476543 0.132101865089754
0.0697049218823255 0.140029754607477
0.0740724229169967 0.148017636018625
0.0784399239516679 0.155763520441354
0.082807424986339 0.163236246691787
0.0871749260210102 0.170726038204407
0.0915424270556814 0.178807529780356
0.0959099280903526 0.188337440733912
0.100277429125024 0.200368411064156
0.104644930159695 0.215814780692943
0.109012431194366 0.234979297415928
0.113379932229037 0.257282653007326
0.117747433263709 0.281404464520537
0.12211493429838 0.30570928656936
0.126482435333051 0.328693924036341
0.130849936367722 0.349329267798359
0.135217437402393 0.367299105987447
0.139584938437064 0.383090175269147
0.143952439471736 0.397824197415866
0.148319940506407 0.412834266811034
0.152687441541078 0.429196279540521
0.157054942575749 0.447491659600424
0.16142244361042 0.467897604200105
0.165789944645092 0.490435050600774
0.170157445679763 0.515113444824695
0.174524946714434 0.541871660150409
0.178892447749105 0.570444852780063
0.183259948783776 0.600379164519101
0.187627449818447 0.631313062893944
0.191994950853119 0.66338925411122
0.19636245188779 0.69740663396136
0.200729952922461 0.734381474778051
0.205097453957132 0.774707055219944
0.209464954991803 0.817625577861047
0.213832456026475 0.861570397826603
0.218199957061146 0.905155567523769
0.222567458095817 0.948063749894221
0.226934959130488 0.991353036202074
0.231302460165159 1.0372417600042
0.235669961199831 1.08851016366527
0.240037462234502 1.14745862819996
0.244404963269173 1.21458017410418
0.248772464303844 1.28771407635312
0.253139965338515 1.36259262508391
0.257507466373186 1.43489659384021
0.261874967407858 1.50280556389394
0.266242468442529 1.56852795932744
0.2706099694772 1.63776697659163
0.274977470511871 1.7172307726897
0.279344971546542 1.81155548322875
0.283712472581214 1.92155315968321
0.288079973615885 2.04483196334415
0.292447474650556 2.17799884861611
0.296814975685227 2.31848418242298
0.301182476719898 2.46470710661998
0.30554997775457 2.61508218932384
0.309917478789241 2.76735523565299
0.314284979823912 2.91905612979785
0.318652480858583 3.06846730464758
0.323019981893254 3.21500835032979
0.327387482927925 3.3586949642051
0.331754983962597 3.49929699632448
0.336122484997268 3.63595922055569
0.340489986031939 3.76745333213184
0.34485748706661 3.89267617010223
0.349224988101281 4.01090919011098
0.353592489135953 4.12158762056567
0.357959990170624 4.22372048295184
0.362327491205295 4.31549119449339
0.366694992239966 4.3945832516954
0.371062493274637 4.45916419259675
0.375429994309309 4.50868559601178
0.37979749534398 4.54365891321978
0.384164996378651 4.56459639917946
0.388532497413322 4.5712932747651
0.392899998447993 4.56330451702599
0.397267499482664 4.54107430316853
0.401635000517336 4.50633439938033
0.406002501552007 4.46112055110023
0.410370002586678 4.40624195148318
0.414737503621349 4.34066946176042
0.41910500465602 4.26248289068955
0.423472505690692 4.17059485398773
0.427840006725363 4.06577909033654
0.432207507760034 3.95017133259255
0.436575008794705 3.82576899219281
0.440942509829376 3.69328359907937
0.445310010864047 3.55230928292094
0.449677511898719 3.40265097670023
0.45404501293339 3.2458020298502
0.458412513968061 3.08547465275032
0.462780015002732 2.92666314711929
0.467147516037403 2.77363657436617
0.471515017072075 2.62802671660836
0.475882518106746 2.488225871776
0.480250019141417 2.35052006036129
0.484617520176088 2.21133446582262
0.488985021210759 2.06935388977428
0.493352522245431 1.92636868169914
0.497720023280102 1.78641490362974
0.502087524314773 1.65381061477407
0.506455025349444 1.53136356779301
0.510822526384115 1.41974364877211
0.515190027418786 1.31804247864543
0.519557528453458 1.22481695570544
0.523925029488129 1.13891772278292
0.5282925305228 1.0597781196375
0.532660031557471 0.987146987497356
0.537027532592142 0.920500116606234
0.541395033626814 0.858609077213188
0.545762534661485 0.799716232974362
0.550130035696156 0.742284272091084
0.554497536730827 0.685724224046806
0.558865037765498 0.630464512962224
0.563232538800169 0.577325591245898
0.567600039834841 0.526823741280192
0.571967540869512 0.479061530495768
0.576335041904183 0.434254336202044
0.580702542938854 0.393334718113894
0.585070043973525 0.358009465585233
0.589437545008197 0.330032414445601
0.593805046042868 0.309919171121824
0.598172547077539 0.295731767167609
0.60254004811221 0.282864709010758
0.606907549146881 0.265561095332443
0.611275050181552 0.239771738373411
0.615642551216224 0.205578155948546
0.620010052250895 0.167276751486055
0.624377553285566 0.130940384214782
0.628745054320237 0.101290098982652
0.633112555354909 0.0800055665350998
0.63748005638958 0.0660956798072825
0.641847557424251 0.0573712537103045
0.646215058458922 0.0517434629548162
0.650582559493593 0.0477675922950147
0.654950060528264 0.0445878461868138
0.659317561562936 0.0416885467417417
0.663685062597607 0.0387228513359907
0.668052563632278 0.0354727632273702
0.672420064666949 0.0318758754142785
0.67678756570162 0.0280463202289921
0.681155066736292 0.0242460660919175
0.685522567770963 0.0207882461646
0.689890068805634 0.0178985967185514
0.694257569840305 0.0156184661392855
0.698625070874976 0.0138304509039569
0.702992571909647 0.0123914061426683
0.707360072944319 0.0112544796577367
0.71172757397899 0.0104682253455883
0.716095075013661 0.0100582441351317
0.720462576048332 0.00991008959050549
0.724830077083003 0.00977447440706415
0.729197578117675 0.00940622989580208
0.733565079152346 0.00872493198393921
0.737932580187017 0.00785941984121836
0.742300081221688 0.00703718605443529
0.746667582256359 0.00641232830109088
0.751035083291031 0.00596874217732876
0.755402584325702 0.00555757954290062
0.759770085360373 0.00501655996138933
0.764137586395044 0.00427455194684895
0.768505087429715 0.00338278938831379
0.772872588464386 0.00247615036074081
0.777240089499058 0.00170275661964971
0.781607590533729 0.00116117313299668
0.7859750915684 0.000868802497154154
0.790342592603071 0.000767202199368769
0.794710093637742 0.000756294351488367
0.799077594672414 0.000740552739718375
0.803445095707085 0.00066609499316892
0.807812596741756 0.000531730444349414
0.812180097776427 0.000371910498169421
0.816547598811098 0.000226879712667477
0.820915099845769 0.000120522382785718
0.825282600880441 5.57198251401685e-05
0.829650101915112 2.24148079519876e-05
0.834017602949783 7.84535484135053e-06
};
\addlegendentry{Illumination Quality}
\addplot [semithick, color2]
table {%
-0.0350646451585482 0.000720421961713618
-0.0306126301318291 0.00211945925937168
-0.02616061510511 0.00540572912676435
-0.021708600078391 0.0119647056234383
-0.0172565850516719 0.0230160698084677
-0.0128045700249528 0.0385740486749229
-0.00835255499823373 0.0565468766186499
-0.00390053997151465 0.072982117887979
0.000551475055204427 0.0838461937953051
0.0050034900819235 0.0873143969401512
0.00945550510864258 0.0847910597911811
0.0139075201353617 0.0798448815410879
0.0183595351620807 0.0760356995616952
0.0228115501887998 0.0753330591265919
0.0272635652155189 0.0779402679431695
0.031715580242238 0.0830695498239248
0.036167595268957 0.0897821290936599
0.0406196102956761 0.0973863499813186
0.0450716253223952 0.105398038539503
0.0495236403491143 0.113334081896312
0.0539756553758334 0.120639059678098
0.0584276704025524 0.126887688993741
0.0628796854292715 0.132129724871397
0.0673317004559906 0.137046747277869
0.0717837154827097 0.142678183218433
0.0762357305094288 0.14982706077888
0.0806877455361478 0.158591452414079
0.0851397605628669 0.168444172780173
0.089591775589586 0.178825486853777
0.0940437906163051 0.189705830169634
0.0984958056430241 0.201574345789892
0.102947820669743 0.214886312768159
0.107399835696462 0.229544661022295
0.111851850723181 0.24490180498352
0.1163038657499 0.260218059027048
0.12075588077662 0.275162173152274
0.125207895803339 0.290046548365216
0.129659910830058 0.305725275119463
0.134111925856777 0.323189836438515
0.138563940883496 0.343008927680967
0.143015955910215 0.364944395925863
0.147467970936934 0.388066554698234
0.151919985963653 0.411327882415222
0.156372000990372 0.434152361188688
0.160824016017091 0.456609489945147
0.16527603104381 0.479188602404006
0.169728046070529 0.502598860810976
0.174180061097248 0.527899492397443
0.178632076123968 0.55667106123937
0.183084091150687 0.590579320878917
0.187536106177406 0.630187375112435
0.191988121204125 0.67386947147485
0.196440136230844 0.718000645244475
0.200892151257563 0.758664499448273
0.205344166284282 0.793792124429614
0.209796181311001 0.824264476023572
0.21424819633772 0.853379887597656
0.218700211364439 0.885264300127387
0.223152226391158 0.923275668880854
0.227604241417877 0.969112442474783
0.232056256444596 1.02274458466133
0.236508271471316 1.08292204307084
0.240960286498035 1.14793730252766
0.245412301524754 1.21637133966063
0.249864316551473 1.2876207850711
0.254316331578192 1.3620649497557
0.258768346604911 1.44081908071017
0.26322036163163 1.52517476300468
0.267672376658349 1.61601158411383
0.272124391685068 1.71352948431597
0.276576406711787 1.81746073095978
0.281028421738506 1.92753840532908
0.285480436765225 2.04374353390803
0.289932451791944 2.16603575033959
0.294384466818664 2.29379389760714
0.298836481845383 2.42553655634684
0.303288496872102 2.55928004857931
0.307740511898821 2.69335995646383
0.31219252692554 2.82725530889238
0.316644541952259 2.96206200815589
0.321096556978978 3.10035139711321
0.325548572005697 3.24504489623588
0.330000587032416 3.3972376923064
0.334452602059135 3.55402597938517
0.338904617085854 3.70827907501064
0.343356632112573 3.85133521617438
0.347808647139292 3.97710289744424
0.352260662166012 4.08456576032253
0.356712677192731 4.17700323839313
0.36116469221945 4.25900991613532
0.365616707246169 4.33376325832029
0.370068722272888 4.40197565126618
0.374520737299607 4.46223777890701
0.378972752326326 4.51185246566139
0.383424767353045 4.54768889539712
0.387876782379764 4.56690822786761
0.392328797406483 4.56737085481697
0.396780812433202 4.54774596944045
0.401232827459921 4.50775174415238
0.40568484248664 4.44874774505935
0.41013685751336 4.37407123581192
0.414588872540079 4.28824620640849
0.419040887566798 4.19518674419549
0.423492902593517 4.0967083307927
0.427944917620236 3.99250286505094
0.432396932646955 3.88131286624207
0.436848947673674 3.76205622799148
0.441300962700393 3.63416964270564
0.445752977727112 3.49756337828504
0.450204992753831 3.35285211251453
0.45465700778055 3.20184024955681
0.459109022807269 3.04765227873023
0.463561037833988 2.89408265498282
0.468013052860708 2.74440280790152
0.472465067887427 2.60036278429845
0.476917082914146 2.46203999551516
0.481369097940865 2.32854063098349
0.485821112967584 2.19891082220933
0.490273127994303 2.07261220813795
0.494725143021022 1.94948415241803
0.499177158047741 1.82953224975301
0.50362917307446 1.71280365691915
0.508081188101179 1.59939029905005
0.512533203127898 1.48953027899005
0.516985218154617 1.38371476281597
0.521437233181337 1.28261403477641
0.525889248208055 1.18675217125553
0.530341263234775 1.09616554455093
0.534793278261494 1.01038165976313
0.539245293288213 0.928773570071109
0.543697308314932 0.850988594634377
0.548149323341651 0.777091900268936
0.55260133836837 0.707354958346145
0.557053353395089 0.641987828752529
0.561505368421808 0.58117949180383
0.565957383448527 0.525426117351237
0.570409398475246 0.475685817227421
0.574861413501965 0.433000049955253
0.579313428528685 0.3978503691291
0.583765443555404 0.369888881362861
0.588217458582123 0.348215629753996
0.592669473608842 0.331590116772235
0.597121488635561 0.317956814036126
0.60157350366228 0.303714416494631
0.606025518688999 0.284094159343099
0.610477533715718 0.255360290336338
0.614929548742437 0.217525716805978
0.619381563769156 0.175119212849079
0.623833578795875 0.134950609244794
0.628285593822594 0.102508455067736
0.632737608849313 0.0796772726533732
0.637189623876033 0.064973432744979
0.641641638902752 0.0553934599950502
0.646093653929471 0.0482561167606066
0.65054566895619 0.0421069784128456
0.654997683982909 0.0366730380657878
0.659449699009628 0.032297889638811
0.663901714036347 0.0293122322078951
0.668353729063066 0.0276433620095631
0.672805744089785 0.0267688128874069
0.677257759116504 0.0259468016634788
0.681709774143223 0.0245611036516512
0.686161789169942 0.0224010563026716
0.690613804196661 0.0197377666077754
0.695065819223381 0.0171503797788503
0.6995178342501 0.0151877625705524
0.703969849276819 0.014059815925247
0.708421864303538 0.0135499638578227
0.712873879330257 0.0131962166654271
0.717325894356976 0.0125984917338311
0.721777909383695 0.0116316999970819
0.726229924410414 0.0104389552588323
0.730681939437133 0.0092536707605601
0.735133954463852 0.00821023620622555
0.739585969490571 0.00728157034523832
0.74403798451729 0.00636141158007544
0.748489999544009 0.00539015759040537
0.752942014570729 0.00440575242621438
0.757394029597448 0.00349471406691143
0.761846044624167 0.00271851992960953
0.766298059650886 0.00209264409520981
0.770750074677605 0.00161692215640736
0.775202089704324 0.00130123885182049
0.779654104731043 0.00115393624056162
0.784106119757762 0.00115418375482346
0.788558134784481 0.00124533942444329
0.7930101498112 0.00135738725443494
0.797462164837919 0.00143603485116849
0.801914179864638 0.0014550098258332
0.806366194891357 0.00140847976279908
0.810818209918077 0.00129768072033641
0.815270224944796 0.00112632812567593
0.819722239971515 0.000906679851441143
0.824174254998234 0.000665419878675457
0.828626270024953 0.000438480145791771
0.833078285051672 0.000256342932231932
0.837530300078391 0.000131817190404263
0.84198231510511 5.92718379898822e-05
0.846434330131829 2.32145161251309e-05
0.850886345158548 7.89963346569628e-06
};
\addlegendentry{Smile}
\addplot [semithick, color3]
table {%
-0.0343785599254906 0.00057169716146592
-0.0301005904790033 0.00164909991511698
-0.0258226210325159 0.00414821028080052
-0.0215446515860286 0.00910808860683782
-0.0172666821395412 0.017481691083092
-0.0129887126930539 0.0293978236403029
-0.00871074324656654 0.0434685186147583
-0.0044327738000792 0.0568380397734791
-0.000154804353591859 0.0663254244360373
0.00412316509289548 0.0700826960168803
0.00840113453938283 0.0685678145213769
0.0126791039858702 0.0641104843545043
0.0169570734323575 0.0595199018033782
0.0212350428788449 0.0568280127416245
0.0255130123253322 0.0568538592524226
0.0297909817718195 0.0594803797013099
0.0340689512183069 0.0641694265624378
0.0383469206647942 0.0703630224259685
0.0426248901112816 0.0776439400237798
0.0469028595577689 0.0856785409769662
0.0511808290042563 0.0940829560070275
0.0554587984507436 0.10240045366262
0.059736767897231 0.110256619701962
0.0640147373437183 0.117551473390405
0.0682927067902056 0.124484614791767
0.072570676236693 0.131379764261943
0.0768486456831803 0.138486878934365
0.0811266151296677 0.145942571115386
0.085404584576155 0.153866518723593
0.0896825540226423 0.162420152472653
0.0939605234691297 0.171736984757111
0.098238492915617 0.181833975775012
0.102516462362104 0.192674136072548
0.106794431808592 0.204387471761983
0.111072401255079 0.217441157087612
0.115350370701566 0.232525770733172
0.119628340148054 0.250163519099279
0.123906309594541 0.270331863814422
0.128184279041028 0.292418311577936
0.132462248487516 0.315535741163331
0.136740217934003 0.33893982981866
0.14101818738049 0.362275527373957
0.145296156826978 0.385567431108039
0.149574126273465 0.409033837069395
0.153852095719953 0.432867675997331
0.15813006516644 0.457112501965921
0.162408034612927 0.481664876502393
0.166686004059415 0.50632637860333
0.170963973505902 0.530863953052388
0.175241942952389 0.555178234683424
0.179519912398877 0.579642675671651
0.183797881845364 0.605383521554523
0.188075851291851 0.634108541644035
0.192353820738339 0.667417945820282
0.196631790184826 0.7060782223425
0.200909759631313 0.749860049348581
0.205187729077801 0.798022871540808
0.209465698524288 0.849908454505268
0.213743667970775 0.905064180477594
0.218021637417263 0.962894899264073
0.22229960686375 1.02239741113208
0.226577576310237 1.08245700405942
0.230855545756725 1.14256751597849
0.235133515203212 1.20333992800266
0.239411484649699 1.26630386085805
0.243689454096187 1.33312919956941
0.247967423542674 1.40489981897169
0.252245392989161 1.48201450504653
0.256523362435649 1.56473092988893
0.260801331882136 1.65375796740693
0.265079301328623 1.75014842672242
0.269357270775111 1.85427985143943
0.273635240221598 1.96464212303855
0.277913209668085 2.0776764901072
0.282191179114573 2.18930747312521
0.28646914856106 2.29733941539337
0.290747118007547 2.402899130299
0.295025087454035 2.50971656489058
0.299303056900522 2.62179283097563
0.30358102634701 2.7412634121647
0.307858995793497 2.86792007738044
0.312136965239984 3.00036728065134
0.316414934686472 3.13751135989317
0.320692904132959 3.2789702097412
0.324970873579446 3.42402600901483
0.329248843025934 3.57012635852979
0.333526812472421 3.71255731246697
0.337804781918908 3.84605485724988
0.342082751365396 3.96733180119118
0.346360720811883 4.07645929872646
0.35063869025837 4.17592886898626
0.354916659704858 4.26820863180534
0.359194629151345 4.35370553859923
0.363472598597832 4.43038053793101
0.36775056804432 4.4948457164821
0.372028537490807 4.54396641654451
0.376306506937294 4.5760594237927
0.380584476383782 4.59120985548243
0.384862445830269 4.59070587306882
0.389140415276756 4.57602385129931
0.393418384723244 4.54795321810875
0.397696354169731 4.50622164704657
0.401974323616218 4.44969706171698
0.406252293062706 4.37714610297971
0.410530262509193 4.2883477885081
0.41480823195568 4.18494768285181
0.419086201402168 4.07031014379834
0.423364170848655 3.94824930113487
0.427642140295142 3.82146499853592
0.43192010974163 3.69086368656327
0.436198079188117 3.55630302688987
0.440476048634604 3.41813806631682
0.444754018081092 3.27823104693418
0.449031987527579 3.13950469368375
0.453309956974067 3.00439376625336
0.457587926420554 2.87353759024873
0.461865895867041 2.74579967666929
0.466143865313529 2.6194411596911
0.470421834760016 2.49327700756494
0.474699804206503 2.36689895386447
0.478977773652991 2.24019451270281
0.483255743099478 2.11310909402224
0.487533712545965 1.98612554921784
0.491811681992453 1.86090351030845
0.49608965143894 1.74013847105963
0.500367620885427 1.62638487033159
0.504645590331915 1.52062883932018
0.508923559778402 1.4217753596987
0.513201529224889 1.3275492373546
0.517479498671377 1.23615569419598
0.521757468117864 1.14742423406932
0.526035437564351 1.06263861604345
0.530313407010839 0.983379300466226
0.534591376457326 0.910419656865745
0.538869345903813 0.843445762404829
0.543147315350301 0.781531824830284
0.547425284796788 0.723750993092134
0.551703254243275 0.669428986463977
0.555981223689763 0.618050423699205
0.56025919313625 0.569149887806217
0.564537162582737 0.522403056946305
0.568815132029225 0.477803209118858
0.573093101475712 0.435690023445559
0.5773710709222 0.396614068686566
0.581649040368687 0.361241100135709
0.585927009815174 0.330375307175293
0.590204979261661 0.304829455803404
0.594482948708149 0.284796700064292
0.598760918154636 0.26888814054002
0.603038887601123 0.253714100650253
0.607316857047611 0.234926507934101
0.611594826494098 0.209552832103579
0.615872795940586 0.178054194896688
0.620150765387073 0.144372349146527
0.62442873483356 0.113770270214823
0.628706704280048 0.0900909573744426
0.632984673726535 0.0743472675599129
0.637262643173022 0.0651936937108793
0.64154061261951 0.0603514641636551
0.645818582065997 0.0577623470587329
0.650096551512484 0.0559864092595374
0.654374520958972 0.054124080524951
0.658652490405459 0.0516896235915901
0.662930459851946 0.0485657963344706
0.667208429298434 0.0449335431809815
0.671486398744921 0.0411081114843485
0.675764368191408 0.037354748571753
0.680042337637896 0.0337957490447695
0.684320307084383 0.0304418814143736
0.68859827653087 0.0272846560094423
0.692876245977358 0.0243493097921809
0.697154215423845 0.0216586347643323
0.701432184870332 0.0191591535971187
0.70571015431682 0.0167138221597404
0.709988123763307 0.0142010549805703
0.714266093209794 0.0116409608956758
0.718544062656282 0.00922697257568687
0.722822032102769 0.00721958973710367
0.727100001549257 0.00577773470566551
0.731377970995744 0.00485448704878775
0.735655940442231 0.00423195200076789
0.739933909888719 0.00366404704782416
0.744211879335206 0.00302095878823637
0.748489848781693 0.00233966801113322
0.752767818228181 0.0017638317408471
0.757045787674668 0.00143209057260604
0.761323757121155 0.00139043968826113
0.765601726567642 0.00156861688966438
0.76987969601413 0.00181702719698602
0.774157665460617 0.00197641095268889
0.778435634907105 0.00194486132414858
0.782713604353592 0.00171118306549486
0.786991573800079 0.00134220727017482
0.791269543246567 0.000937841418989216
0.795547512693054 0.000583275168588478
0.799825482139541 0.000322409184561757
0.804103451586029 0.000158051737434309
0.808381421032516 6.8538618517218e-05
0.812659390479003 2.62207548294545e-05
0.816937359925491 8.82684645030214e-06
};
\addlegendentry{Yaw}
\end{axis}

\end{tikzpicture}

%% file: images/serfiq-nisk-qscores.tex

\begin{tikzpicture}

\tikzset{lw/.style = {line width=4pt}}

\definecolor{color0}{rgb}{0.823529411764706,0.411764705882353,0.117647058823529}
\definecolor{color1}{rgb}{0,1,1}
\definecolor{color2}{rgb}{0.43921568627451,0.501960784313725,0.564705882352941}
\definecolor{color3}{rgb}{0.580392156862745,0,0.827450980392157}

\begin{axis}[
legend cell align={left},
legend style={
  fill opacity=0.8,
  draw opacity=1,
  text opacity=1,
  at={(0.5,-0.3)},
  anchor=north,
  nodes={scale=0.6, transform shape},
  legend columns=2,
  draw=white!80!black
},
tick align=outside,
tick pos=left,
width=1.8in,
scale only axis,
x grid style={white!69.0196078431373!black},
xlabel={Quality Scores},
xmin=0.82, xmax=0.908691738774647,
xtick style={
    color=black,
    /pgf/number format/.cd,
    fixed,
    fixed zerofill,
    precision=2
},
xtick={0.78, 0.8, 0.82,0.84,0.86,0.88,0.9,0.92},
xticklabels={0.78, 0.8,0.82,0.84,0.86,0.88,0.90,0.92},
y grid style={white!69.0196078431373!black},
ylabel={},
ymin=0, ymax=69.1395988320628,
ytick style={color=black}
]
\addplot [semithick, color0]
table {%
0.803675091270536 0.000247920861670693
0.804170114825008 0.000965503017276545
0.804665138379479 0.00298265721159361
0.805160161933951 0.00735228876116331
0.805655185488422 0.0145989883634805
0.806150209042894 0.0236795032164028
0.806645232597365 0.0319398094495641
0.807140256151837 0.0364709736029308
0.807635279706308 0.0356346002870375
0.80813030326078 0.0297318281035941
0.808625326815251 0.0208964000270005
0.809120350369723 0.0121447956360195
0.809615373924194 0.00573884401393946
0.810110397478666 0.00217726482326936
0.810605421033137 0.000657748401213345
0.811100444587609 0.000157424708558002
0.81159546814208 2.97618995341708e-05
0.812090491696552 4.43693834254225e-06
0.812585515251023 5.21101835081878e-07
0.813080538805495 4.81884766385425e-08
0.813575562359966 3.50762933448772e-09
0.814070585914438 2.00937372115591e-10
0.814565609468909 9.05823076024181e-12
0.815060633023381 3.21320849661132e-13
0.815555656577852 8.96880579651184e-15
0.816050680132324 1.96980604239866e-16
0.816545703686795 3.40410413138741e-18
0.817040727241267 4.7464530839714e-20
0.817535750795738 1.05662926868265e-19
0.81803077435021 7.39560628462849e-18
0.818525797904681 4.09212641850079e-16
0.819020821459153 1.78159434726869e-14
0.819515845013624 6.10313381832417e-13
0.820010868568096 1.6450581233968e-11
0.820505892122567 3.4889432690758e-10
0.821000915677039 5.82225104675903e-09
0.82149593923151 7.64491154900987e-08
0.821990962785982 7.89839761870364e-07
0.822485986340453 6.42084117921297e-06
0.822981009894925 4.10710226831123e-05
0.823476033449396 0.000206720024534089
0.823971057003868 0.000818788361902856
0.824466080558339 0.00255282000371663
0.824961104112811 0.00627027122277033
0.825456127667282 0.0121637563367374
0.825951151221754 0.0187809411589772
0.826446174776226 0.0236151803561256
0.826941198330697 0.0257106145845403
0.827436221885169 0.0273539183899372
0.82793124543964 0.031736991476599
0.828426268994112 0.0385661269915561
0.828921292548583 0.0431773308603127
0.829416316103055 0.040884530283479
0.829911339657526 0.0317278165608313
0.830406363211998 0.0204205495302662
0.830901386766469 0.0123472723400167
0.831396410320941 0.0102634039995356
0.831891433875412 0.0137311923775493
0.832386457429884 0.020074663615196
0.832881480984355 0.0261392654512084
0.833376504538827 0.0306091980501893
0.833871528093298 0.0347317870488785
0.83436655164777 0.0399703524532035
0.834861575202241 0.0456367084307716
0.835356598756713 0.0503108550237294
0.835851622311184 0.055069895949098
0.836346645865656 0.0629709534407995
0.836841669420127 0.0743262102785548
0.837336692974599 0.0843856900123674
0.83783171652907 0.0879363059199464
0.838326740083542 0.08566653157705
0.838821763638013 0.084107451835789
0.839316787192485 0.0892073744042483
0.839811810746956 0.101123570393183
0.840306834301428 0.115520921880648
0.840801857855899 0.128897098204861
0.841296881410371 0.14139398664373
0.841791904964842 0.154470033857757
0.842286928519314 0.167154536036604
0.842781952073785 0.175321976838409
0.843276975628257 0.175376472075781
0.843771999182728 0.171125391163842
0.8442670227372 0.177509387669314
0.844762046291671 0.212425053682589
0.845257069846143 0.279571227403023
0.845752093400614 0.359388102898148
0.846247116955086 0.419237717004302
0.846742140509557 0.4344148561972
0.847237164064029 0.402446869167152
0.8477321876185 0.341318803766367
0.848227211172972 0.276436006666435
0.848722234727443 0.227666735451386
0.849217258281915 0.203502726334967
0.849712281836386 0.203223721176352
0.850207305390858 0.224538478246709
0.850702328945329 0.269816168827025
0.851197352499801 0.343572317255435
0.851692376054272 0.443414973950375
0.852187399608744 0.555145451648648
0.852682423163215 0.658913594609429
0.853177446717687 0.742265831678681
0.853672470272158 0.807037170345003
0.85416749382663 0.860932994684534
0.854662517381101 0.903671143574249
0.855157540935573 0.926168661893697
0.855652564490044 0.923044001287092
0.856147588044516 0.903360211569444
0.856642611598987 0.892060699713512
0.857137635153459 0.922624167381454
0.85763265870793 1.02102367644806
0.858127682262402 1.18757057963783
0.858622705816873 1.39295799958419
0.859117729371345 1.5974069291776
0.859612752925816 1.77794030487372
0.860107776480288 1.93921088529978
0.860602800034759 2.10444654753662
0.861097823589231 2.30061437976109
0.861592847143702 2.54210049111891
0.862087870698174 2.81704367183363
0.862582894252645 3.09750743966506
0.863077917807117 3.37597274066681
0.863572941361588 3.68748676209608
0.86406796491606 4.0847309443971
0.864562988470531 4.58880187085809
0.865058012025003 5.1669763295431
0.865553035579474 5.76179700957206
0.866048059133946 6.34446854809352
0.866543082688417 6.93751502441893
0.867038106242889 7.58348990928011
0.86753312979736 8.30187086241846
0.868028153351832 9.08290053752439
0.868523176906303 9.90246613358642
0.869018200460775 10.7299404823879
0.869513224015246 11.5554487970294
0.870008247569718 12.4341168166593
0.870503271124189 13.4677331538167
0.870998294678661 14.7118409299665
0.871493318233132 16.1249557412519
0.871988341787604 17.6280922247989
0.872483365342075 19.1942378872479
0.872978388896547 20.8708745136972
0.873473412451018 22.7289515192712
0.87396843600549 24.7914511346963
0.874463459559961 27.0071379109725
0.874958483114433 29.2918458620469
0.875453506668904 31.5772535092856
0.875948530223376 33.813868208213
0.876443553777847 35.9699280347547
0.876938577332319 38.0612570863515
0.87743360088679 40.1475912425179
0.877928624441262 42.2632031412249
0.878423647995733 44.3855738740329
0.878918671550205 46.5160817135347
0.879413695104676 48.7549645471305
0.879908718659148 51.2173034353685
0.88040374221362 53.8622224774383
0.880898765768091 56.4644181099956
0.881393789322563 58.7810147398302
0.881888812877034 60.7200940194638
0.882383836431506 62.3516361263499
0.882878859985977 63.7815572081917
0.883373883540449 64.9915447249961
0.88386890709492 65.7744582411334
0.884363930649392 65.847236982917
0.884858954203863 65.0412536193229
0.885353977758335 63.3920039013328
0.885849001312806 61.1012982834471
0.886344024867278 58.4443127239036
0.886839048421749 55.6378312987352
0.887334071976221 52.7108355620182
0.887829095530692 49.5033922012623
0.888324119085164 45.8302862209877
0.888819142639635 41.6639719868291
0.889314166194107 37.16985009719
0.889809189748578 32.5881591902058
0.89030421330305 28.1049905213675
0.890799236857521 23.821891296426
0.891294260411993 19.7981902971625
0.891789283966464 16.0941020718814
0.892284307520936 12.7841325178217
0.892779331075407 9.93803621755348
0.893274354629879 7.58519504661621
0.89376937818435 5.69842685160888
0.894264401738822 4.21291880114196
0.894759425293293 3.05671726535293
0.895254448847765 2.16919335783146
0.895749472402236 1.50579438002631
0.896244495956708 1.03154544767426
0.896739519511179 0.708391687886913
0.897234543065651 0.491575725818
0.897729566620122 0.340744994315964
0.898224590174594 0.230695885178762
0.898719613729065 0.149646906548485
0.899214637283537 0.0914459372683915
0.899709660838008 0.0514048379154086
0.90020468439248 0.0257452397911877
0.900699707946951 0.0111160085915532
0.901194731501423 0.00402638338160577
0.901689755055894 0.00119939085184442
0.902184778610366 0.000289826691177517
};
\addlegendentry{FRGC v2.0}
\addplot [semithick, green!39.2156862745098!black]
table {%
0.788287172329539 0.00011847568957754
0.788832143754803 0.000379405576917297
0.789377115180067 0.00102113045791824
0.789922086605331 0.00230973313794132
0.790467058030596 0.004390817990699
0.79101202945586 0.00701507178621342
0.791557000881124 0.009419370051568
0.792101972306389 0.0106295444730885
0.792646943731653 0.0100811598816991
0.793191915156917 0.0080354352610863
0.793736886582181 0.00538283869037419
0.794281858007446 0.00303051459187534
0.79482682943271 0.00143391847832139
0.795371800857974 0.000570211538378513
0.795916772283239 0.000190573808043891
0.796461743708503 5.35783335778199e-05
0.797006715133767 1.30220409401469e-05
0.797551686559032 4.92816681325938e-06
0.798096657984296 1.32066437849479e-05
0.79864162940956 5.70512329474611e-05
0.799186600834824 0.000214417375383112
0.799731572260089 0.000681258816091243
0.800276543685353 0.0018290820356436
0.800821515110617 0.00415201160257833
0.801366486535881 0.00797491821244424
0.801911457961146 0.0129787299330078
0.80245642938641 0.0179506481694199
0.803001400811674 0.0212542481940601
0.803546372236939 0.021936925161198
0.804091343662203 0.020572188541231
0.804636315087467 0.0189241954580949
0.805181286512732 0.0185974391478563
0.805726257937996 0.0198767062034947
0.80627122936326 0.0217956304166836
0.806816200788524 0.0232941875586696
0.807361172213789 0.0243347085430048
0.807906143639053 0.0259158209849149
0.808451115064317 0.028951468051196
0.808996086489582 0.0330747466144705
0.809541057914846 0.0365434528593248
0.81008602934011 0.0373637831934977
0.810631000765374 0.0345963517525967
0.811175972190639 0.0288004172320913
0.811720943615903 0.0215397005540916
0.812265915041167 0.0145856991756457
0.812810886466432 0.0093383656171064
0.813355857891696 0.00655643644054293
0.81390082931696 0.00628622762592843
0.814445800742225 0.00791995698294766
0.814990772167489 0.010453378048366
0.815535743592753 0.0130376439115031
0.816080715018017 0.0156164812183239
0.816625686443282 0.0190662732236392
0.817170657868546 0.0244834045021978
0.81771562929381 0.0321283109167104
0.818260600719074 0.0410649116415915
0.818805572144339 0.0499783834852972
0.819350543569603 0.0585654396112469
0.819895514994867 0.0683726414392559
0.820440486420132 0.0823321515008383
0.820985457845396 0.103069590371297
0.82153042927066 0.130906536283181
0.822075400695925 0.162929030865636
0.822620372121189 0.194001470492661
0.823165343546453 0.219289119561692
0.823710314971717 0.236759199952618
0.824255286396982 0.248107707304832
0.824800257822246 0.257550586159948
0.82534522924751 0.269322469996587
0.825890200672775 0.28556094586691
0.826435172098039 0.30575955578933
0.826980143523303 0.327835411229653
0.827525114948567 0.350253851167032
0.828070086373832 0.374107270465863
0.828615057799096 0.403084625547442
0.82916002922436 0.440065598303987
0.829705000649625 0.482799483208931
0.830249972074889 0.523483663057869
0.830794943500153 0.55368348035585
0.831339914925417 0.570395305031761
0.831884886350682 0.578082967734589
0.832429857775946 0.585713205094638
0.83297482920121 0.601915578548211
0.833519800626475 0.631936762048213
0.834064772051739 0.677530614000303
0.834609743477003 0.73801391321886
0.835154714902267 0.810559327675744
0.835699686327532 0.890149478893681
0.836244657752796 0.970464407380005
0.83678962917806 1.04526892777611
0.837334600603325 1.10924164271536
0.837879572028589 1.15862326240228
0.838424543453853 1.19237682866253
0.838969514879117 1.21332269698857
0.839514486304382 1.22874905398307
0.840059457729646 1.25085023787155
0.84060442915491 1.29612992389989
0.841149400580175 1.38136193425524
0.841694372005439 1.51620108635943
0.842239343430703 1.69726338855071
0.842784314855967 1.90902196132513
0.843329286281232 2.13195513579107
0.843874257706496 2.35282810676873
0.84441922913176 2.57004420234829
0.844964200557025 2.79040228849395
0.845509171982289 3.02096521280182
0.846054143407553 3.26423236066988
0.846599114832817 3.52015358304508
0.847144086258082 3.78996760929304
0.847689057683346 4.07598571647053
0.84823402910861 4.37864429056421
0.848779000533875 4.69609378775113
0.849323971959139 5.02727797504984
0.849868943384403 5.37428571503774
0.850413914809668 5.74063661179646
0.850958886234932 6.127086513809
0.851503857660196 6.52970470362253
0.85204882908546 6.94321167385871
0.852593800510725 7.3672363736375
0.853138771935989 7.8096600025643
0.853683743361253 8.28403914352211
0.854228714786518 8.80424459296782
0.854773686211782 9.38138690464526
0.855318657637046 10.0239579158326
0.85586362906231 10.7375917604583
0.856408600487575 11.5211128780264
0.856953571912839 12.3598861554514
0.857498543338103 13.2217823217896
0.858043514763368 14.0618904035091
0.858588486188632 14.8379198206064
0.859133457613896 15.5306314981922
0.85967842903916 16.1585066158318
0.860223400464425 16.7775543964215
0.860768371889689 17.4625286762934
0.861313343314953 18.2727449632769
0.861858314740218 19.2188772398117
0.862403286165482 20.2589887392431
0.862948257590746 21.3366222736981
0.86349322901601 22.4306931104655
0.864038200441275 23.5668190060737
0.864583171866539 24.778507226134
0.865128143291803 26.0608014796984
0.865673114717068 27.3624945827011
0.866218086142332 28.6197155663972
0.866763057567596 29.7982070612132
0.867308028992861 30.9113449691545
0.867853000418125 32.0039086267726
0.868397971843389 33.1173528420511
0.868942943268653 34.2630267288554
0.869487914693918 35.4184010384866
0.870032886119182 36.5426422080945
0.870577857544446 37.6001515254268
0.87112282896971 38.5813868751319
0.871667800394975 39.5095531873456
0.872212771820239 40.4252495486959
0.872757743245503 41.3562735696064
0.873302714670768 42.2947552290286
0.873847686096032 43.2005277383566
0.874392657521296 44.0270421653707
0.87493762894656 44.7436234736226
0.875482600371825 45.3306451117918
0.876027571797089 45.7552493461128
0.876572543222353 45.9606947710808
0.877117514647618 45.8877203650563
0.877662486072882 45.5062817185364
0.878207457498146 44.8201462055619
0.878752428923411 43.8355190550885
0.879297400348675 42.5259610296143
0.879842371773939 40.8345544606602
0.880387343199203 38.7201288823973
0.880932314624468 36.2102549313454
0.881477286049732 33.4131993129195
0.882022257474996 30.4742819273509
0.882567228900261 27.5080723973325
0.883112200325525 24.5578965063575
0.883657171750789 21.6126693347852
0.884202143176053 18.6641662317876
0.884747114601318 15.7550395397356
0.885292086026582 12.9800997050762
0.885837057451846 10.4470310737245
0.886382028877111 8.23290915984837
0.886927000302375 6.3657891666074
0.887471971727639 4.83364937410164
0.888016943152903 3.60448871538
0.888561914578168 2.64128402322853
0.889106886003432 1.90634956901155
0.889651857428696 1.36006670084539
0.890196828853961 0.961488184228136
0.890741800279225 0.672903125238112
0.891286771704489 0.464360618709771
0.891831743129753 0.314314461315255
0.892376714555018 0.20726767360233
0.892921685980282 0.131652378371853
0.893466657405546 0.0791051899363273
0.894011628830811 0.0439466040452209
0.894556600256075 0.0220571499659927
0.895101571681339 0.00980659892836806
0.895646543106604 0.00380463806491697
0.896191514531868 0.00127433126423304
0.896736485957132 0.000365769089553747
};
\addlegendentry{Base Images}
\addplot [semithick, color1]
table {%
0.7917565317312 8.35292269607226e-05
0.792282732884092 0.000296920967980927
0.792808934036984 0.000856005734030389
0.793335135189877 0.00200145533734467
0.793861336342769 0.00379532279652029
0.794387537495661 0.00583695434189647
0.794913738648553 0.00728059868885255
0.795439939801445 0.00736634406340046
0.795966140954337 0.00605172187800482
0.796492342107229 0.00406610292760729
0.797018543260121 0.00234734070587108
0.797544744413013 0.00150979906956157
0.798070945565905 0.0017688536073705
0.798597146718798 0.00306460430950404
0.79912334787169 0.00501092085523633
0.799649549024582 0.00680462652572702
0.800175750177474 0.00751939251217955
0.800701951330366 0.00674218097428918
0.801228152483258 0.00490356085514554
0.80175435363615 0.0028948732395002
0.802280554789042 0.00139973297246237
0.802806755941934 0.000610666227652671
0.803332957094826 0.00044137748408508
0.803859158247719 0.000863855403742417
0.804385359400611 0.00203931856796243
0.804911560553503 0.00415375401887272
0.805437761706395 0.00708101880792454
0.805963962859287 0.0101746003640574
0.806490164012179 0.0124547967403064
0.807016365165071 0.0131196376699457
0.807542566317963 0.0120309729682318
0.808068767470855 0.00987779034075421
0.808594968623747 0.00788211361152708
0.80912116977664 0.00711980144585508
0.809647370929532 0.00790196598653048
0.810173572082424 0.00981208826423068
0.810699773235316 0.0124028735807036
0.811225974388208 0.015714941633265
0.8117521755411 0.019933533796424
0.812278376693992 0.0247212829299001
0.812804577846884 0.0292669774502548
0.813330778999776 0.0329581454683482
0.813856980152668 0.0356602667951623
0.814383181305561 0.0373671953339195
0.814909382458453 0.0380117977956047
0.815435583611345 0.0378678863192604
0.815961784764237 0.0380157032324401
0.816487985917129 0.040136062216292
0.817014187070021 0.0455854353656351
0.817540388222913 0.0545551248918558
0.818066589375805 0.0660645320991574
0.818592790528697 0.0784991658149431
0.81911899168159 0.0899764655374456
0.819645192834482 0.0986905593932525
0.820171393987374 0.103804907807845
0.820697595140266 0.10633758200722
0.821223796293158 0.108628727171304
0.82174999744605 0.112412553378751
0.822276198598942 0.117740499454861
0.822802399751834 0.124366924290508
0.823328600904726 0.13388410781554
0.823854802057618 0.149632939149836
0.824381003210511 0.174012378412277
0.824907204363403 0.205594336684246
0.825433405516295 0.238711249992943
0.825959606669187 0.266265540667715
0.826485807822079 0.283681353634344
0.827012008974971 0.291053907610074
0.827538210127863 0.292805249794919
0.828064411280755 0.295772499262081
0.828590612433647 0.306227485946387
0.829116813586539 0.326890641277372
0.829643014739432 0.355614868857072
0.830169215892324 0.386041722367461
0.830695417045216 0.410444016270674
0.831221618198108 0.425527502537059
0.831747819351 0.437690875729238
0.832274020503892 0.460315056972541
0.832800221656784 0.502541523790281
0.833326422809676 0.560482841963161
0.833852623962568 0.620854670834166
0.83437882511546 0.672844601257201
0.834905026268353 0.715331846430646
0.835431227421245 0.754271795849249
0.835957428574137 0.796374290339769
0.836483629727029 0.845575435869429
0.837009830879921 0.903125112259341
0.837536032032813 0.969643764080576
0.838062233185705 1.04678528447884
0.838588434338597 1.13603842492527
0.839114635491489 1.23535516095047
0.839640836644381 1.33714069497696
0.840167037797274 1.42959967530218
0.840693238950166 1.50119229763924
0.841219440103058 1.54711512405334
0.84174564125595 1.57485610969367
0.842271842408842 1.60453771988905
0.842798043561734 1.66130805409247
0.843324244714626 1.7611603667825
0.843850445867518 1.89929303913394
0.84437664702041 2.05432181035545
0.844902848173303 2.20804182450979
0.845429049326195 2.35992031512124
0.845955250479087 2.52047371552376
0.846481451631979 2.69537383125872
0.847007652784871 2.88147087102189
0.847533853937763 3.07577662166165
0.848060055090655 3.28267416650716
0.848586256243547 3.5110415836152
0.849112457396439 3.76409788728511
0.849638658549331 4.03001002943664
0.850164859702224 4.28476467026162
0.850691060855116 4.51272991949256
0.851217262008008 4.72911488010708
0.8517434631609 4.97721167415102
0.852269664313792 5.29676064458373
0.852795865466684 5.69364054610237
0.853322066619576 6.13982529856699
0.853848267772468 6.59896736804471
0.85437446892536 7.05064468824215
0.854900670078252 7.4945714795638
0.855426871231145 7.93832352253008
0.855953072384037 8.38729190052763
0.856479273536929 8.84988885503325
0.857005474689821 9.34913489930848
0.857531675842713 9.92156053298661
0.858057876995605 10.5985907708521
0.858584078148497 11.3827682765626
0.859110279301389 12.2360873808989
0.859636480454281 13.0938037422189
0.860162681607173 13.9014989970827
0.860688882760066 14.651076499626
0.861215083912958 15.3866096148342
0.86174128506585 16.1722653555619
0.862267486218742 17.0466835508283
0.862793687371634 18.0049348935634
0.863319888524526 19.0230869208211
0.863846089677418 20.0920672719422
0.86437229083031 21.2232983761914
0.864898491983202 22.4307868030735
0.865424693136094 23.7141579395807
0.865950894288987 25.0513429849282
0.866477095441879 26.402110127716
0.867003296594771 27.7252180782621
0.867529497747663 29.000020823247
0.868055698900555 30.2350009667511
0.868581900053447 31.4597113305461
0.869108101206339 32.7099321909241
0.869634302359231 34.0093990105269
0.870160503512123 35.3513465101222
0.870686704665016 36.7021810982289
0.871212905817908 38.0407329230242
0.8717391069708 39.3942054728389
0.872265308123692 40.8140956009677
0.872791509276584 42.301685979986
0.873317710429476 43.7672309422371
0.873843911582368 45.0848635150579
0.87437011273526 46.195393412203
0.874896313888152 47.1430500394792
0.875422515041044 48.0060836132287
0.875948716193937 48.8058301245905
0.876474917346829 49.4889703099366
0.877001118499721 49.9731991790441
0.877527319652613 50.1836327250053
0.878053520805505 50.0559228314305
0.878579721958397 49.5387467090816
0.879105923111289 48.6052896468825
0.879632124264181 47.2465995803105
0.880158325417073 45.4495515576549
0.880684526569965 43.2015033751183
0.881210727722858 40.5309894131864
0.88173692887575 37.5360678002163
0.882263130028642 34.3594211089886
0.882789331181534 31.129355871809
0.883315532334426 27.9188289897431
0.883841733487318 24.749651698581
0.88436793464021 21.6250708124611
0.884894135793102 18.5586251967929
0.885420336945994 15.5853582373426
0.885946538098886 12.7622335791386
0.886472739251779 10.1647269924695
0.886998940404671 7.87456393517609
0.887525141557563 5.95364103987026
0.888051342710455 4.41701589740355
0.888577543863347 3.22850684628693
0.889103745016239 2.32465073829973
0.889629946169131 1.64505106100675
0.890156147322023 1.14446424802214
0.890682348474915 0.786121047940052
0.891208549627807 0.534421799677485
0.8917347507807 0.357425862036663
0.892260951933592 0.23224655572841
0.892787153086484 0.14475603954174
0.893313354239376 0.0854395174211072
0.893839555392268 0.0468617817439517
0.89436575654516 0.0232506324874587
0.894891957698052 0.0101310956713924
0.895418158850944 0.00377866216369687
0.895944360003836 0.00118382998029514
0.896470561156729 0.000307655699728223
};
\addlegendentry{PCA-FR}
\addplot [semithick, red!54.5098039215686!black]
table {%
0.787849886295078 9.11638029831745e-05
0.788399306688997 0.000335430627310523
0.788948727082917 0.0009883914973058
0.789498147476837 0.00233239065557838
0.790047567870756 0.00440777675973627
0.790596988264676 0.00667089177187489
0.791146408658595 0.00808526407881291
0.791695829052515 0.00784784839382564
0.792245249446434 0.00610034217431178
0.792794669840354 0.00379771752039312
0.793344090234273 0.00189468114096104
0.793893510628193 0.0007652540419222
0.794442931022112 0.000289342640376407
0.794992351416032 0.000255430653261544
0.795541771809951 0.000641993233997356
0.796091192203871 0.00164068701827547
0.79664061259779 0.00341718032628501
0.79719003299171 0.00570611518352202
0.797739453385629 0.00763114582699373
0.798288873779549 0.00817310192967923
0.798838294173469 0.00701019932230318
0.799387714567388 0.00481529139245478
0.799937134961308 0.0026491548254679
0.800486555355227 0.00116930493507504
0.801035975749147 0.000426051751323422
0.801585396143066 0.000185254822895746
0.802134816536986 0.000291263270984513
0.802684236930905 0.000820628241243237
0.803233657324825 0.00201566905610659
0.803783077718744 0.00400259122410217
0.804332498112664 0.00644507869275112
0.804881918506583 0.00859508255869305
0.805431338900503 0.00999820636290973
0.805980759294422 0.0111480652070793
0.806530179688342 0.0129888963011183
0.807079600082261 0.0155205909499038
0.807629020476181 0.0173220207984252
0.8081784408701 0.0169728919206508
0.80872786126402 0.0148019924260542
0.80927728165794 0.0127945280518106
0.809826702051859 0.0129182968592828
0.810376122445779 0.0159994631901124
0.810925542839698 0.0219286046910227
0.811474963233618 0.0298798397653457
0.812024383627537 0.0378819985865809
0.812573804021457 0.0431687664210918
0.813123224415376 0.0441402580254335
0.813672644809296 0.0419079466153198
0.814222065203215 0.0392421251959556
0.814771485597135 0.0381846867979069
0.815320905991054 0.0392658024997796
0.815870326384974 0.0425974381184039
0.816419746778893 0.048560542846318
0.816969167172813 0.0569728404954626
0.817518587566732 0.0661050933718546
0.818068007960652 0.073292316651875
0.818617428354571 0.0770461749919899
0.819166848748491 0.0785467552946098
0.81971626914241 0.0805193658915317
0.82026568953633 0.0842554910739858
0.82081510993025 0.0880830480331065
0.821364530324169 0.0894516708701157
0.821913950718089 0.0885507028846874
0.822463371112008 0.0891951624413654
0.823012791505928 0.0958327106705021
0.823562211899847 0.110025538541403
0.824111632293767 0.129955992803317
0.824661052687686 0.152711617719535
0.825210473081606 0.176170264691378
0.825759893475525 0.198504006994224
0.826309313869445 0.217250181533467
0.826858734263364 0.230929903596339
0.827408154657284 0.24166872957608
0.827957575051203 0.254377042553081
0.828506995445123 0.272796090025819
0.829056415839042 0.297364182155735
0.829605836232962 0.326439235628317
0.830155256626881 0.357599132130573
0.830704677020801 0.387771232976172
0.831254097414721 0.414716568275373
0.83180351780864 0.439940572198138
0.83235293820256 0.4686004477936
0.832902358596479 0.504489734502647
0.833451778990399 0.54548599445916
0.834001199384318 0.585524086866672
0.834550619778238 0.620623138378444
0.835100040172157 0.651560493505493
0.835649460566077 0.682556668675001
0.836198880959996 0.721097875017472
0.836748301353916 0.777582687492174
0.837297721747835 0.858552007502631
0.837847142141755 0.957478629503578
0.838396562535674 1.05640265432562
0.838945982929594 1.14034539639831
0.839495403323513 1.20870601070023
0.840044823717433 1.27106855199848
0.840594244111352 1.33476762235016
0.841143664505272 1.40014996477504
0.841693084899191 1.46669126052808
0.842242505293111 1.53818019251054
0.842791925687031 1.61923177336771
0.84334134608095 1.71094563311894
0.84389076647487 1.81431532483075
0.844440186868789 1.93509762704388
0.844989607262709 2.07979946410039
0.845539027656628 2.24662146938309
0.846088448050548 2.42534964974771
0.846637868444467 2.61031649221286
0.847187288838387 2.8116820165236
0.847736709232306 3.04772916137192
0.848286129626226 3.32326565944986
0.848835550020145 3.61990220514698
0.849384970414065 3.91297273757948
0.849934390807984 4.198744906826
0.850483811201904 4.50172152076548
0.851033231595823 4.8519879951126
0.851582651989743 5.25667172599332
0.852132072383663 5.69631939781564
0.852681492777582 6.14369492356498
0.853230913171502 6.57787239384061
0.853780333565421 6.98734214449124
0.854329753959341 7.37717971487098
0.85487917435326 7.77624366662355
0.85542859474718 8.22394566936209
0.855978015141099 8.74063906617332
0.856527435535019 9.31333018213625
0.857076855928938 9.91615233290803
0.857626276322858 10.5445932977101
0.858175696716777 11.2211033117312
0.858725117110697 11.9619918076922
0.859274537504616 12.7490295665022
0.859823957898536 13.5480679751709
0.860373378292455 14.3548642436447
0.860922798686375 15.2088455243336
0.861472219080294 16.1532779759848
0.862021639474214 17.1869421463538
0.862571059868134 18.2629771321857
0.863120480262053 19.3332351916843
0.863669900655972 20.3866222883481
0.864219321049892 21.4449453749693
0.864768741443812 22.5298527975526
0.865318161837731 23.6412580269749
0.865867582231651 24.7728173158382
0.86641700262557 25.9487441927782
0.86696642301949 27.2320732306804
0.867515843413409 28.6749158569222
0.868065263807329 30.2552776314199
0.868614684201248 31.8784521784256
0.869164104595168 33.4485731383883
0.869713524989087 34.9293374390709
0.870262945383007 36.3375076578015
0.870812365776926 37.7053103285889
0.871361786170846 39.0656374944224
0.871911206564765 40.4504898446737
0.872460626958685 41.8711198977202
0.873010047352604 43.2953876629739
0.873559467746524 44.6572388277032
0.874108888140444 45.8914515332269
0.874658308534363 46.9572788095634
0.875207728928283 47.8370240750555
0.875757149322202 48.5238401654903
0.876306569716122 49.0077456697732
0.876855990110041 49.2589028660816
0.877405410503961 49.225180751841
0.87795483089788 48.8653147650871
0.8785042512918 48.1908885540377
0.879053671685719 47.2559243429614
0.879603092079639 46.0827476196601
0.880152512473558 44.5966827216871
0.880701932867478 42.6539725837
0.881251353261397 40.1561456055446
0.881800773655317 37.1437111173449
0.882350194049236 33.7796216939234
0.882899614443156 30.2515839882811
0.883449034837075 26.6917603563273
0.883998455230995 23.1701013306766
0.884547875624915 19.7421259411993
0.885097296018834 16.4924235438578
0.885646716412753 13.5270617325016
0.886196136806673 10.9233065338291
0.886745557200593 8.69093555589654
0.887294977594512 6.78458714898529
0.887844397988432 5.15317468867368
0.888393818382351 3.77869148389603
0.888943238776271 2.67107771751511
0.88949265917019 1.83234684633772
0.89004207956411 1.23262663928053
0.890591499958029 0.819320057128984
0.891140920351949 0.540189072281768
0.891690340745868 0.354929742565696
0.892239761139788 0.233591875173561
0.892789181533707 0.153380286531603
0.893338601927627 0.0984460308464754
0.893888022321546 0.059776317319866
0.894437442715466 0.0331514658322698
0.894986863109385 0.0162790238631797
0.895536283503305 0.00690202365662257
0.896085703897225 0.00247758236468183
0.896635124291144 0.000741816392959466
0.897184544685064 0.000183205308816012
};
\addlegendentry{Illumination Quality}
\addplot [semithick, color2]
table {%
0.772045575324742 7.56565082859814e-05
0.772688260514144 0.000325424748217975
0.773330945703545 0.00105270362916614
0.773973630892946 0.00256101971498586
0.774616316082348 0.00468566226574181
0.775259001271749 0.00644733596664907
0.77590168646115 0.00667176713875687
0.776544371650552 0.00519221559401782
0.777187056839953 0.00303889583098782
0.777829742029354 0.00133761203744833
0.778472427218756 0.000442787938690641
0.779115112408157 0.00011023326102903
0.779757797597558 2.06386262877469e-05
0.78040048278696 2.9060305657786e-06
0.781043167976361 3.07730702022102e-07
0.781685853165762 2.45071458930597e-08
0.782328538355163 1.46779668895176e-09
0.782971223544565 6.61135275710251e-11
0.783613908733966 2.23957708625479e-12
0.784256593923367 5.70549014714126e-14
0.784899279112769 1.09312836293162e-15
0.78554196430217 1.57507277564076e-17
0.786184649491571 1.70679500076143e-19
0.786827334680973 1.39249091968012e-21
0.787470019870374 2.88748087609787e-22
0.788112705059775 3.85184685668514e-20
0.788755390249177 3.98735516598607e-18
0.789398075438578 3.10885098971624e-16
0.790040760627979 1.82583812199553e-14
0.79068344581738 8.07834413689516e-13
0.791326131006782 2.69297339489623e-11
0.791968816196183 6.76464953706527e-10
0.792611501385584 1.28061705699285e-08
0.793254186574986 1.82733435620068e-07
0.793896871764387 1.96571445562481e-06
0.794539556953788 1.59455070010151e-05
0.79518224214319 9.75787774492178e-05
0.795824927332591 0.000450826895013072
0.796467612521992 0.00157495327113375
0.797110297711394 0.0041732724877528
0.797752982900795 0.00844040585579743
0.798395668090196 0.0131918950366778
0.799038353279598 0.0163067823446381
0.799681038468999 0.0165856721252287
0.8003237236584 0.0147351283408322
0.800966408847801 0.0123653887010562
0.801609094037203 0.010594787144524
0.802251779226604 0.00944616473754538
0.802894464416005 0.00813337763526124
0.803537149605407 0.00618982300916081
0.804179834794808 0.00429156073300473
0.804822519984209 0.00362470971296453
0.805465205173611 0.00457509803945327
0.806107890363012 0.00617462711709975
0.806750575552413 0.00683585937414335
0.807393260741815 0.00579723368930019
0.808035945931216 0.00378307070617684
0.808678631120617 0.00220871999952696
0.809321316310019 0.00222551905238184
0.80996400149942 0.00469642981477743
0.810606686688821 0.0103408535733299
0.811249371878222 0.0190963650804826
0.811892057067624 0.0292537984002352
0.812534742257025 0.0377570209016044
0.813177427446426 0.0422274508316549
0.813820112635828 0.043370868965023
0.814462797825229 0.0451971163553459
0.81510548301463 0.0519451517547864
0.815748168204032 0.0636819804467049
0.816390853393433 0.0748165088948584
0.817033538582834 0.0790427459578403
0.817676223772236 0.0771876986132543
0.818318908961637 0.0779191283797743
0.818961594151038 0.0890812383525803
0.81960427934044 0.110611449309167
0.820246964529841 0.136442454313033
0.820889649719242 0.159924540414187
0.821532334908643 0.177139976918774
0.822175020098045 0.18885313307928
0.822817705287446 0.200375096434844
0.823460390476847 0.217852771279365
0.824103075666249 0.244702239983433
0.82474576085565 0.280681241920183
0.825388446045051 0.32045488709578
0.826031131234453 0.353782765632836
0.826673816423854 0.373742002566661
0.827316501613255 0.38506234855498
0.827959186802657 0.399313015906676
0.828601871992058 0.423594594611017
0.829244557181459 0.455933511041065
0.829887242370861 0.48788930127523
0.830529927560262 0.513182392470803
0.831172612749663 0.537808219436961
0.831815297939064 0.576747122949292
0.832457983128466 0.637696433738924
0.833100668317867 0.713868053059616
0.833743353507268 0.791832085752453
0.83438603869667 0.858679039269472
0.835028723886071 0.905111868741572
0.835671409075472 0.933693957996593
0.836314094264874 0.964323329460467
0.836956779454275 1.02055195349273
0.837599464643676 1.11070576518686
0.838242149833078 1.23104536636863
0.838884835022479 1.37828084372513
0.83952752021188 1.54217885247765
0.840170205401282 1.69329076219726
0.840812890590683 1.80088005890916
0.841455575780084 1.86723960180955
0.842098260969485 1.92758499044356
0.842740946158887 2.01276538204222
0.843383631348288 2.12991693206173
0.844026316537689 2.2802433543914
0.844669001727091 2.47246015001709
0.845311686916492 2.71012262708765
0.845954372105893 2.9808386137972
0.846597057295295 3.2719971667517
0.847239742484696 3.59297179577606
0.847882427674097 3.96210240470434
0.848525112863499 4.36184923683241
0.8491677980529 4.72782083458237
0.849810483242301 5.00844180763033
0.850453168431702 5.22937703243347
0.851095853621104 5.48070515958768
0.851738538810505 5.84776831572402
0.852381223999906 6.36106007447871
0.853023909189308 6.98612445022432
0.853666594378709 7.64033334631185
0.85430927956811 8.2382499731849
0.854951964757512 8.74544612915284
0.855594649946913 9.19697970525372
0.856237335136314 9.67556157922739
0.856880020325716 10.2693550384719
0.857522705515117 11.014154049178
0.858165390704518 11.8658789205037
0.858808075893919 12.7594044254717
0.859450761083321 13.6803687071357
0.860093446272722 14.6432178008571
0.860736131462123 15.6322889590133
0.861378816651525 16.6164264341118
0.862021501840926 17.6057478004471
0.862664187030327 18.6537046475976
0.863306872219729 19.8028147591312
0.86394955740913 21.044766666381
0.864592242598531 22.3352196602203
0.865234927787933 23.6448897316679
0.865877612977334 24.9857943828449
0.866520298166735 26.3710528180392
0.867162983356137 27.7597571773383
0.867805668545538 29.0770974736032
0.868448353734939 30.3007673625386
0.86909103892434 31.4938332854494
0.869733724113742 32.7275505061122
0.870376409303143 34.000465164447
0.871019094492544 35.2610823086646
0.871661779681946 36.4965285316855
0.872304464871347 37.783558118982
0.872947150060748 39.2488377341363
0.87358983525015 40.9535663767943
0.874232520439551 42.7849185845263
0.874875205628952 44.479190813718
0.875517890818354 45.7917663817866
0.876160576007755 46.6511006705144
0.876803261197156 47.1521967227135
0.877445946386557 47.4160141023805
0.878088631575959 47.4526728168802
0.87873131676536 47.1548332757792
0.879374001954761 46.4088060982523
0.880016687144163 45.1638271353121
0.880659372333564 43.3694166100944
0.881302057522965 40.9188540918402
0.881944742712367 37.7491804288827
0.882587427901768 33.9757942202307
0.883230113091169 29.8510408849335
0.883872798280571 25.610827769792
0.884515483469972 21.4204209409001
0.885158168659373 17.4255366820378
0.885800853848774 13.7720696513568
0.886443539038176 10.5708559532601
0.887086224227577 7.87672842569083
0.887728909416978 5.69897310043017
0.88837159460638 4.01065537679399
0.889014279795781 2.74924567250481
0.889656964985182 1.83034914508343
0.890299650174584 1.17568305535265
0.890942335363985 0.726775064137798
0.891585020553386 0.435104295091545
0.892227705742788 0.253259606426874
0.892870390932189 0.140974240968522
0.89351307612159 0.0724005719443067
0.894155761310992 0.0335394250341276
0.894798446500393 0.0151539284773246
0.895441131689794 0.00890578480737409
0.896083816879196 0.00757850480549262
0.896726502068597 0.00662093023961058
0.897369187257998 0.00471143892641415
0.898011872447399 0.00256396628041399
0.898654557636801 0.00105296159769183
0.899297242826202 0.000325441982866761
0.899939928015603 7.56573846386466e-05
};
\addlegendentry{Smile}
\addplot [semithick, color3]
table {%
0.799513147366371 0.000206949894786777
0.799984072089263 0.000633964952046884
0.800454996812155 0.00166210133013969
0.800925921535046 0.00373364361264829
0.801396846257938 0.00719984350465993
0.801867770980831 0.0119568980176868
0.802338695703723 0.017190735481422
0.802809620426615 0.0215750937631165
0.803280545149506 0.0239349295193574
0.803751469872398 0.0238939392761827
0.80422239459529 0.0219933988835925
0.804693319318182 0.0193027326239516
0.805164244041074 0.0169557035413066
0.805635168763966 0.015868354203386
0.806106093486858 0.0165522185666645
0.80657701820975 0.0189639383754433
0.807047942932642 0.0225375891863676
0.807518867655534 0.0264919176434995
0.807989792378426 0.0302513950079725
0.808460717101318 0.0336890307724708
0.80893164182421 0.0369765420518489
0.809402566547102 0.0400714519891663
0.809873491269994 0.0422414455256851
0.810344415992886 0.0422212025608416
0.810815340715778 0.0391849384005606
0.81128626543867 0.0338835535257098
0.811757190161562 0.0289255533978739
0.812228114884454 0.0277862627722934
0.812699039607346 0.0331267688639734
0.813169964330238 0.0454739541163553
0.81364088905313 0.0630115532251183
0.814111813776022 0.0825186914132001
0.814582738498914 0.100846743077322
0.815053663221806 0.116170931534621
0.815524587944698 0.128593867677386
0.81599551266759 0.13994867877512
0.816466437390482 0.152640833194915
0.816937362113374 0.167746399459534
0.817408286836266 0.183534596676014
0.817879211559158 0.195882561265651
0.81835013628205 0.200766169779778
0.818821061004942 0.197218242813443
0.819291985727834 0.188676177351913
0.819762910450726 0.181796946284749
0.820233835173618 0.183425319041202
0.82070475989651 0.197436852299767
0.821175684619402 0.223235794420656
0.821646609342294 0.256610257777364
0.822117534065186 0.291978757429128
0.822588458788078 0.324287293267978
0.82305938351097 0.349792223464206
0.823530308233862 0.366453626429431
0.824001232956754 0.374685150326087
0.824472157679646 0.377807008464689
0.824943082402538 0.380915538289538
0.82541400712543 0.388272705921094
0.825884931848322 0.401162896390844
0.826355856571214 0.418088680219151
0.826826781294106 0.437054411479243
0.827297706016998 0.45779121885649
0.82776863073989 0.482087820684114
0.828239555462782 0.512302312322455
0.828710480185674 0.549411041955363
0.829181404908566 0.591716873964515
0.829652329631458 0.6347472759244
0.83012325435435 0.672764913891848
0.830594179077242 0.701840253425064
0.831065103800134 0.722844708661665
0.831536028523026 0.741483964334382
0.832006953245917 0.764108896054622
0.832477877968809 0.792199778519898
0.832948802691701 0.820767281518063
0.833419727414593 0.842824077598073
0.833890652137485 0.856128313973568
0.834361576860377 0.866204859711108
0.834832501583269 0.883593514566321
0.835303426306161 0.918436392105922
0.835774351029053 0.976445045924552
0.836245275751945 1.05761454706914
0.836716200474837 1.15697337476378
0.837187125197729 1.26673396320315
0.837658049920621 1.37963586584614
0.838128974643513 1.49252846996692
0.838599899366405 1.60804262711297
0.839070824089297 1.73243874276537
0.839541748812189 1.87022306857423
0.840012673535081 2.0192111383721
0.840483598257973 2.1703441901096
0.840954522980865 2.31341610123263
0.841425447703757 2.44497986797397
0.841896372426649 2.57211097651066
0.842367297149541 2.7083488078546
0.842838221872433 2.86465250616653
0.843309146595325 3.04278924984715
0.843780071318217 3.23639896490092
0.844250996041109 3.43814218103075
0.844721920764001 3.64670529095448
0.845192845486893 3.8684934136293
0.845663770209785 4.11300117569639
0.846134694932677 4.3847555507959
0.846605619655569 4.67744759766346
0.847076544378461 4.97599796856547
0.847547469101353 5.26738392250925
0.848018393824245 5.55310223379749
0.848489318547137 5.85297400357058
0.848960243270029 6.19626827369071
0.849431167992921 6.60611604200778
0.849902092715813 7.08746247590028
0.850373017438705 7.62512187098254
0.850843942161597 8.19171260640706
0.851314866884489 8.76017670371225
0.851785791607381 9.31445820394187
0.852256716330273 9.85472246445105
0.852727641053165 10.3971034602765
0.853198565776057 10.9689832392976
0.853669490498949 11.6009252444125
0.854140415221841 12.3188210576196
0.854611339944733 13.1409905725522
0.855082264667625 14.0795980849152
0.855553189390517 15.1396355514139
0.856024114113409 16.3127200153435
0.856495038836301 17.5731384000311
0.856965963559193 18.8842749197861
0.857436888282085 20.2125074323765
0.857907813004977 21.5382668237901
0.858378737727869 22.8587366220205
0.858849662450761 24.1842387463564
0.859320587173653 25.5316148202353
0.859791511896545 26.9165340450411
0.860262436619437 28.3476115962278
0.860733361342329 29.8253293868006
0.86120428606522 31.3448931045398
0.861675210788112 32.8984762016757
0.862146135511004 34.4740315926954
0.862617060233897 36.0534034438146
0.863087984956789 37.614278067515
0.86355890967968 39.1341170300583
0.864029834402572 40.587529911261
0.864500759125464 41.935346699525
0.864971683848356 43.1206833861074
0.865442608571248 44.0891668128622
0.86591353329414 44.824095269654
0.866384458017032 45.361455084437
0.866855382739924 45.7638349206225
0.867326307462816 46.0770759134558
0.867797232185708 46.3136465422007
0.8682681569086 46.4732575675744
0.868739081631492 46.5646178647528
0.869210006354384 46.5929288371131
0.869680931077276 46.526807414206
0.870151855800168 46.2925477787461
0.87062278052306 45.8156376454031
0.871093705245952 45.0721200250244
0.871564629968844 44.0984693432385
0.872035554691736 42.9551952637632
0.872506479414628 41.6875354258078
0.87297740413752 40.3214473279186
0.873448328860412 38.8886155507372
0.873919253583304 37.443889011898
0.874390178306196 36.049919767433
0.874861103029088 34.7392436812192
0.87533202775198 33.4881167668413
0.875802952474872 32.2270131434797
0.876273877197764 30.8788015450322
0.876744801920656 29.3918421276069
0.877215726643548 27.7465662439939
0.87768665136644 25.9446653681622
0.878157576089332 24.0026367869362
0.878628500812224 21.9560609378683
0.879099425535116 19.8628222549193
0.879570350258008 17.7940711797112
0.8800412749809 15.8151595069801
0.880512199703792 13.9684981454289
0.880983124426684 12.2690258272619
0.881454049149576 10.7132018068438
0.881924973872468 9.29209060963065
0.88239589859536 7.99803365529121
0.882866823318252 6.8239384496735
0.883337748041144 5.76241398644384
0.883808672764036 4.80864139800841
0.884279597486928 3.96253249729805
0.88475052220982 3.22542855207531
0.885221446932712 2.59414673302068
0.885692371655604 2.05907910499931
0.886163296378496 1.60824825721366
0.886634221101388 1.23260081282257
0.88710514582428 0.927114447940059
0.887576070547172 0.687227570010772
0.888046995270064 0.504791308739769
0.888517919992956 0.367689251275563
0.888988844715848 0.263240521442859
0.88945976943874 0.182020716125958
0.889930694161632 0.119053355576604
0.890401618884524 0.072241421775739
0.890872543607416 0.0400142569173348
0.891343468330308 0.019965809879068
0.8918143930532 0.00887700640181635
0.892285317776091 0.0034851058096037
0.892756242498983 0.00119921059299065
0.893227167221875 0.000359500743817186
};
\addlegendentry{Yaw}
\end{axis}

\end{tikzpicture}

%% file: images/comparison-frgc.tex
%
%
\definecolor{mycolor1}{rgb}{0.84314,0.48235,0.22353}%
\begin{tikzpicture}

\begin{axis}[%
width=1.8in,
height=1.35in,
at={(0.807in,0.648in)},
scale only axis,
xmin=0.102027626633644,
xmax=1.04201106548309,
xlabel style={font=\bfseries\color{white!4!black}},
xlabel={comparison scores: FRGC},
ymin=0,
ymax=0.1,
ylabel style={font=\bfseries\color{white!4!black}},
ylabel={Relative frequency},
axis background/.style={fill=white},
axis x line*=bottom,
axis y line*=left,
ytick={0, 0.02, 0.04, 0.06, 0.08, 0.1},
scaled ticks=false, tick label style={/pgf/number format/fixed},
]
\addplot [color=mycolor1, line width=2.0pt]
  table[row sep=crcr]{%
0.13254790186882	1.80867077332159e-08\\
0.141734398470985	1.19567704624417e-07\\
0.150920895073149	5.63205588525394e-07\\
0.160107391675313	1.98408268990042e-06\\
0.169293888277478	5.35875957746928e-06\\
0.178480384879642	1.13111781596128e-05\\
0.187666881481806	1.90194823060878e-05\\
0.196853378083971	2.60237638578397e-05\\
0.206039874686135	2.96648882947254e-05\\
0.2152263712883	2.88216789741495e-05\\
0.224412867890464	2.44066785406727e-05\\
0.233599364492628	1.87312269845483e-05\\
0.242785861094793	1.44556849026417e-05\\
0.251972357696957	1.33029963107337e-05\\
0.261158854299121	1.48812401063535e-05\\
0.270345350901286	1.68830766723429e-05\\
0.27953184750345	1.68955480453709e-05\\
0.288718344105615	1.41219397693029e-05\\
0.297904840707779	9.78696122975005e-06\\
0.307091337309943	6.01081498691058e-06\\
0.316277833912108	4.04864768704489e-06\\
0.325464330514272	3.82600404822391e-06\\
0.334650827116436	4.6672859949717e-06\\
0.343837323718601	5.91627235126272e-06\\
0.353023820320765	7.59615229666524e-06\\
0.36221031692293	1.08254187815929e-05\\
0.371396813525094	1.72241575666201e-05\\
0.380583310127258	2.84514410432813e-05\\
0.389769806729423	4.73493858511803e-05\\
0.398956303331587	7.86626858551669e-05\\
0.408142799933751	0.000126931960144311\\
0.417329296535916	0.000193677199293664\\
0.42651579313808	0.000278789185852961\\
0.435702289740244	0.000385201765512375\\
0.444888786342409	0.000519326372748102\\
0.454075282944573	0.000684532723371285\\
0.463261779546738	0.000878351642803895\\
0.472448276148902	0.00110270708582465\\
0.481634772751066	0.00137574947021904\\
0.490821269353231	0.00172955637836497\\
0.500007765955395	0.00219728982305808\\
0.50919426255756	0.0027992061848489\\
0.518380759159724	0.00353820024019267\\
0.527567255761888	0.00441389008902725\\
0.536753752364053	0.00544112776162741\\
0.545940248966217	0.00664903321606398\\
0.555126745568381	0.00806573808491802\\
0.564313242170546	0.0097047493091758\\
0.57349973877271	0.0115444263593799\\
0.582686235374875	0.0135146118284665\\
0.591872731977039	0.0155344490693045\\
0.601059228579203	0.0175541472768198\\
0.610245725181368	0.0195358495303032\\
0.619432221783532	0.0214324861765164\\
0.628618718385696	0.0232096935924443\\
0.637805214987861	0.0248699179741052\\
0.646991711590025	0.026461871336435\\
0.656178208192189	0.0280610013782979\\
0.665364704794354	0.0297134672792587\\
0.674551201396518	0.0314039120525604\\
0.683737697998683	0.0330919251447803\\
0.692924194600847	0.034761498058164\\
0.702110691203011	0.0364059302716345\\
0.711297187805176	0.0379868014059682\\
0.72048368440734	0.0394435586276588\\
0.729670181009505	0.0407140927785753\\
0.738856677611669	0.0417502223145292\\
0.748043174213833	0.0425606153170427\\
0.757229670815998	0.0432115389417628\\
0.766416167418162	0.0437518849829933\\
0.775602664020326	0.0441901689498824\\
0.784789160622491	0.0445709130610831\\
0.793975657224655	0.0450085911198744\\
0.803162153826819	0.0456129955283707\\
0.812348650428984	0.0464153188530791\\
0.821535147031148	0.0473852721344399\\
0.830721643633312	0.0485042446152999\\
0.839908140235477	0.0498050967113811\\
0.849094636837641	0.0513441248882782\\
0.858281133439806	0.0531502993803608\\
0.86746763004197	0.0551729558003428\\
0.876654126644135	0.0572494320901529\\
0.885840623246299	0.0591347842463108\\
0.895027119848463	0.0605504733891388\\
0.904213616450628	0.061180453852268\\
0.913400113052792	0.060660010030607\\
0.922586609654956	0.0585977818813717\\
0.931773106257121	0.0545984485842849\\
0.940959602859285	0.0483696014730881\\
0.950146099461449	0.0400353026265877\\
0.959332596063614	0.0304198100734322\\
0.968519092665778	0.0208868967006419\\
0.977705589267942	0.0127603505798418\\
0.986892085870107	0.00680295186770182\\
0.996078582472271	0.00308156055045969\\
1.00526507907444	0.00114690434774567\\
1.0144515756766	0.000338260635266782\\
1.02363807227876	7.62775160598485e-05\\
1.03282456888093	1.26015714368865e-05\\
1.04201106548309	1.38038738550051e-06\\
};

\addplot [color=mycolor1, dotted, line width=2.0pt]
  table[row sep=crcr]{%
0.102027626633644	4.07473764242781e-06\\
0.10845705394793	1.65459725870237e-05\\
0.114886481262217	5.52680285819363e-05\\
0.121315908576503	0.000158953952395138\\
0.127745335890789	0.00039999941971517\\
0.134174763205076	0.000893157832338421\\
0.140604190519362	0.00178556875154553\\
0.147033617833648	0.00323276544425229\\
0.153463045147934	0.00537154028978351\\
0.159892472462221	0.0083094297070263\\
0.166321899776507	0.0121438468957362\\
0.172751327090793	0.0169763304223028\\
0.179180754405079	0.0228752042440318\\
0.185610181719366	0.0298221307881009\\
0.192039609033652	0.0376865095407315\\
0.198469036347938	0.046244564161379\\
0.204898463662225	0.0551783466008955\\
0.211327890976511	0.0640522988691708\\
0.217757318290797	0.0723642726230041\\
0.224186745605083	0.0796776874067055\\
0.23061617291937	0.0857270910682427\\
0.237045600233656	0.0904284856538138\\
0.243475027547942	0.093812627820556\\
0.249904454862229	0.095954392264661\\
0.256333882176515	0.0969254062369674\\
0.262763309490801	0.0967961661198115\\
0.269192736805087	0.0956760696083536\\
0.275622164119374	0.0937180947742186\\
0.28205159143366	0.0910495703207747\\
0.288481018747946	0.0877346943519311\\
0.294910446062233	0.0838486680865019\\
0.301339873376519	0.0795936963630177\\
0.307769300690805	0.0753061969950723\\
0.314198728005091	0.0713186361207458\\
0.320628155319378	0.067770339696423\\
0.327057582633664	0.0645405552615811\\
0.33348700994795	0.0613552032371797\\
0.339916437262236	0.0579960876239801\\
0.346345864576523	0.0544441212746764\\
0.352775291890809	0.0508362652201355\\
0.359204719205095	0.0473137435708635\\
0.365634146519382	0.0439106303868395\\
0.372063573833668	0.0405686328864506\\
0.378493001147954	0.0372420020192919\\
0.384922428462241	0.0339809914732552\\
0.391351855776527	0.0308980060660996\\
0.397781283090813	0.0280840457997835\\
0.404210710405099	0.0255762252548429\\
0.410640137719386	0.0234107822428338\\
0.417069565033672	0.0216392601710393\\
0.423498992347958	0.0202538907025046\\
0.429928419662244	0.0191060266395324\\
0.436357846976531	0.0179527564674298\\
0.442787274290817	0.0166007484227349\\
0.449216701605103	0.0150056105793863\\
0.45564612891939	0.013257770167522\\
0.462075556233676	0.0115131587239672\\
0.468504983547962	0.00993484018118427\\
0.474934410862248	0.00864994726696779\\
0.481363838176535	0.00770071150889047\\
0.487793265490821	0.00702690032833316\\
0.494222692805107	0.00651092280356269\\
0.500652120119394	0.00606174095910844\\
0.50708154743368	0.00565775407982712\\
0.513510974747966	0.00530824081353183\\
0.519940402062252	0.00498809985006538\\
0.526369829376539	0.00462218825151618\\
0.532799256690825	0.00414278934026127\\
0.539228684005111	0.00355645233888944\\
0.545658111319397	0.00295013794630856\\
0.552087538633684	0.0024312400057553\\
0.55851696594797	0.00205692731775974\\
0.564946393262256	0.00180791313621616\\
0.571375820576543	0.00161699635473645\\
0.577805247890829	0.00142332979638808\\
0.584234675205115	0.00120793516447558\\
0.590664102519401	0.000990627968001919\\
0.597093529833688	0.000800041442293217\\
0.603522957147974	0.00064893449648372\\
0.60995238446226	0.00053085431098173\\
0.616381811776546	0.000432673932478047\\
0.622811239090833	0.000346031526562941\\
0.629240666405119	0.000270371765991332\\
0.635670093719405	0.000209534213731283\\
0.642099521033692	0.000167617174362276\\
0.648528948347978	0.000145903154408583\\
0.654958375662264	0.000141883521220642\\
0.661387802976551	0.00015002328416757\\
0.667817230290837	0.000164489204879958\\
0.674246657605123	0.00018001288786253\\
0.680676084919409	0.000190240530985224\\
0.687105512233696	0.00018750419483128\\
0.693534939547982	0.000166692100690626\\
0.699964366862268	0.000130321470918773\\
0.706393794176554	8.81195627963134e-05\\
0.712823221490841	5.09795554607208e-05\\
0.719252648805127	2.50597472591075e-05\\
0.725682076119413	1.04058330595317e-05\\
0.7321115034337	3.6461181603612e-06\\
0.738540930747986	1.07298590625255e-06\\
};

\addplot [color=black, dashed]
  table[row sep=crcr]{%
0.55	0\\
0.55	0.1\\
};

\end{axis}
\end{tikzpicture}%

%% file: images/comparison-frpca.tex
%
%
\definecolor{mycolor1}{rgb}{0.00000,1.00000,1.00000}%
\begin{tikzpicture}

\begin{axis}[%
width=1.8in,
height=1.35in,
at={(0.807in,0.648in)},
scale only axis,
xmin=0.0933318454027176,
xmax=1.04195599079132,
xlabel style={font=\bfseries\color{white!4!black}},
xlabel={comparison scores: FR-PCA},
ymin=0,
ymax=0.20,
axis background/.style={fill=white},
axis x line*=bottom,
axis y line*=left,
ytick={0, 0.05, 0.1, 0.15, 0.20},
scaled ticks=false, tick label style={/pgf/number format/fixed},
]
\addplot [color=mycolor1, line width=2.0pt]
  table[row sep=crcr]{%
0.174525083899498	1.06145241261532e-06\\
0.183287012251941	5.36934615633864e-06\\
0.192048940604383	1.96773813605887e-05\\
0.200810868956826	5.2620232973017e-05\\
0.209572797309269	0.000103507961933262\\
0.218334725661711	0.000150914204461615\\
0.227096654014154	0.000163948266410917\\
0.235858582366597	0.000132879593065569\\
0.244620510719039	8.00971141872013e-05\\
0.253382439071482	3.56740741538533e-05\\
0.262144367423925	1.16463372270496e-05\\
0.270906295776367	2.76532014212606e-06\\
0.27966822412881	4.45799102001104e-07\\
0.288430152481252	5.61936601539034e-08\\
0.297192080833695	6.15522445080616e-69\\
0.305954009186138	1.19045948733177e-64\\
0.31471593753858	1.63682175860999e-60\\
0.323477865891023	1.599941861002e-56\\
0.332239794243466	1.11179074927576e-52\\
0.341001722595908	5.49234682246821e-49\\
0.349763650948351	1.9288963614034e-45\\
0.358525579300794	4.8158824402842e-42\\
0.367287507653236	8.54788956643966e-39\\
0.376049436005679	1.07859392858165e-35\\
0.384811364358122	9.67549254620188e-33\\
0.393573292710564	6.17027097283189e-30\\
0.402335221063007	2.79737979571963e-27\\
0.41109714941545	9.01601599647874e-25\\
0.419859077767892	2.06582703247634e-22\\
0.428621006120335	3.36503264224323e-20\\
0.437382934472778	3.89673300093988e-18\\
0.44614486282522	3.20795347003617e-16\\
0.454906791177663	1.8774635496748e-14\\
0.463668719530106	7.81143836777502e-13\\
0.472430647882548	2.31050306927498e-11\\
0.481192576234991	4.85845440974759e-10\\
0.489954504587434	7.2628279563522e-09\\
0.498716432939876	7.72170861149122e-08\\
0.507478361292319	5.83797721976843e-07\\
0.516240289644761	3.14156844941862e-06\\
0.525002217997204	1.20580298262979e-05\\
0.533764146349647	3.31907316766802e-05\\
0.542526074702089	6.64859998718807e-05\\
0.551288003054532	0.000100569310821801\\
0.560049931406975	0.000124528526906039\\
0.568811859759417	0.000142635017623953\\
0.57757378811186	0.000163851174318453\\
0.586335716464303	0.000184311513506933\\
0.595097644816745	0.000192295286069749\\
0.603859573169188	0.000180040485737909\\
0.612621501521631	0.000147141629266262\\
0.621383429874073	0.000105559940461095\\
0.630145358226516	8.32787830667488e-05\\
0.638907286578959	0.000111923498000614\\
0.647669214931401	0.000206468463059222\\
0.656431143283844	0.000357042093989354\\
0.665193071636287	0.000546480045085882\\
0.673954999988729	0.000774440665127114\\
0.682716928341172	0.00103224666783702\\
0.691478856693615	0.00124883956750344\\
0.700240785046057	0.0013331057165154\\
0.7090027133985	0.00129202550000092\\
0.717764641750942	0.00124653578968723\\
0.726526570103385	0.00134558538141381\\
0.735288498455828	0.00169674431408086\\
0.744050426808271	0.00229746902011811\\
0.752812355160713	0.00302376269227208\\
0.761574283513156	0.0037865698556677\\
0.770336211865599	0.00467241965038078\\
0.779098140218041	0.00588140671258383\\
0.787860068570484	0.00760448674686092\\
0.796621996922926	0.00999094977689844\\
0.805383925275369	0.0131721190849382\\
0.814145853627812	0.0173112826576899\\
0.822907781980254	0.0226770768594582\\
0.831669710332697	0.029680983420365\\
0.84043163868514	0.0388342485437071\\
0.849193567037582	0.050577213544156\\
0.857955495390025	0.065048525355132\\
0.866717423742468	0.0818541319443962\\
0.87547935209491	0.0999128670766173\\
0.884241280447353	0.117534032464033\\
0.893003208799796	0.132935305313561\\
0.901765137152238	0.144957297983467\\
0.910527065504681	0.152899563797255\\
0.919288993857124	0.155406749375378\\
0.928050922209566	0.150645606162596\\
0.936812850562009	0.138311843807454\\
0.945574778914452	0.120507162219587\\
0.954336707266894	0.100133830513888\\
0.963098635619337	0.0790941820690454\\
0.971860563971779	0.0583449136777647\\
0.980622492324222	0.0391158776795488\\
0.989384420676665	0.0231748013520973\\
0.998146349029107	0.0118124328634835\\
1.00690827738155	0.00503400401273596\\
1.01567020573399	0.00173894978306094\\
1.02443213408644	0.00047187166305581\\
1.03319406243888	9.71019127511642e-05\\
1.04195599079132	1.44649008808025e-05\\
};

\addplot [color=mycolor1, dotted, line width=2.0pt]
  table[row sep=crcr]{%
0.0933318454027176	2.96418687301364e-06\\
0.101447424298585	1.5063873535527e-05\\
0.109563003194453	5.90776960252164e-05\\
0.11767858209032	0.000180864582153475\\
0.125794160986187	0.000444700521147644\\
0.133909739882055	0.000906933751071341\\
0.142025318777922	0.00161010995719237\\
0.15014089767379	0.00262843238369268\\
0.158256476569657	0.00412876357414306\\
0.166372055465525	0.00633640833986518\\
0.174487634361392	0.00937163026292068\\
0.18260321325726	0.0131059772575654\\
0.190718792153127	0.017275925402533\\
0.198834371048995	0.0218006645399154\\
0.206949949944862	0.0269402197377444\\
0.21506552884073	0.0330943996563936\\
0.223181107736597	0.040396245628802\\
0.231296686632465	0.0483808428756082\\
0.239412265528332	0.0560357939215258\\
0.2475278444242	0.0623880109894821\\
0.255643423320067	0.0672051786228274\\
0.263759002215935	0.0710830616316637\\
0.271874581111802	0.0747799558516622\\
0.279990160007669	0.0784723895422345\\
0.288105738903537	0.0815986513322083\\
0.296221317799404	0.0832893703369319\\
0.304336896695272	0.083017939922767\\
0.312452475591139	0.0810534705431736\\
0.320568054487007	0.0782704911610694\\
0.328683633382874	0.0753887966902232\\
0.336799212278742	0.072499177931734\\
0.344914791174609	0.0694159876261901\\
0.353030370070477	0.0662217622152855\\
0.361145948966344	0.0631188265158283\\
0.369261527862212	0.0599144232154132\\
0.377377106758079	0.0561537050603825\\
0.385492685653947	0.0518156196400009\\
0.393608264549814	0.04745087532469\\
0.401723843445682	0.043535620894988\\
0.409839422341549	0.0400069902342887\\
0.417955001237417	0.0365387507819161\\
0.426070580133284	0.0330926750584194\\
0.434186159029151	0.0299904610709748\\
0.442301737925019	0.0274470241841745\\
0.450417316820886	0.025232156470639\\
0.458532895716754	0.0229325069976547\\
0.466648474612621	0.0204253665707854\\
0.474764053508489	0.0179854339614561\\
0.482879632404356	0.0159905067404209\\
0.490995211300224	0.014591002883341\\
0.499110790196091	0.0136093155402602\\
0.507226369091959	0.01268449051227\\
0.515341947987826	0.0115349374416183\\
0.523457526883694	0.0101553091347658\\
0.531573105779561	0.00876842636487525\\
0.539688684675429	0.00755834887416075\\
0.547804263571296	0.00652936193466643\\
0.555919842467164	0.0056414017471638\\
0.564035421363031	0.00492837022226609\\
0.572151000258899	0.00440023262813292\\
0.580266579154766	0.00397290328549939\\
0.588382158050633	0.00357816941613904\\
0.596497736946501	0.00323561670494948\\
0.604613315842368	0.00294938973508481\\
0.612728894738236	0.00263536222804709\\
0.620844473634103	0.00222469117302917\\
0.628960052529971	0.00177439785134908\\
0.637075631425838	0.00141372154221\\
0.645191210321706	0.00120954515544646\\
0.653306789217573	0.00112139737353343\\
0.661422368113441	0.00106793153823323\\
0.669537947009308	0.00100030689849026\\
0.677653525905176	0.000909060835989459\\
0.685769104801043	0.000800041378316299\\
0.693884683696911	0.000692193242739305\\
0.702000262592778	0.000618370661046399\\
0.710115841488645	0.000595173922766607\\
0.718231420384513	0.000594916015022689\\
0.72634699928038	0.000564315085736966\\
0.734462578176248	0.000475613129361391\\
0.742578157072115	0.000354783102802632\\
0.750693735967983	0.000259736487828819\\
0.75880931486385	0.000231927525665715\\
0.766924893759718	0.000265330311743329\\
0.775040472655585	0.000311607217409972\\
0.783156051551453	0.000314544480636751\\
0.79127163044732	0.000253370645259464\\
0.799387209343188	0.000158140061820663\\
0.807502788239055	7.61863027355617e-05\\
0.815618367134923	3.18075312833702e-05\\
0.82373394603079	2.27294980895852e-05\\
0.831849524926658	3.81598209690053e-05\\
0.839965103822525	6.52528974027723e-05\\
0.848080682718393	8.60066276376581e-05\\
0.85619626161426	8.48838418268743e-05\\
0.864311840510128	6.25158667342419e-05\\
0.872427419405995	3.43579787673294e-05\\
0.880542998301862	1.40908463142367e-05\\
0.88865857719773	4.31239635560492e-06\\
0.896774156093597	9.84855202652959e-07\\
};

\addplot [color=black, dashed]
  table[row sep=crcr]{%
0.55	0\\
0.55	0.2\\
};

\end{axis}
\end{tikzpicture}%

%% file: images/comparison-base.tex
%
%
\definecolor{mycolor1}{rgb}{0.12549,0.46667,0.12549}%
\begin{tikzpicture}

\begin{axis}[%
width=1.8in,
height=1.35in,
at={(0.807in,0.648in)},
scale only axis,
xmin=0.0992306935787201,
xmax=1,
xlabel style={font=\bfseries\color{white!4!black}},
xlabel={comparison scores: Base Images},
ymin=0,
ymax=0.1,
ylabel style={font=\bfseries\color{white!4!black}},
ylabel={Relative frequency},
axis background/.style={fill=white},
axis x line*=bottom,
axis y line*=left,
ytick={0, 0.02, 0.04, 0.06, 0.08, 0.1},
scaled ticks=false, tick label style={/pgf/number format/fixed},
]
\addplot [color=mycolor1, dashed, line width=2.0pt]
  table[row sep=crcr]{%
0.0992306935787201	2.94291904893874e-06\\
0.106924717570796	1.44189563719775e-05\\
0.114618741562872	5.5953741366346e-05\\
0.122312765554948	0.000172883284747592\\
0.130006789547024	0.000434576772543706\\
0.1377008135391	0.000916869152001169\\
0.145394837531176	0.00168065490988131\\
0.153088861523253	0.00277715232570527\\
0.160782885515329	0.00425681207300455\\
0.168476909507405	0.00617773792558282\\
0.176170933499481	0.00864904683874214\\
0.183864957491557	0.0118567169803139\\
0.191558981483633	0.0159393805787175\\
0.199253005475709	0.0208080114089738\\
0.206947029467785	0.0262389062438562\\
0.214641053459861	0.0321485773427981\\
0.222335077451937	0.0386072839025337\\
0.230029101444013	0.0455453694764833\\
0.237723125436089	0.0526639877626697\\
0.245417149428165	0.0597023281230024\\
0.253111173420241	0.0665377548619538\\
0.260805197412318	0.0728416088386829\\
0.268499221404394	0.0779385240847026\\
0.27619324539647	0.0813680467189831\\
0.283887269388546	0.0833475761862085\\
0.291581293380622	0.0843305822918698\\
0.299275317372698	0.0843769618755492\\
0.306969341364774	0.0833764681026433\\
0.31466336535685	0.0816789380089841\\
0.322357389348926	0.080023844128907\\
0.330051413341002	0.0787731014786222\\
0.337745437333078	0.0773749537330921\\
0.345439461325154	0.0748025367511007\\
0.35313348531723	0.0706021392158933\\
0.360827509309306	0.065365669023648\\
0.368521533301382	0.0600756087156849\\
0.376215557293458	0.0552261893495358\\
0.383909581285535	0.0507495793233661\\
0.391603605277611	0.0465207290500615\\
0.399297629269687	0.042638776783629\\
0.406991653261763	0.039269067689431\\
0.414685677253839	0.0364241832095569\\
0.422379701245915	0.0339556444216069\\
0.430073725237991	0.0316681310375537\\
0.437767749230067	0.0293754884302043\\
0.445461773222143	0.0269227735638744\\
0.453155797214219	0.0242954236515925\\
0.460849821206295	0.0216619601400888\\
0.468543845198371	0.019233049630209\\
0.476237869190447	0.0170976144013612\\
0.483931893182523	0.0152485000922989\\
0.491625917174599	0.0137015759619664\\
0.499319941166676	0.0124954722127922\\
0.507013965158752	0.0115821573732384\\
0.514707989150828	0.0107840693441149\\
0.522402013142904	0.00988649223353244\\
0.53009603713498	0.00877369278858493\\
0.537790061127056	0.00749239475624171\\
0.545484085119132	0.00620316692965531\\
0.553178109111208	0.00506948199977103\\
0.560872133103284	0.00417710902158341\\
0.56856615709536	0.0035278398779611\\
0.576260181087436	0.00307135964606733\\
0.583954205079512	0.00273977669620614\\
0.591648229071588	0.00247881743744832\\
0.599342253063664	0.00226728823447976\\
0.60703627705574	0.002114828741649\\
0.614730301047816	0.00203676979844988\\
0.622424325039893	0.002017830783172\\
0.630118349031969	0.00199983868900372\\
0.637812373024045	0.00192263781658925\\
0.645506397016121	0.00177560828269523\\
0.653200421008197	0.0015913468883208\\
0.660894445000273	0.00138992038318728\\
0.668588468992349	0.00115730892822686\\
0.676282492984425	0.000886454224777223\\
0.683976516976501	0.000613893928009685\\
0.691670540968577	0.000396173531414133\\
0.699364564960653	0.000263622989372592\\
0.707058588952729	0.000208630294460208\\
0.714752612944805	0.000203166750849092\\
0.722446636936881	0.000215684915987293\\
0.730140660928957	0.000220101467773247\\
0.737834684921033	0.000203124439020926\\
0.74552870891311	0.000165012302113529\\
0.753222732905186	0.000115301954509893\\
0.760916756897262	6.75138702052616e-05\\
0.768610780889338	3.27449541563333e-05\\
0.776304804881414	1.49249541954883e-05\\
0.78399882887349	1.28319678009532e-05\\
0.791692852865566	2.48353480233794e-05\\
0.799386876857642	4.83777619047749e-05\\
0.807080900849718	7.44678827390866e-05\\
0.814774924841794	8.83872496260396e-05\\
0.82246894883387	8.06392315534493e-05\\
0.830162972825946	5.65508696193562e-05\\
0.837856996818022	3.04837465153204e-05\\
0.845551020810098	1.26308780540481e-05\\
0.853245044802174	4.02285972768882e-06\\
0.860939068794251	9.84855202652959e-07\\
};

\addplot [color=black, dashed]
  table[row sep=crcr]{%
0.55	0\\
0.55	0.1\\
};

\end{axis}
\end{tikzpicture}%

%% file: images/comparison-illumination-quality.tex
%
%
\definecolor{mycolor1}{rgb}{0.53333,0.00000,0.00000}%
\begin{tikzpicture}

\begin{axis}[%
width=1.8in,
height=1.35in,
at={(0.807in,0.648in)},
scale only axis,
xmin=0.103162037134171,
xmax=1.04102776765823,
xlabel style={font=\bfseries\color{white!4!black}},
xlabel={comparison scores: Illumination},
ymin=0,
ymax=0.20,
axis background/.style={fill=white},
axis x line*=bottom,
axis y line*=left,
ytick={0, 0.05, 0.1, 0.15, 0.20},
scaled ticks=false, tick label style={/pgf/number format/fixed},
]
\addplot [color=mycolor1, line width=2.0pt]
  table[row sep=crcr]{%
0.180484162569046	1.91472751879162e-06\\
0.189176522216412	9.25838646490083e-06\\
0.197868881863777	3.1999024369865e-05\\
0.206561241511143	7.90517116668951e-05\\
0.215253601158508	0.00013959148575047\\
0.223945960805874	0.000176189366444302\\
0.232638320453239	0.000158954772209952\\
0.241330680100605	0.000102503940456629\\
0.25002303974797	4.72476989414408e-05\\
0.258715399395336	1.55666038595465e-05\\
0.267407759042701	3.6658964378818e-06\\
0.276100118690067	6.17077361403579e-07\\
0.284792478337432	7.42458416989859e-08\\
0.293484837984798	4.68068024499083e-16\\
0.302177197632163	2.57666022121669e-14\\
0.310869557279529	1.01383005099335e-12\\
0.319561916926895	2.85123488132504e-11\\
0.32825427657426	5.73139920459481e-10\\
0.336946636221626	8.23470699678919e-09\\
0.345638995868991	8.45659383399404e-08\\
0.354331355516357	6.20729846990297e-07\\
0.363023715163722	3.25663936465117e-06\\
0.371716074811088	1.22122862694444e-05\\
0.380408434458453	3.27331359651864e-05\\
0.389100794105819	6.27142096668674e-05\\
0.397793153753184	8.59262537843966e-05\\
0.40648551340055	8.44900216155444e-05\\
0.415177873047915	6.12818627053515e-05\\
0.423870232695281	3.93013560972952e-05\\
0.432562592342646	3.86774512657121e-05\\
0.441254951990012	5.99593067990042e-05\\
0.449947311637377	8.37252394725946e-05\\
0.458639671284743	8.65488326028e-05\\
0.467332030932109	6.44568746393895e-05\\
0.476024390579474	3.57197068635685e-05\\
0.48471675022684	2.0153019429353e-05\\
0.493409109874205	2.61830092249927e-05\\
0.502101469521571	5.23659581187866e-05\\
0.510793829168936	8.77578174336748e-05\\
0.519486188816302	0.00011617883330041\\
0.528178548463667	0.000130202335223299\\
0.536870908111033	0.000134209470799695\\
0.545563267758398	0.0001353213935351\\
0.554255627405764	0.000143202864817044\\
0.562947987053129	0.000168464821285141\\
0.571640346700495	0.00020922566530062\\
0.58033270634786	0.000251224815073408\\
0.589025065995226	0.000289791381417162\\
0.597717425642591	0.000343136441899947\\
0.606409785289957	0.000421521223203257\\
0.615102144937323	0.000485766497692624\\
0.623794504584688	0.000480757423835688\\
0.632486864232054	0.000418577603839551\\
0.641179223879419	0.000387098918648218\\
0.649871583526785	0.000466537688677729\\
0.65856394317415	0.000651058155358515\\
0.667256302821516	0.000863047120400403\\
0.675948662468881	0.00104078388053581\\
0.684641022116247	0.00118124105481627\\
0.693333381763612	0.00129719188806742\\
0.702025741410978	0.00138999109763082\\
0.710718101058343	0.00148076139896463\\
0.719410460705709	0.00163012125723689\\
0.728102820353074	0.00191167867358408\\
0.73679518000044	0.00235112533818439\\
0.745487539647806	0.00289135779968698\\
0.754179899295171	0.00343974807536705\\
0.762872258942537	0.00394670760585642\\
0.771564618589902	0.00442069200049107\\
0.780256978237268	0.00490222133604914\\
0.788949337884633	0.00547336626340469\\
0.797641697531999	0.00627535222595411\\
0.806334057179364	0.00745266990684612\\
0.81502641682673	0.00911384636691756\\
0.823718776474095	0.0113880137992807\\
0.832411136121461	0.0143844706711224\\
0.841103495768826	0.0180084729071226\\
0.849795855416192	0.0220265795907153\\
0.858488215063557	0.026488146346299\\
0.867180574710923	0.0320525165875201\\
0.875872934358289	0.039879093581082\\
0.884565294005654	0.0510477600655409\\
0.89325765365302	0.0660147735241136\\
0.901950013300385	0.0847385051638957\\
0.910642372947751	0.107109352014637\\
0.919334732595116	0.132634660113202\\
0.928027092242482	0.159278849093922\\
0.936719451889847	0.18260698037246\\
0.945411811537213	0.196406528548518\\
0.954104171184578	0.194845441495369\\
0.962796530831944	0.175158311648422\\
0.971488890479309	0.139778153112785\\
0.980181250126675	0.0966874863913174\\
0.98887360977404	0.0565370484181837\\
0.997565969421406	0.0272914695659514\\
1.00625832906877	0.0106505870056348\\
1.01495068871614	0.00329963053786908\\
1.0236430483635	0.000796259000354798\\
1.03233540801087	0.000144934693918297\\
1.04102776765823	1.81857953293507e-05\\
};

\addplot [color=mycolor1, dotted, line width=2.0pt]
  table[row sep=crcr]{%
0.103162037134171	1.98556063755611e-06\\
0.110871292822289	1.01090884770743e-05\\
0.118580548510407	4.10380315530157e-05\\
0.126289804198525	0.000134572449873951\\
0.133999059886643	0.000368665386238449\\
0.141708315574762	0.000856176962532042\\
0.14941757126288	0.00171338419154919\\
0.157126826950998	0.00300421491093275\\
0.164836082639116	0.00472143483314977\\
0.172545338327234	0.00685552511618946\\
0.180254594015353	0.00951344902939897\\
0.187963849703471	0.0129237378801619\\
0.195673105391589	0.0172503563898423\\
0.203382361079707	0.022351193126638\\
0.211091616767826	0.0277716832530115\\
0.218800872455944	0.0330972249712423\\
0.226510128144062	0.0383114055008558\\
0.23421938383218	0.0437221804953654\\
0.241928639520298	0.0495621371439674\\
0.249637895208417	0.0556773671754274\\
0.257347150896535	0.0615590419541154\\
0.265056406584653	0.0667004629800546\\
0.272765662272771	0.0709746603083532\\
0.280474917960889	0.0745480142836339\\
0.288184173649008	0.0775017893491084\\
0.295893429337126	0.0797787027625014\\
0.303602685025244	0.0813729573531107\\
0.311311940713362	0.0821453184129121\\
0.319021196401481	0.0815854762070494\\
0.326730452089599	0.0792284217739853\\
0.334439707777717	0.0754357230725728\\
0.342148963465835	0.0714475137280097\\
0.349858219153953	0.0684706695469644\\
0.357567474842072	0.0666528210365562\\
0.36527673053019	0.0649963388194942\\
0.372985986218308	0.0623415720644461\\
0.380695241906426	0.0584301886838579\\
0.388404497594544	0.0539638758154751\\
0.396113753282663	0.0497985049547328\\
0.403823008970781	0.0462837905105026\\
0.411532264658899	0.0432898686516545\\
0.419241520347017	0.0405158473697126\\
0.426950776035135	0.0377009656441596\\
0.434660031723254	0.034740886523925\\
0.442369287411372	0.0317683216057058\\
0.45007854309949	0.0290243812965754\\
0.457787798787608	0.0266088837738338\\
0.465497054475727	0.0243914035172218\\
0.473206310163845	0.0221518750391944\\
0.480915565851963	0.0197817360899776\\
0.488624821540081	0.0173858666514331\\
0.496334077228199	0.0152072120448333\\
0.504043332916318	0.0134260465147232\\
0.511752588604436	0.012021275591936\\
0.519461844292554	0.0108258583658885\\
0.527171099980672	0.00969115817992039\\
0.53488035566879	0.00859434248764715\\
0.542589611356909	0.00759291722398044\\
0.550298867045027	0.00670587756490152\\
0.558008122733145	0.00587662831664476\\
0.565717378421263	0.00505128711707606\\
0.573426634109381	0.00426026104694232\\
0.5811358897975	0.00360122515448702\\
0.588845145485618	0.00314003868773407\\
0.596554401173736	0.00284582729425016\\
0.604263656861854	0.00263964289030574\\
0.611972912549973	0.00248361344225025\\
0.619682168238091	0.00237903057292658\\
0.627391423926209	0.00229744799052908\\
0.635100679614327	0.00215908277610025\\
0.642809935302445	0.00189560082514138\\
0.650519190990564	0.00152498358703468\\
0.658228446678682	0.0011516507278711\\
0.6659377023668	0.000880864176159134\\
0.673646958054918	0.000737537626328778\\
0.681356213743036	0.000672014904075399\\
0.689065469431155	0.000630381735942893\\
0.696774725119273	0.000594480332635787\\
0.704483980807391	0.000566051292378729\\
0.712193236495509	0.00054428382153436\\
0.719902492183627	0.000525881332725528\\
0.727611747871746	0.000507951694434868\\
0.735321003559864	0.000482512958645939\\
0.743030259247982	0.000436369138197098\\
0.7507395149361	0.000361205357891414\\
0.758448770624219	0.000263248997257452\\
0.766158026312337	0.000163386010420334\\
0.773867282000455	8.4407422796111e-05\\
0.781576537688573	3.75047058482242e-05\\
0.789285793376691	2.05298839221049e-05\\
0.79699504906481	2.66771359153915e-05\\
0.804704304752928	4.83995993860593e-05\\
0.812413560441046	7.42358919837739e-05\\
0.820122816129164	8.83438258578616e-05\\
0.827832071817282	8.08166196828745e-05\\
0.835541327505401	5.67686343578624e-05\\
0.843250583193519	3.06195990749696e-05\\
0.850959838881637	1.26815938577792e-05\\
0.858669094569755	4.03303014513295e-06\\
0.866378350257874	9.84855202652959e-07\\
};

\addplot [color=black, dashed]
  table[row sep=crcr]{%
0.55	0\\
0.55	0.2\\
};

\end{axis}
\end{tikzpicture}%

%% file: images/comparison-yaw.tex
%
%
\definecolor{mycolor1}{rgb}{0.65490,0.17647,0.85882}%
\begin{tikzpicture}

\begin{axis}[%
width=1.8in,
height=1.35in,
at={(0.807in,0.648in)},
scale only axis,
xmin=0.0951054710149765,
xmax=1.041850669384,
xlabel style={font=\bfseries\color{white!4!black}},
xlabel={comparison scores: Yaw},
ymin=0,
ymax=0.1,
ylabel style={font=\bfseries\color{white!4!black}},
ylabel={Relative frequency},
axis background/.style={fill=white},
axis x line*=bottom,
axis y line*=left,
ytick={0, 0.02, 0.04, 0.06, 0.08, 0.1},
scaled ticks=false, tick label style={/pgf/number format/fixed},
]
\addplot [color=mycolor1, line width=2.0pt]
  table[row sep=crcr]{%
0.179281042814255	1.44666885026415e-06\\
0.18799386732506	7.31784475940132e-06\\
0.196706691835866	2.65358114667789e-05\\
0.205419516346671	6.89908597514596e-05\\
0.214132340857477	0.000128623039523196\\
0.222845165368282	0.000171970132991348\\
0.231557989879088	0.000164897039708334\\
0.240270814389894	0.000113395435171103\\
0.248983638900699	5.59213132151589e-05\\
0.257696463411505	1.97748972821427e-05\\
0.26640928792231	5.01353080444131e-06\\
0.275122112433116	9.11143900089156e-07\\
0.283834936943921	1.1867297644139e-07\\
0.292547761454727	5.85357068664812e-16\\
0.301260585965532	3.19338331649963e-14\\
0.309973410476338	1.24323588981794e-12\\
0.318686234987143	3.45404767527674e-11\\
0.327399059497949	6.84818396714231e-10\\
0.336111884008754	9.68935373819515e-09\\
0.34482470851956	1.09681095066473e-07\\
0.353537533030365	8.21319455082063e-07\\
0.362250357541171	4.4406644613397e-06\\
0.370963182051977	1.73826311621401e-05\\
0.379676006562782	4.94019005517049e-05\\
0.388388831073588	0.000102211432399723\\
0.397101655584393	0.000154291990577216\\
0.405814480095199	0.000170170755032084\\
0.414527304606004	0.000137182325008331\\
0.42324012911681	8.08410767203878e-05\\
0.431952953627615	3.52842152111135e-05\\
0.440665778138421	1.410528224252e-05\\
0.449378602649226	1.50468708228093e-05\\
0.458091427160032	3.55608095416499e-05\\
0.466804251670837	7.46478417549226e-05\\
0.475517076181643	0.000129956160054316\\
0.484229900692448	0.00020583096368849\\
0.492942725203254	0.000310647743893789\\
0.50165554971406	0.00042787386091824\\
0.510368374224865	0.000513319965409945\\
0.51908119873567	0.000544166041337894\\
0.527794023246476	0.000551757572502046\\
0.536506847757282	0.000598198231767499\\
0.545219672268087	0.00072532255728868\\
0.553932496778893	0.000916691845700565\\
0.562645321289698	0.00110979549372191\\
0.571358145800504	0.00126778553176784\\
0.580070970311309	0.00145150373819107\\
0.588783794822115	0.00177079397625326\\
0.59749661933292	0.00221998625305476\\
0.606209443843726	0.00263540408074696\\
0.614922268354531	0.00289804086747205\\
0.623635092865337	0.00310796104721053\\
0.632347917376142	0.00347137808499058\\
0.641060741886948	0.00408876570451369\\
0.649773566397754	0.00493681111336753\\
0.658486390908559	0.00594134168532346\\
0.667199215419365	0.00697225096881678\\
0.67591203993017	0.00790008827449429\\
0.684624864440976	0.0087673574365602\\
0.693337688951781	0.00983550317649343\\
0.702050513462587	0.0114257086626233\\
0.710763337973392	0.0136991330850917\\
0.719476162484198	0.0165632192265256\\
0.728188986995003	0.0198177555266266\\
0.736901811505809	0.0233538025743823\\
0.745614636016614	0.0270981331457568\\
0.75432746052742	0.0308557989581319\\
0.763040285038225	0.0344047438231186\\
0.771753109549031	0.037897456598209\\
0.780465934059837	0.0420595762996074\\
0.789178758570642	0.0474839605072478\\
0.797891583081448	0.0537334246260629\\
0.806604407592253	0.0597411732912678\\
0.815317232103059	0.0649784316228472\\
0.824030056613864	0.0696015440782603\\
0.83274288112467	0.0736115166838793\\
0.841455705635475	0.0765596905423846\\
0.850168530146281	0.0782466300996564\\
0.858881354657086	0.0791940782539916\\
0.867594179167892	0.0798590132746077\\
0.876307003678698	0.0798972105114303\\
0.885019828189503	0.0788965580055399\\
0.893732652700309	0.0771937216429829\\
0.902445477211114	0.0752950823602757\\
0.91115830172192	0.0732753300289176\\
0.919871126232725	0.0711087296508606\\
0.928583950743531	0.0689029887856786\\
0.937296775254336	0.0665638838863201\\
0.946009599765142	0.0635902462275359\\
0.954722424275947	0.0593732253230265\\
0.963435248786753	0.0535223799655931\\
0.972148073297558	0.0457219524222072\\
0.980860897808364	0.035801262753421\\
0.989573722319169	0.0245719757468133\\
0.998286546829975	0.0141357515327591\\
1.00699937134078	0.00656443706460968\\
1.01571219585159	0.00238960307824676\\
1.02442502036239	0.000666429528606447\\
1.0331378448732	0.000139155823055189\\
1.041850669384	2.07247924816422e-05\\
};

\addplot [color=mycolor1, dotted, line width=2.0pt]
  table[row sep=crcr]{%
0.0951054710149765	1.99886423719545e-06\\
0.103409827754955	1.0025472941303e-05\\
0.111714184494934	3.76871248247585e-05\\
0.120018541234912	0.000107461781256122\\
0.128322897974891	0.000239150210806029\\
0.13662725471487	0.00043812567814421\\
0.144931611454848	0.000729939207811517\\
0.153235968194827	0.00122689214393061\\
0.161540324934805	0.00213776286050972\\
0.169844681674784	0.00368317188076609\\
0.178149038414762	0.00600465697167908\\
0.186453395154741	0.0091413187170564\\
0.19475775189472	0.013050500693207\\
0.203062108634698	0.0176658262952093\\
0.211366465374677	0.0229719579176975\\
0.219670822114655	0.0289747809502964\\
0.227975178854634	0.0355410034923969\\
0.236279535594613	0.0423316824409031\\
0.244583892334591	0.0489352860080821\\
0.25288824907457	0.055081698139231\\
0.261192605814548	0.060727325608911\\
0.269496962554527	0.0659841418360936\\
0.277801319294506	0.0709456858045785\\
0.286105676034484	0.0755139113158856\\
0.294410032774463	0.0792714300487694\\
0.302714389514441	0.0815750950378708\\
0.31101874625442	0.0820243544176469\\
0.319323102994399	0.080920299741797\\
0.327627459734377	0.0790356021622791\\
0.335931816474356	0.0768904822843106\\
0.344236173214334	0.0744775337644553\\
0.352540529954313	0.0716216354826462\\
0.360844886694292	0.0683260871528868\\
0.36914924343427	0.0647676466255436\\
0.377453600174249	0.0611327330606518\\
0.385757956914227	0.0574107274002916\\
0.394062313654206	0.0534194855892523\\
0.402366670394185	0.0491762702843557\\
0.410671027134163	0.0451252212076837\\
0.418975383874142	0.0416516180445442\\
0.42727974061412	0.0385284101285028\\
0.435584097354099	0.0352721242811826\\
0.443888454094078	0.0318836058310543\\
0.452192810834056	0.0287452831346668\\
0.460497167574035	0.0259869521127099\\
0.468801524314013	0.0233886533496921\\
0.477105881053992	0.0207550197223149\\
0.485410237793971	0.0181393842227568\\
0.493714594533949	0.0158092791956937\\
0.502018951273928	0.0140964221667471\\
0.510323308013906	0.0131218626562036\\
0.518627664753885	0.0125983687118673\\
0.526932021493864	0.0120179941624491\\
0.535236378233842	0.011075700909519\\
0.543540734973821	0.00985610157922289\\
0.551845091713799	0.00862897258290189\\
0.560149448453778	0.00756007532261927\\
0.568453805193757	0.00664457718250329\\
0.576758161933735	0.00582004987746034\\
0.585062518673714	0.00505447798894655\\
0.593366875413692	0.00436177885881761\\
0.601671232153671	0.00377625272853351\\
0.60997558889365	0.00331108967769039\\
0.618279945633628	0.00292889797063996\\
0.626584302373607	0.00256762475302971\\
0.634888659113585	0.00220353773280243\\
0.643193015853564	0.00186278143959961\\
0.651497372593543	0.00157172909404733\\
0.659801729333521	0.00133904615679517\\
0.6681060860735	0.00116840579663564\\
0.676410442813478	0.00103992125515879\\
0.684714799553457	0.000906481765116642\\
0.693019156293436	0.000746600628727465\\
0.701323513033414	0.000596143326607674\\
0.709627869773393	0.000503409529653224\\
0.717932226513371	0.000474151573639683\\
0.72623658325335	0.000478197066609655\\
0.734540939993329	0.000489342898616322\\
0.742845296733307	0.000493337289862967\\
0.751149653473286	0.000474646918475471\\
0.759454010213264	0.00042315247572343\\
0.767758366953243	0.000349569544650669\\
0.776062723693222	0.000273500823677065\\
0.7843670804332	0.00020474909987163\\
0.792671437173179	0.000149694918895169\\
0.800975793913157	0.000117535746564474\\
0.809280150653136	0.000106252362644482\\
0.817584507393115	9.70595027631577e-05\\
0.825888864133093	7.53487418273795e-05\\
0.834193220873072	4.77383316844425e-05\\
0.84249757761305	3.19611138961703e-05\\
0.850801934353029	3.83162503937659e-05\\
0.859106291093007	6.19743457194724e-05\\
0.867410647832986	8.44544623146641e-05\\
0.875715004572965	8.63377546197567e-05\\
0.884019361312943	6.51199138742535e-05\\
0.892323718052922	3.61655343559492e-05\\
0.900628074792901	1.47830826018489e-05\\
0.908932431532879	4.4475841033442e-06\\
0.917236788272858	9.84855202652959e-07\\
};

\addplot [color=black, dashed]
  table[row sep=crcr]{%
0.55	0\\
0.55	0.1\\
};

\end{axis}
\end{tikzpicture}%

%% file: images/comparison-smile.tex
%
%
\definecolor{mycolor1}{rgb}{0.40784,0.47451,0.54118}%
\begin{tikzpicture}

\begin{axis}[%
width=1.8in,
height=1.35in,
at={(0.807in,0.648in)},
scale only axis,
xmin=0.103542106151581,
xmax=1.03906695365906,
xlabel style={font=\bfseries\color{white!4!black}},
xlabel={comparison scores: Smile},
ymin=0,
ymax=0.20,
axis background/.style={fill=white},
axis x line*=bottom,
axis y line*=left,
ytick={0, 0.05, 0.1, 0.15, 0.20},
scaled ticks=false, tick label style={/pgf/number format/fixed},
]
\addplot [color=mycolor1, line width=2.0pt]
  table[row sep=crcr]{%
0.158541561961174	9.84860314436529e-07\\
0.16743575783691	4.89303466424208e-06\\
0.176329953712646	1.71052613468837e-05\\
0.185224149588383	4.20914026006047e-05\\
0.194118345464119	7.30498294815505e-05\\
0.203012541339855	9.03132756352645e-05\\
0.211906737215591	8.34929714069003e-05\\
0.220800933091327	6.94740514219237e-05\\
0.229695128967064	7.15326150666127e-05\\
0.2385893248428	8.63523836856473e-05\\
0.247483520718536	8.88009014082484e-05\\
0.256377716594272	6.71711683486282e-05\\
0.265271912470008	3.60506940830982e-05\\
0.274166108345745	1.3649390782719e-05\\
0.283060304221481	3.63597840953165e-06\\
0.291954500097217	6.81455834584603e-07\\
0.300848695972953	8.98591949752131e-08\\
0.309742891848689	8.55500289818748e-23\\
0.318637087724426	1.59011505367029e-20\\
0.327531283600162	2.07943489614855e-18\\
0.336425479475898	1.91324460453335e-16\\
0.345319675351634	1.23852294670616e-14\\
0.35421387122737	5.64086892346594e-13\\
0.363108067103107	1.80757509791633e-11\\
0.372002262978843	4.07525774797542e-10\\
0.380896458854579	6.46431043882779e-09\\
0.389790654730315	7.21435847320952e-08\\
0.398684850606051	5.66476140280439e-07\\
0.407579046481788	3.12949251497744e-06\\
0.416473242357524	1.21639493388165e-05\\
0.42536743823326	3.32647051956385e-05\\
0.434261634108996	6.40031200723994e-05\\
0.443155829984732	8.66417076572438e-05\\
0.452050025860468	8.2520386165339e-05\\
0.460944221736205	5.52972954217623e-05\\
0.469838417611941	2.60708337490427e-05\\
0.478732613487677	8.64797330772817e-06\\
0.487626809363413	2.01828341631838e-06\\
0.496521005239149	3.31404538071056e-07\\
0.505415201114886	3.82891584608637e-08\\
0.514309396990622	7.90215827340385e-11\\
0.523203592866358	1.51581924882491e-09\\
0.532097788742094	2.04577328077377e-08\\
0.54099198461783	3.94026688513801e-07\\
0.549886180493567	2.78624108870236e-06\\
0.558780376369303	1.39120253263956e-05\\
0.567674572245039	4.95151283450484e-05\\
0.576568768120775	0.000126123533007305\\
0.585462963996511	0.000237283951634021\\
0.594357159872248	0.000355649907013917\\
0.603251355747984	0.000483229877294576\\
0.61214555162372	0.000649632560220223\\
0.621039747499456	0.000817455404273688\\
0.629933943375192	0.000881016128010074\\
0.638828139250929	0.000809249258758199\\
0.647722335126665	0.000696185229382552\\
0.656616531002401	0.000655332927156555\\
0.665510726878137	0.000728999702650394\\
0.674404922753874	0.000888720274825555\\
0.68329911862961	0.00109396095844235\\
0.692193314505346	0.00134582888679872\\
0.701087510381082	0.0016922132129102\\
0.709981706256818	0.00219885797407351\\
0.718875902132554	0.00288122782481628\\
0.727770098008291	0.00365506861294556\\
0.736664293884027	0.00438275649784527\\
0.745558489759763	0.00497218226872084\\
0.754452685635499	0.00550774976614221\\
0.763346881511235	0.00629228504808492\\
0.772241077386972	0.00763008023826355\\
0.781135273262708	0.00957395143945134\\
0.790029469138444	0.0119486558357536\\
0.79892366501418	0.0145561568126872\\
0.807817860889916	0.0173364594447024\\
0.816712056765653	0.0204298178183425\\
0.825606252641389	0.0241533058268188\\
0.834500448517125	0.0289639588920924\\
0.843394644392861	0.0354428122256233\\
0.852288840268597	0.0440632274171168\\
0.861183036144334	0.0547618660240738\\
0.87007723202007	0.0670141584691768\\
0.878971427895806	0.08045822380799\\
0.887865623771542	0.0950171304144247\\
0.896759819647278	0.110287471986711\\
0.905654015523015	0.125163713008353\\
0.914548211398751	0.13816531161643\\
0.923442407274487	0.147738946672716\\
0.932336603150223	0.152073916771823\\
0.941230799025959	0.149340460813522\\
0.950124994901696	0.138620885571671\\
0.959019190777432	0.12022050546135\\
0.967913386653168	0.0955389810212515\\
0.976807582528904	0.0676529724073143\\
0.98570177840464	0.0413568523067333\\
0.994595974280376	0.0211525563891523\\
1.00349017015611	0.00878758503326608\\
1.01238436603185	0.00288578986071544\\
1.02127856190759	0.000729622619585072\\
1.03017275778332	0.000137761225554461\\
1.03906695365906	1.80677298038545e-05\\
};

\addplot [color=mycolor1, dotted, line width=2.0pt]
  table[row sep=crcr]{%
0.103542106151581	1.97965329380939e-06\\
0.111263889953344	9.96522019043613e-06\\
0.118985673755106	3.96548934385684e-05\\
0.126707457556869	0.0001281811092785\\
0.134429241358632	0.000340849572728803\\
0.142151025160395	0.000764192112860968\\
0.149872808962157	0.00148755408693074\\
0.15759459276392	0.00260500075796697\\
0.165316376565683	0.00422878642420737\\
0.173038160367446	0.00646908925880183\\
0.180759944169208	0.00937168119063731\\
0.188481727970971	0.0129204739309914\\
0.196203511772734	0.0171366118961522\\
0.203925295574497	0.0221308997066144\\
0.211647079376259	0.0279340280259778\\
0.219368863178022	0.0342839897117612\\
0.227090646979785	0.0406900009240984\\
0.234812430781548	0.046796131829352\\
0.24253421458331	0.0526445795344851\\
0.250255998385073	0.058472372988677\\
0.257977782186836	0.0642494924130048\\
0.265699565988598	0.0695424620011255\\
0.273421349790361	0.0739040029972644\\
0.281143133592124	0.0772323826618271\\
0.288864917393887	0.0796162382440532\\
0.296586701195649	0.0810176659276088\\
0.304308484997412	0.0813654251289873\\
0.312030268799175	0.0808570352760976\\
0.319752052600938	0.0798900540521928\\
0.3274738364027	0.0786697221946227\\
0.335195620204463	0.0770388761329465\\
0.342917404006226	0.074725981082969\\
0.350639187807989	0.0716220865961072\\
0.358360971609751	0.0678245839334743\\
0.366082755411514	0.0636016861460449\\
0.373804539213277	0.0593551467685545\\
0.38152632301504	0.0554481299768342\\
0.389248106816802	0.0519787334908226\\
0.396969890618565	0.0487846740589784\\
0.404691674420328	0.0456751677214083\\
0.412413458222091	0.0425550069477142\\
0.420135242023853	0.0393258428705843\\
0.427857025825616	0.03583555682215\\
0.435578809627379	0.0320921230788624\\
0.443300593429142	0.028454778777444\\
0.451022377230904	0.0254209674715198\\
0.458744161032667	0.0231883041158295\\
0.46646594483443	0.0215327106579875\\
0.474187728636193	0.0200796727495238\\
0.481909512437955	0.0185929315543584\\
0.489631296239718	0.0170349260138299\\
0.497353080041481	0.0154870634855211\\
0.505074863843244	0.0140387113032415\\
0.512796647645006	0.0127081469948793\\
0.520518431446769	0.0114639687353941\\
0.528240215248532	0.0102951904146453\\
0.535961999050295	0.00921226340297045\\
0.543683782852057	0.00819814399564371\\
0.55140556665382	0.00722708379287802\\
0.559127350455583	0.006304686482579\\
0.566849134257346	0.00545588388692362\\
0.574570918059108	0.0046948802424275\\
0.582292701860871	0.00403117182096843\\
0.590014485662634	0.00347314014102941\\
0.597736269464397	0.00301285876672011\\
0.605458053266159	0.00262793755827908\\
0.613179837067922	0.00230459515235588\\
0.620901620869685	0.00204404223670463\\
0.628623404671448	0.00183736773993751\\
0.63634518847321	0.00164520694898445\\
0.644066972274973	0.00142487820009187\\
0.651788756076736	0.00117500769691852\\
0.659510539878498	0.000932185112712046\\
0.667232323680261	0.000725398247450321\\
0.674954107482024	0.000556815029174333\\
0.682675891283787	0.00042234408635411\\
0.690397675085549	0.000325542561039743\\
0.698119458887312	0.000262610838173264\\
0.705841242689075	0.000212912398825671\\
0.713563026490838	0.000157354448944184\\
0.7212848102926	9.75526139701538e-05\\
0.729006594094363	4.85709686649436e-05\\
0.736728377896126	1.9020351340583e-05\\
0.744450161697889	5.86688064055824e-06\\
0.752171945499651	1.75964908620464e-06\\
0.759893729301414	2.08286092431187e-06\\
0.767615513103177	6.81366631114987e-06\\
0.77533729690494	1.9443835667092e-05\\
0.783059080706702	4.34456158978788e-05\\
0.790780864508465	7.67781827027348e-05\\
0.798502648310228	0.000110002113826097\\
0.806224432111991	0.000132187276055305\\
0.813946215913753	0.000137495417835998\\
0.821667999715516	0.000124925319626131\\
0.829389783517279	9.71373124808055e-05\\
0.837111567319042	6.24297026484469e-05\\
0.844833351120804	3.21569246038976e-05\\
0.852555134922567	1.30077423665119e-05\\
0.86027691872433	4.08492344843652e-06\\
0.867998702526093	9.84855202652959e-07\\
};

\addplot [color=black, dashed]
  table[row sep=crcr]{%
0.55	0\\
0.55	0.2\\
};

\end{axis}
\end{tikzpicture}%

%% file: content/06-conclusion.tex
\newpage
\section{Conclusion and Future Work}

To solve the privacy-related issue with real datasets and overcome the shortage of training data, we introduce PCA-FR-Guided sampling for generating mated samples in a non-deterministic manner. Unlike controlled face image editing techniques operating in the latent space, we apply PCA to find semantically meaningful directions. While moving latent vectors into these directions, the identity of the underlying face image is preserved by progressive supervision with a pre-trained face recognition system. With the newly created Synthetic Mated samples dataset (SymFace Dataset) with 77,034 images, we have evaluated state-of-the-art face quality assessment algorithms and biometric comparison score analysis to validate the applicability of the proposed approach. The well-separated distributions between mated and non-mated comparison scores indicate that synthetic mated samples generated with PCA-FR-Guided sampling are well suited for biometric performance tests. Furthermore, the analysis of face quality and the comparison scores is comparable to observations made in real datasets, indicating the usefulness of the proposed approach. 

Although this work has illustrated to include synthetic samples in face recognition performance tests, we emphasise the open challenge to mimic the full extent of intra-identity variation measurable in bona fide datasets. Future works should also focus on an exploratory analysis of the different principal components, thereby exploring the latent space of StyleGAN and strengthening the understanding of the internal data representation. We foresee using the proposed approach to reduce the need for large training sets and minimise the demographic bias by diversifying latent space in synthetic generation schemes.

%% file: content/appendix.tex
\section*{Appendix}
\label{sec:appendix}

\input{tables/qscores-kl-divergences}

\input{tables/cscores-kl-divergences}

%% file: tables/qscores-kl-divergences.tex
\begin{table}[]
\centering
\begin{tabular}{|l|l|c|c|c|c|}
\hline
\multicolumn{2}{|l|}{Datasets}                                 & \multicolumn{1}{l|}{PCA-FR} & \multicolumn{1}{l|}{Illumination Quality} & \multicolumn{1}{l|}{Smile} & \multicolumn{1}{l|}{Yaw} \\ \hline
\multicolumn{1}{|c|}{\multirow{2}{*}{FRGC v2.0}} & SER-FIQ     & 1.17                        & 1.19                                      & 1.13                       & 3.25                     \\ \cline{2-2}
\multicolumn{1}{|c|}{}                           & FaceQnet v1 & 0.11                        & 0.12                                      & 0.11                       & 0.13                     \\ \hline
\multirow{2}{*}{Base Images}                     & SER-FIQ     & 0.02                        & 0.01                                      & 0.01                       & 0.27                     \\ \cline{2-2}
                                                 & FaceQnet v1 & 0.01                        & 0.01                                      & 0.01                       & 0.01                     \\ \hline
\end{tabular}
\caption{KL-Divergences between quality score distributions in Figure \ref{fig:qscore-distributions}.}
\label{tab:qscores-kl-divergences}
\end{table}

%% file: tables/cscores-kl-divergences.tex
\begin{table}[]
\centering
\begin{tabular}{|l|l|c|c|c|c|}
\hline
\multicolumn{2}{|l|}{Datasets}                               & \multicolumn{1}{l|}{PCA-FR} & \multicolumn{1}{l|}{Illumination Quality} & \multicolumn{1}{l|}{Smile} & \multicolumn{1}{l|}{Yaw} \\ \hline
\multicolumn{1}{|c|}{\multirow{2}{*}{FRGC v2.0}} & Mated     & 0.42                        & 0.79                                      & 0.17                       & 0.72                     \\ \cline{2-2}
\multicolumn{1}{|c|}{}                           & Non-mated & 0.27                        & 0.28                                      & 0.32                       & 0.33                     \\ \hline
\multirow{2}{*}{Base Images}                     & Mated     & /                           & /                                         & /                          & /                        \\ \cline{2-2}
                                                 & Non-mated & 0.01                        & 0.20                                      & 0.02                       & 0.01                     \\ \hline
\end{tabular}
\caption{KL-Divergences between comparison score distributions in Figure \ref{fig:comparison_scores}.}
\label{tab:cscores-kl-divergences}
\end{table}